%% file: main.tex
\documentclass[fleqn,10pt]{wlscirep}
\usepackage{hyphenat}
\usepackage[T1]{fontenc}
\usepackage{amssymb}
\usepackage{booktabs}
\usepackage{multirow}
\usepackage{tabularx}   
\usepackage{array}
\usepackage{ragged2e}
\usepackage{geometry}
\usepackage{multirow}
\usepackage{array}
\usepackage{makecell}
\usepackage{float}
\usepackage{array}
\usepackage{ragged2e} 
\usepackage{longtable}
\usepackage{xltabular}
\usepackage{booktabs}
\usepackage{multirow}
\usepackage{makecell}
\usepackage[most]{tcolorbox}
\title{Predicting first-episode homelessness among US Veterans using longitudinal EHR data: time-varying models and social risk factors}

\author[1,2]{Rohan Pandey}
\author[1]{Haijuan Yan}
\author[1,2,3]{Hong Yu}
\author[4,5,6*]{Jack Tsai}
\affil[1]{Center for Healthcare Organization and Implementation Research, VA Bedford Health Care, MA, USA}
\affil[2]{Manning College of Information and Computer Sciences, UMass Amherst, MA, USA}
\affil[3]{Miner School of Computer and Information Sciences, UMass Lowell, MA, USA}
\affil[4]{National Center on Homelessness among Veterans, VA Homeless Programs Office, Washington, DC, USA }
\affil[5]{School of Public Health, University of Texas Health Science Center at Houston, Houston, TX, USA }
\affil[6]{Department of Psychiatry, Yale University School of Medicine, New Haven, CT, USA}
\affil[*]{*Corresponding author: Jack Tsai (Jack.Tsai@uth.tmc.edu)}

\begin{abstract}
Homelessness among US veterans remains a critical public health challenge, yet risk prediction offers a pathway for proactive intervention. In this retrospective prognostic study, we analyzed electronic health record (EHR) data from 4,276,403 Veterans Affairs patients during a 2016 observation period to predict first-episode homelessness occurring 3-12 months later in 2017 (prevalence: 0.32-1.19\%). We constructed static and time-varying EHR representations, utilizing clinician-informed logic to model the persistence of clinical conditions and social risks over time. We then compared the performance of classical machine learning, transformer-based masked language models, and fine-tuned large language models (LLMs). We demonstrate that incorporating social and behavioral factors into longitudinal models improved precision-recall area under the curve (PR-AUC) by 15-30\%. In the top 1\% risk tier, models yielded positive predictive values ranging from 3.93-4.72\% at 3 months, 7.39-8.30\% at 6 months, 9.84-11.41\% at 9 months, and 11.65-13.80\% at 12 months across model architectures. Large language models underperformed encoder-based models on discrimination but showed smaller performance disparities across racial groups. These results demonstrate that longitudinal, socially informed EHR modeling concentrates homelessness risk into actionable strata, enabling targeted and data-informed prevention strategies for at-risk veterans.
\end{abstract}
\begin{document}

\flushbottom
\maketitle
%
%
\section{Introduction}

Homelessness remains a major public health challenge in the United States (U.S.). On a single night in January 2024, an estimated 771,480 people were experiencing homelessness, an 18\% increase from 2023. Veterans account for about 5\% of all US adults experiencing homelessness, with 32,882 veterans experiencing homelessness on a single night in 2024, underscoring persistent risk in this population\cite{2024AnnualHomelessness}. While there has been substantial progress in reducing veteran homelessness in the past 15 years, veteran homelessness remains one of the top priorities of the U.S. Department of Veterans Affairs (VA)\cite{tsai2021problem}. In the broader U.S. population, homelessness is at record high levels\cite{2024AnnualHomelessness}, underscoring the need for scalable prediction and prevention strategies.

Electronic health records (EHRs) offer system-wide visibility into clinical, utilization, and social-needs data that could enable earlier identification of patients at risk for first-episode homelessness. The VA operates the largest integrated healthcare system in the U.S. and has maintained a national EHR system for more than four decades. Prior EHR-based studies in the VA have documented associations between physical comorbidities, mental health conditions (e.g., PTSD), and substance use disorders among veterans \cite{saleem2013next, fink2022comparing}. Yet most of this work has been cross-sectional or relied on regression-based analyses with limited use of predictive modeling.

Only a handful of studies have focused on predicting homelessness in the veteran population\cite{tsai2017one, elbogen2025identifying, tsai2025retrospective, tsai2024predicting}, and these studies have not compared diverse modeling approaches using contemporary machine-learning methods. In addition, although longitudinal EHR data are available, time-varying representations of patient histories are rarely used, and key social factors, including socioeconomic status, education, housing, and access to healthy food, remain inconsistently captured and underused at scale \cite{chatterjee2025measurement, devanarayan2025association, hau2025social}. Homelessness risk and access to preventive services vary across sociodemographic groups, including racial, age, and gender subgroups among Veterans\cite{montgomery2020housing}, raising concerns that prediction models could either exacerbate or mitigate existing inequities. These gaps are particularly important for clinical translation, as models must inform proactive outreach and triage. Early, accurate identification of patients at elevated risk for first-episode homelessness could enable targeted outreach, social-work referral, and prevention services within large health systems such as the VA.

To address these gaps, we used national VA EHR data to conduct a retrospective prognostic modeling study. We compared the performance of different models to predict first-episode homelessness among veterans over 3, 6, 9, and 12 months (multiple future horizons) following a 1-year observation. Our contributions are threefold: (1) construction of time-varying EHR representations using clinically informed condition persistence rules that govern how long diagnoses and social/behavioral problems remain active after they are last recorded; (2) explicit incorporation of social and behavioral factors of health (SBFH) alongside demographics, service utilization, and medical and behavioral health conditions; and (3) a comparison of architecturally distinct models under a consistent evaluation protocol. A schematic overview of the study design, observation/prediction windows, and longitudinal representation construction is shown in Fig. 1.

\begin{figure}[H]
    \centering
    \includegraphics[width=0.75\linewidth, height=11cm]{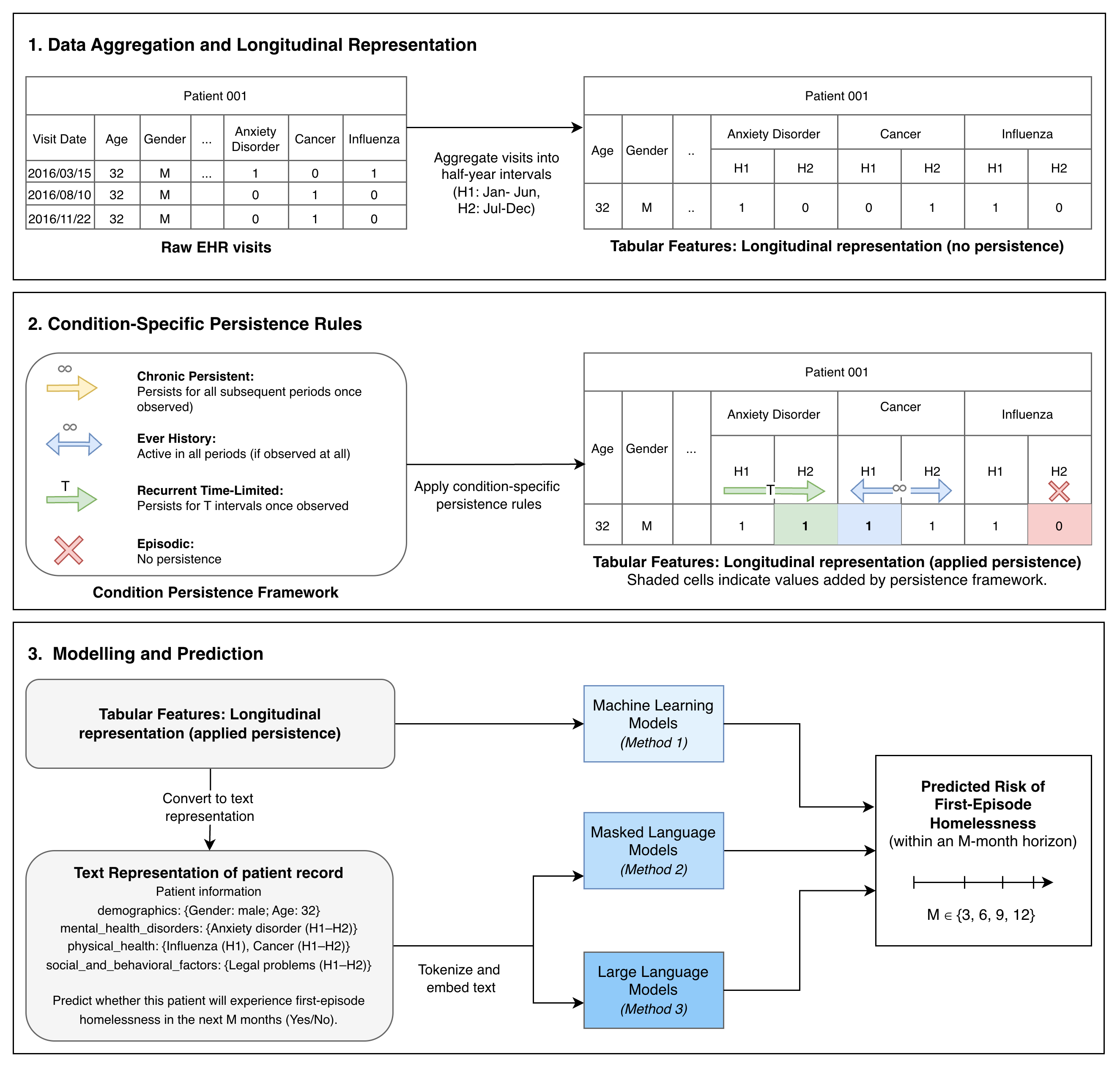}
    \caption{\textbf{Study schematic: longitudinal EHR representations and homelessness risk prediction. \\}
    Raw visit-level EHR data for an example patient are aggregated into fixed half-year intervals to form a baseline longitudinal representation without persistence (H1: January-June 2016; H2: July-December 2016). A condition persistence framework applies domain-specific persistence policies (chronic persistent, ever-history, recurrent time-limited, episodic) to carry forward or limit condition activity across intervals, yielding a longitudinal representation in which shaded cells indicate values added by the framework. The resulting patient-level profiles are converted into natural-language prompts organized by clinical domain (example shown) and used as inputs to three model classes: machine-learning models, masked-language models, and large-language models to independently predict the risk of first-episode homelessness within M months after baseline. \textit{The example illustrates the half-year temporal representation; analogous representations were also constructed at quarterly and yearly aggregation levels for comparative analyses.}
}
    \label{fig:1}
\end{figure}
\section{Results}
\subsection{Prevalence of Homelessness}
Of 6,105,401 veterans with a VA visit in 2016, 4,276,403 (70.0\%) met eligibility for prediction (Supplementary Figure \hyperref[supp_fig_1]{1}). Within the 3-month window, 13,728 of 4,276,403 veterans experienced homelessness (0.32\%). Prevalence increased with longer windows: 26,818 (0.63\%) within 6 months, 39,420 (0.92\%) within 9 months, and 51,002 (1.19\%) within 12 months. These four cumulative prediction cohorts therefore contained 13,728, 26,818, 39,420, and 51,002 cases, respectively, with the remainder serving as non-cases. Detailed cohort characteristics are presented in Supplementary Table \hyperref[supp_table_1]{1}.

\begin{table*}[p] 
\centering
\scriptsize 
\renewcommand{\arraystretch}{0.9} 
\begin{tabularx}{\textwidth}{l X l l l}
\toprule
\textbf{Model Class} & \textbf{Model} & \textbf{Input Representation} & \textbf{PR AUC, \%} & \textbf{ROC AUC, \%} \\
\midrule

\multicolumn{5}{c}{\textbf{Prediction Window = 3 Months}} \\
\midrule
\multirow{6}{*}{Machine Learning} 
 & \multirow{2}{*}{Elastic Net Logistic Regression} & Static & 1.73 (1.44, 2.19) & 76.88 (75.03, 78.64) \\ 
 & & Time Varying & 1.87 (1.54, 2.45) & 76.71 (74.79, 78.46) \\ \cmidrule(lr){2-5}
 & \multirow{2}{*}{Random Forest} & Static & 1.89 (1.57, 2.36) & 78.38 (76.57, 80.04) \\ 
 & & Time Varying & 2.11 (1.76, 2.68) & 79.27 (77.54, 80.94) \\ \cmidrule(lr){2-5}
 & \multirow{2}{*}{XGBoost} & Static & 2.08 (1.71, 2.8) & 79.57 (77.87, 81.22) \\ 
 & & Time Varying & 2.22 (1.83, 2.91) & 79.73 (77.95, 81.37) \\ \midrule
 
\multirow{4}{*}{\begin{tabular}[c]{@{}l@{}}Masked Language\\ Models\end{tabular}} 
 & \multirow{2}{*}{ModernBERT} & Static & 1.91 (1.57, 2.43) & 77.84 (76, 79.56) \\ 
 & & Time Varying & \textbf{2.39 (1.8, 3.34)} & 77.28 (75.39, 79) \\ \cmidrule(lr){2-5}
 & \multirow{2}{*}{BioClinical ModernBERT} & Static & 1.96 (1.63, 2.45) & 78.74 (76.93, 80.42) \\ 
 & & Time Varying & 2.34 (1.91, 3.09) & 78.78 (76.97, 80.53) \\ \midrule

\multirow{4}{*}{\begin{tabular}[c]{@{}l@{}}Large Language\\ Models\end{tabular}} 
 & \multirow{2}{*}{Llama-3.1-8B} & Static & 1.75 (1.47, 2.21) & 78.36 (76.59, 80.04) \\ 
 & & Time Varying & 2.16 (1.77, 2.76) & 79.12 (77.43, 80.79) \\ \cmidrule(lr){2-5}
 & \multirow{2}{*}{OpenBioLLM-8B} & Static & 1.42 (1.23, 1.76) & 76.97 (75.11, 78.82) \\ 
 & & Time Varying & 2.32 (1.81, 3.1) & 78.31 (76.49, 79.95) \\ 
\midrule

\multicolumn{5}{c}{\textbf{Prediction Window = 6 Months}} \\
\midrule
\multirow{6}{*}{Machine Learning} 
 & \multirow{2}{*}{Elastic Net Logistic Regression} & Static & 3.23 (2.8, 3.84) & 74.82 (73.45, 76.18) \\ 
 & & Time Varying & 3.28 (2.83, 3.97) & 74.83 (73.46, 76.18) \\ \cmidrule(lr){2-5}
 & \multirow{2}{*}{Random Forest} & Static & 3.73 (3.24, 4.42) & 77.82 (76.59, 79.05) \\ 
 & & Time Varying & 3.86 (3.32, 4.61) & 77.98 (76.77, 79.24) \\ \cmidrule(lr){2-5}
 & \multirow{2}{*}{XGBoost} & Static & 3.62 (3.16, 4.28) & 78.53 (77.3, 79.75) \\ 
 & & Time Varying & \textbf{4.13 (3.48, 5.01)} & 78.45 (77.2, 79.67) \\ \midrule
 
\multirow{4}{*}{\begin{tabular}[c]{@{}l@{}}Masked Language\\ Models\end{tabular}} 
 & \multirow{2}{*}{ModernBERT} & Static & 3.43 (2.98, 4.03) & 77.32 (76.11, 78.59) \\ 
 & & Time Varying & 3.54 (3.07, 4.14) & 77.49 (76.23, 78.73) \\ \cmidrule(lr){2-5}
 & \multirow{2}{*}{BioClinical ModernBERT} & Static & 3.43 (2.96, 4.03) & 77.02 (75.75, 78.22) \\ 
 & & Time Varying & 3.58 (3.12, 4.21) & 76.6 (75.33, 77.93) \\ \midrule

\multirow{4}{*}{\begin{tabular}[c]{@{}l@{}}Large Language\\ Models\end{tabular}} 
 & \multirow{2}{*}{Llama-3.1-8B} & Static & 3.44 (3.01, 4.02) & 77.26 (76.02, 78.53) \\ 
 & & Time Varying & 4.12 (3.52, 4.98) & 78.43 (77.21, 79.66) \\ \cmidrule(lr){2-5}
 & \multirow{2}{*}{OpenBioLLM-8B} & Static & 3.55 (3.15, 4.13) & 79.02 (77.83, 80.19) \\ 
 & & Time Varying & 3.65 (3.19, 4.36) & 78.33 (77.11, 79.54) \\ 
\midrule

\multicolumn{5}{c}{\textbf{Prediction Window = 9 Months}} \\
\midrule
\multirow{6}{*}{Machine Learning} 
 & \multirow{2}{*}{Elastic Net Logistic Regression} & Static & 4.47 (4.02, 5.11) & 76.4 (75.32, 77.39) \\ 
 & & Time Varying & 4.52 (4.05, 5.22) & 76.3 (75.22, 77.32) \\ \cmidrule(lr){2-5}
 & \multirow{2}{*}{Random Forest} & Static & 5.07 (4.56, 5.78) & 78.97 (78.02, 79.87) \\ 
 & & Time Varying & 5.23 (4.67, 5.95) & 79.03 (78.11, 79.95) \\ \cmidrule(lr){2-5}
 & \multirow{2}{*}{XGBoost} & Static & 4.93 (4.43, 5.62) & 78.62 (77.64, 79.55) \\ 
 & & Time Varying & 5.14 (4.59, 5.89) & 78.37 (77.39, 79.31) \\ \midrule
 
\multirow{4}{*}{\begin{tabular}[c]{@{}l@{}}Masked Language\\ Models\end{tabular}} 
 & \multirow{2}{*}{ModernBERT} & Static & 4.95 (4.44, 5.63) & 78.38 (77.35, 79.32) \\ 
 & & Time Varying & \textbf{5.27 (4.68, 6.01)} & 78.08 (77.08, 79.03) \\ \cmidrule(lr){2-5}
 & \multirow{2}{*}{BioClinical ModernBERT} & Static & 5.14 (4.62, 5.84) & 78.7 (77.75, 79.64) \\ 
 & & Time Varying & 5.16 (4.65, 5.84) & 78.72 (77.75, 79.68) \\ \midrule

\multirow{4}{*}{\begin{tabular}[c]{@{}l@{}}Large Language\\ Models\end{tabular}} 
 & \multirow{2}{*}{Llama-3.1-8B} & Static & 4.03 (3.65, 4.53) & 77.47 (76.46, 78.38) \\ 
 & & Time Varying & 5.19 (4.64, 5.92) & 78.56 (77.6, 79.46) \\ \cmidrule(lr){2-5}
 & \multirow{2}{*}{OpenBioLLM-8B} & Static & 5.07 (4.58, 5.66) & 80.2 (79.28, 81.11) \\ 
 & & Time Varying & 4.42 (3.97, 5.03) & 76.68 (75.64, 77.68) \\ 
\midrule

\multicolumn{5}{c}{\textbf{Prediction Window = 12 Months}} \\
\midrule
\multirow{6}{*}{Machine Learning} 
 & \multirow{2}{*}{Elastic Net Logistic Regression} & Static & 5.66 (5.17, 6.33) & 75.75 (74.77, 76.71) \\ 
 & & Time Varying & 5.72 (5.22, 6.41) & 75.75 (74.74, 76.72) \\ \cmidrule(lr){2-5}
 & \multirow{2}{*}{Random Forest} & Static & 6.21 (5.68, 6.85) & 78.32 (77.43, 79.16) \\ 
 & & Time Varying & 6.39 (5.83, 7.06) & 78.39 (77.5, 79.24) \\ \cmidrule(lr){2-5}
 & \multirow{2}{*}{XGBoost} & Static & 6.47 (5.86, 7.23) & 78.1 (77.2, 78.96) \\ 
 & & Time Varying & \textbf{6.72 (6.06, 7.53)} & 78.05 (77.14, 78.92) \\ \midrule
 
\multirow{4}{*}{\begin{tabular}[c]{@{}l@{}}Masked Language\\ Models\end{tabular}} 
 & \multirow{2}{*}{ModernBERT} & Static & 6.11 (5.62, 6.74) & 78.5 (77.59, 79.35) \\ 
 & & Time Varying & 6.29 (5.76, 6.97) & 77.84 (76.9, 78.75) \\ \cmidrule(lr){2-5}
 & \multirow{2}{*}{BioClinical ModernBERT} & Static & 6.19 (5.65, 6.89) & 78.65 (77.78, 79.5) \\ 
 & & Time Varying & 6.65 (6.03, 7.4) & 77.99 (77.07, 78.86) \\ \midrule

\multirow{4}{*}{\begin{tabular}[c]{@{}l@{}}Large Language\\ Models\end{tabular}} 
 & \multirow{2}{*}{Llama-3.1-8B} & Static & 5.01 (4.57, 5.54) & 73.72 (72.69, 74.74) \\ 
 & & Time Varying & 5.66 (5.16, 6.32) & 77.3 (76.38, 78.15) \\ \cmidrule(lr){2-5}
 & \multirow{2}{*}{OpenBioLLM-8B} & Static & 5.65 (5.17, 6.28) & 77.85 (76.93, 78.69) \\ 
 & & Time Varying & 5.99 (5.49, 6.66) & 78.06 (77.18, 78.93) \\ 
\bottomrule

\end{tabularx}
\caption{Performance of predictive models across prediction windows for static and proposed time-varying EHR representations. Bold indicates highest PR-AUC within each prediction window.}
\label{tab:model_performance}
\end{table*}

\begin{table*}[p]
\centering
\scriptsize
\renewcommand{\arraystretch}{0.9}
\begin{tabularx}{\textwidth}{l X l l l l l}
\toprule
\textbf{Model Class} & \textbf{Model} & \makecell[c]{\textbf{Risk Group}\\\textbf{Size, P}} & \makecell[c]{\textbf{Sensitivity (\%)}\\\textbf{@ top-P}} & \makecell[c]{\textbf{Specificity (\%)}\\\textbf{@ top-P}} & \makecell[c]{\textbf{PPV (\%)}\\\textbf{@ top-P}} & \makecell[c]{\textbf{O/E Ratio}\\\textbf{@ top-P}} \\
\midrule

\multicolumn{7}{c}{\textbf{Prediction Window = 3 Months}} \\
\midrule
\multirow{6}{*}{Machine Learning} 
 & Elastic Net Logistic Regression & 0.01 & 12.68 (10.35, 15.16) & 99.04 (99.03, 99.05) & 4.07 (3.32, 4.86) & 12.68 (10.35, 15.15) \\
 & & 0.05 & 30.76 (27.26, 34.11) & 95.08 (95.07, 95.09) & 1.97 (1.75, 2.19) & 6.15 (5.45, 6.82) \\ \cmidrule(lr){2-7}
 & Random Forest & 0.01 & 13.41 (10.93, 16.03) & 99.04 (99.03, 99.05) & 4.30 (3.51, 5.14) & 13.41 (10.93, 16.03) \\
 & & 0.05 & 33.09 (29.59, 36.59) & 95.09 (95.08, 95.10) & 2.12 (1.90, 2.35) & 6.62 (5.92, 7.32) \\ \cmidrule(lr){2-7}
 & XGBoost & 0.01 & 12.83 (10.64, 15.45) & 99.04 (99.03, 99.05) & 4.12 (3.41, 4.96) & 12.83 (10.64, 15.45) \\
 & & 0.05 & 32.36 (29.01, 35.71) & 95.09 (95.08, 95.10) & 2.08 (1.86, 2.29) & 6.47 (5.80, 7.14) \\ \midrule
 
\multirow{4}{*}{\begin{tabular}[c]{@{}l@{}}Masked Language\\ Models\end{tabular}} 
 & ModernBERT & 0.01 & 12.24 (9.77, 14.58) & 99.04 (99.03, 99.04) & 3.93 (3.13, 4.68) & 12.25 (9.76, 14.58) \\
 & & 0.05 & 30.17 (26.96, 33.53) & 95.08 (95.07, 95.09) & 1.94 (1.73, 2.15) & 6.04 (5.39, 6.70) \\ \cmidrule(lr){2-7}
 & BioClinical ModernBERT & 0.01 & 14.72 (12.39, 17.35) & 99.04 (99.04, 99.05) & 4.72 (3.97, 5.56) & 14.72 (12.39, 17.34) \\
 & & 0.05 & 32.80 (29.30, 36.30) & 95.09 (95.08, 95.10) & 2.10 (1.88, 2.33) & 6.56 (5.86, 7.26) \\ \midrule

\multirow{4}{*}{\begin{tabular}[c]{@{}l@{}}Large Language\\ Models\end{tabular}} 
 & Llama-3.1-8B & 0.01 & 13.85 (11.37, 16.33) & 99.04 (99.03, 99.05) & 4.44 (3.65, 5.24) & 13.85 (11.37, 16.32) \\
 & & 0.05 & 32.51 (29.15, 36.15) & 95.09 (95.08, 95.10) & 2.09 (1.87, 2.32) & 6.50 (5.83, 7.23) \\ \cmidrule(lr){2-7}
 & OpenBioLLM-8B & 0.01 & 13.41 (11.37, 16.62) & 99.04 (99.03, 99.05) & 4.30 (3.65, 5.33) & 13.41 (11.37, 16.61) \\
 & & 0.05 & 31.92 (28.57, 35.57) & 95.09 (95.08, 95.10) & 2.05 (1.83, 2.28) & 6.38 (5.71, 7.11) \\ 

\midrule

\multicolumn{7}{c}{\textbf{Prediction Window = 6 Months}} \\
\midrule
\multirow{6}{*}{Machine Learning} 
 & Elastic Net Logistic Regression & 0.01 & 11.78 (10.14, 13.50) & 99.07 (99.06, 99.08) & 7.39 (6.36, 8.46) & 11.78 (10.14, 13.49) \\
 & & 0.05 & 27.59 (25.21, 29.90) & 95.14 (95.13, 95.16) & 3.46 (3.16, 3.75) & 5.52 (5.04, 5.98) \\ \cmidrule(lr){2-7}
 & Random Forest & 0.01 & 12.53 (10.81, 14.24) & 99.07 (99.06, 99.08) & 7.86 (6.78, 8.93) & 12.53 (10.81, 14.24) \\
 & & 0.05 & 29.53 (27.22, 31.99) & 95.15 (95.14, 95.17) & 3.70 (3.41, 4.01) & 5.91 (5.44, 6.40) \\ \cmidrule(lr){2-7}
 & XGBoost & 0.01 & 12.53 (10.81, 14.32) & 99.07 (99.06, 99.08) & 7.86 (6.78, 8.98) & 12.53 (10.81, 14.31) \\
 & & 0.05 & 29.31 (26.92, 31.69) & 95.15 (95.14, 95.17) & 3.68 (3.38, 3.98) & 5.86 (5.38, 6.34) \\ \midrule
 
\multirow{4}{*}{\begin{tabular}[c]{@{}l@{}}Masked Language\\ Models\end{tabular}} 
 & ModernBERT & 0.01 & 12.53 (10.81, 14.32) & 99.07 (99.06, 99.08) & 7.86 (6.78, 8.98) & 12.53 (10.81, 14.31) \\
 & & 0.05 & 30.05 (27.67, 32.44) & 95.16 (95.14, 95.17) & 3.77 (3.47, 4.07) & 6.01 (5.53, 6.49) \\ \cmidrule(lr){2-7}
 & BioClinical ModernBERT & 0.01 & 13.27 (11.41, 14.91) & 99.08 (99.07, 99.09) & 8.33 (7.15, 9.35) & 13.27 (11.40, 14.91) \\
 & & 0.05 & 29.53 (27.07, 31.99) & 95.15 (95.14, 95.17) & 3.70 (3.39, 4.01) & 5.91 (5.41, 6.40) \\ \midrule

\multirow{4}{*}{\begin{tabular}[c]{@{}l@{}}Large Language\\ Models\end{tabular}} 
 & Llama-3.1-8B & 0.01 & 12.83 (11.04, 14.47) & 99.07 (99.06, 99.08) & 8.04 (6.92, 9.07) & 12.83 (11.03, 14.46) \\
 & & 0.05 & 30.95 (28.49, 33.41) & 95.16 (95.15, 95.18) & 3.88 (3.57, 4.19) & 6.19 (5.70, 6.68) \\ \cmidrule(lr){2-7}
 & OpenBioLLM-8B & 0.01 & 11.63 (10.37, 14.10) & 99.07 (99.06, 99.08) & 7.30 (6.50, 8.84) & 11.63 (10.36, 14.09) \\
 & & 0.05 & 30.28 (28.11, 32.96) & 95.16 (95.15, 95.18) & 3.80 (3.53, 4.13) & 6.06 (5.62, 6.59) \\ 

\midrule

\multicolumn{7}{c}{\textbf{Prediction Window = 9 Months}} \\
\midrule
\multirow{6}{*}{Machine Learning} 
 & Elastic Net Logistic Regression & 0.01 & 10.71 (9.34, 12.02) & 99.09 (99.08, 99.10) & 9.87 (8.60, 11.08) & 10.71 (9.33, 12.02) \\
 & & 0.05 & 26.69 (24.71, 28.61) & 95.20 (95.18, 95.22) & 4.92 (4.55, 5.27) & 5.34 (4.94, 5.72) \\ \cmidrule(lr){2-7}
 & Random Forest & 0.01 & 12.23 (10.86, 13.65) & 99.10 (99.09, 99.12) & 11.27 (10.00, 12.58) & 12.23 (10.85, 13.64) \\
 & & 0.05 & 29.17 (27.19, 31.05) & 95.22 (95.21, 95.24) & 5.38 (5.01, 5.72) & 5.83 (5.44, 6.21) \\ \cmidrule(lr){2-7}
 & XGBoost & 0.01 & 11.97 (10.55, 13.39) & 99.10 (99.09, 99.11) & 11.04 (9.72, 12.34) & 11.97 (10.55, 13.39) \\
 & & 0.05 & 28.92 (26.89, 30.95) & 95.22 (95.20, 95.24) & 5.33 (4.96, 5.71) & 5.78 (5.38, 6.19) \\ \midrule
 
\multirow{4}{*}{\begin{tabular}[c]{@{}l@{}}Masked Language\\ Models\end{tabular}} 
 & ModernBERT & 0.01 & 12.08 (10.65, 13.50) & 99.10 (99.09, 99.12) & 11.13 (9.82, 12.44) & 12.08 (10.65, 13.49) \\
 & & 0.05 & 29.98 (27.90, 32.01) & 95.23 (95.21, 95.25) & 5.53 (5.14, 5.90) & 6.00 (5.58, 6.40) \\ \cmidrule(lr){2-7}
 & BioClinical ModernBERT & 0.01 & 11.52 (10.25, 12.99) & 99.10 (99.09, 99.11) & 10.62 (9.44, 11.97) & 11.52 (10.24, 12.98) \\
 & & 0.05 & 30.44 (28.36, 32.32) & 95.24 (95.22, 95.25) & 5.61 (5.23, 5.96) & 6.09 (5.67, 6.46) \\ \midrule

\multirow{4}{*}{\begin{tabular}[c]{@{}l@{}}Large Language\\ Models\end{tabular}} 
 & Llama-3.1-8B & 0.01 & 12.38 (11.06, 13.90) & 99.11 (99.09, 99.12) & 11.41 (10.19, 12.81) & 12.38 (11.06, 13.90) \\
 & & 0.05 & 29.07 (27.09, 30.95) & 95.22 (95.21, 95.24) & 5.36 (4.99, 5.71) & 5.81 (5.42, 6.19) \\ \cmidrule(lr){2-7}
 & OpenBioLLM-8B & 0.01 & 10.10 (9.28, 12.03) & 99.08 (99.08, 99.10) & 9.31 (8.56, 11.08) & 10.10 (9.28, 12.02) \\
 & & 0.05 & 27.45 (26.18, 30.19) & 95.21 (95.20, 95.23) & 5.06 (4.83, 5.56) & 5.49 (5.24, 6.04) \\ 

\midrule

\multicolumn{7}{c}{\textbf{Prediction Window = 12 Months}} \\
\midrule
\multirow{6}{*}{Machine Learning} 
 & Elastic Net Logistic Regression & 0.01 & 10.90 (9.76, 11.96) & 99.12 (99.11, 99.13) & 13.00 (11.64, 14.26) & 10.90 (9.76, 11.96) \\
 & & 0.05 & 26.39 (24.78, 28.00) & 95.26 (95.24, 95.28) & 6.30 (5.91, 6.68) & 5.28 (4.96, 5.60) \\ \cmidrule(lr){2-7}
 & Random Forest & 0.01 & 11.29 (10.12, 12.47) & 99.12 (99.11, 99.14) & 13.47 (12.06, 14.87) & 11.30 (10.11, 12.47) \\
 & & 0.05 & 28.12 (26.47, 29.84) & 95.28 (95.26, 95.30) & 6.71 (6.31, 7.12) & 5.62 (5.29, 5.97) \\ \cmidrule(lr){2-7}
 & XGBoost & 0.01 & 11.57 (10.43, 12.78) & 99.13 (99.11, 99.14) & 13.80 (12.44, 15.24) & 11.57 (10.43, 12.78) \\
 & & 0.05 & 27.88 (26.20, 29.57) & 95.28 (95.26, 95.30) & 6.65 (6.25, 7.05) & 5.58 (5.24, 5.91) \\ \midrule
 
\multirow{4}{*}{\begin{tabular}[c]{@{}l@{}}Masked Language\\ Models\end{tabular}} 
 & ModernBERT & 0.01 & 11.10 (10.00, 12.24) & 99.12 (99.11, 99.14) & 13.24 (11.92, 14.59) & 11.10 (10.00, 12.23) \\
 & & 0.05 & 29.69 (27.88, 31.29) & 95.30 (95.28, 95.32) & 7.08 (6.65, 7.46) & 5.94 (5.58, 6.26) \\ \cmidrule(lr){2-7}
 & BioClinical ModernBERT & 0.01 & 11.41 (10.24, 12.59) & 99.13 (99.11, 99.14) & 13.61 (12.20, 15.01) & 11.41 (10.24, 12.58) \\
 & & 0.05 & 29.18 (27.41, 30.78) & 95.29 (95.27, 95.31) & 6.96 (6.54, 7.34) & 5.84 (5.48, 6.16) \\ \midrule

\multirow{4}{*}{\begin{tabular}[c]{@{}l@{}}Large Language\\ Models\end{tabular}} 
 & Llama-3.1-8B & 0.01 & 9.76 (8.71, 10.90) & 99.11 (99.09, 99.12) & 11.65 (10.38, 13.00) & 9.77 (8.70, 10.90) \\
 & & 0.05 & 26.67 (25.14, 28.47) & 95.26 (95.24, 95.28) & 6.36 (6.00, 6.79) & 5.33 (5.03, 5.69) \\ \cmidrule(lr){2-7}
 & OpenBioLLM-8B & 0.01 & 10.63 (10.04, 12.39) & 99.12 (99.11, 99.14) & 12.68 (11.97, 14.77) & 10.63 (10.04, 12.39) \\
 & & 0.05 & 28.00 (26.71, 30.04) & 95.28 (95.26, 95.30) & 6.68 (6.37, 7.16) & 5.60 (5.34, 6.01) \\ 
\bottomrule

\end{tabularx}
\caption{Performance of time-varying models by risk tier (top 1\% and 5\%) and prediction window.}
\label{tab:risk_tier_performance}
\end{table*}

\subsection{Overall Model Performance}
Across prediction windows, PR-AUCs increased with longer prediction horizons, mirroring higher event prevalence. For the 3-month window (prevalence 0.32\%), models achieved PR-AUCs of 1.42-2.39\%. For 6, 9, and 12 months (prevalence rates of 0.63\%, 0.92\%, and 1.19\%, respectively), the ranges were 3.23-4.13\%, 4.03-5.27\%, and 5.01-6.72\%, respectively (Table 1). The best-performing models by window were ModernBERT with time-varying features at 3 months (PR-AUC 2.39\%, 95\% CI 1.80-3.34) and 9 months (5.27\%, 4.68-6.01), and XGBoost with time-varying features at 6 months (4.13\%, 3.48-5.01) and 12 months (6.72\%, 6.06-7.53). Compared with static representations, incorporating clinically informed time-varying features improved PR-AUCs in all but one model-window comparison. In ablation analyses, adding condition-persistence rules on top of the time-varying representation improved PR-AUC for most model-window combinations, with the largest average gains at 6- and 12-month horizons (Supplementary Table \hyperref[supp_table_2]{2}). This supports the added value of clinically informed condition persistence. Because time-varying representations with our condition persistence framework yielded the highest PR-AUCs across models and prediction windows, we focused subsequent analyses on the time-varying versions of each model.

\subsection{Concentration of Risk}
We compared PR-AUC and ROC-AUC across three model classes (machine learning, masked language models, large language models) using both static and time-varying representations. Across prediction windows, time-varying models identified 9.76-14.72\% of all future homelessness cases when screening only the top 1\% of patients by predicted risk, with a specificity of 99.04-99.13\% (Table 2). Expanding the risk tier to the top 5\% increased case capture to 26.39-33.09\% (specificity 95.08-95.30\%). Consistent with increasing outcome prevalence over longer horizons, PPV rose with longer windows and narrower risk tiers. Among the best-performing time-varying models at each horizon (ModernBERT at 3 months and XGBoost at 12 months), PPV at the 1\% tier ranged from 3.93\% (95\% CI, 3.13-4.68) at 3 months to 13.80\% (12.44-15.24) at 12 months, and at the 5\% tier from 1.94\% (1.73-2.15) to 6.65\% (6.25-7.05). The highest PPV observed was 18.71\% (16.36-20.84) for XGBoost at 12 months, when screening the top 0.5\% at a specificity of 99.59\% (Supplementary Table \hyperref[supp_table_3]{3}). This means that, in the observed data, approximately 1 in 5 patients flagged subsequently experienced homelessness within 12 months. Risk enrichment within these tiers was substantial: observed-to-expected (O/E) ratios ranged from 9.76 to 14.72 at the 1\% tier and from 5.28 to 6.62 at the 5\% tier, indicating that flagged groups are roughly 5-15 times more likely to become homeless than the cohort average (Supplementary Table \hyperref[supp_table_3]{3}). Together, these results suggest that small, high-risk tiers (top 1-5\%) can substantially concentrate future homelessness risk while preserving very high specificity, making them promising targets for proactive outreach and prevention efforts.

\subsection{Impact of predictors}
\begin{figure}[H]
    \centering
    \includegraphics[width=0.8\linewidth]{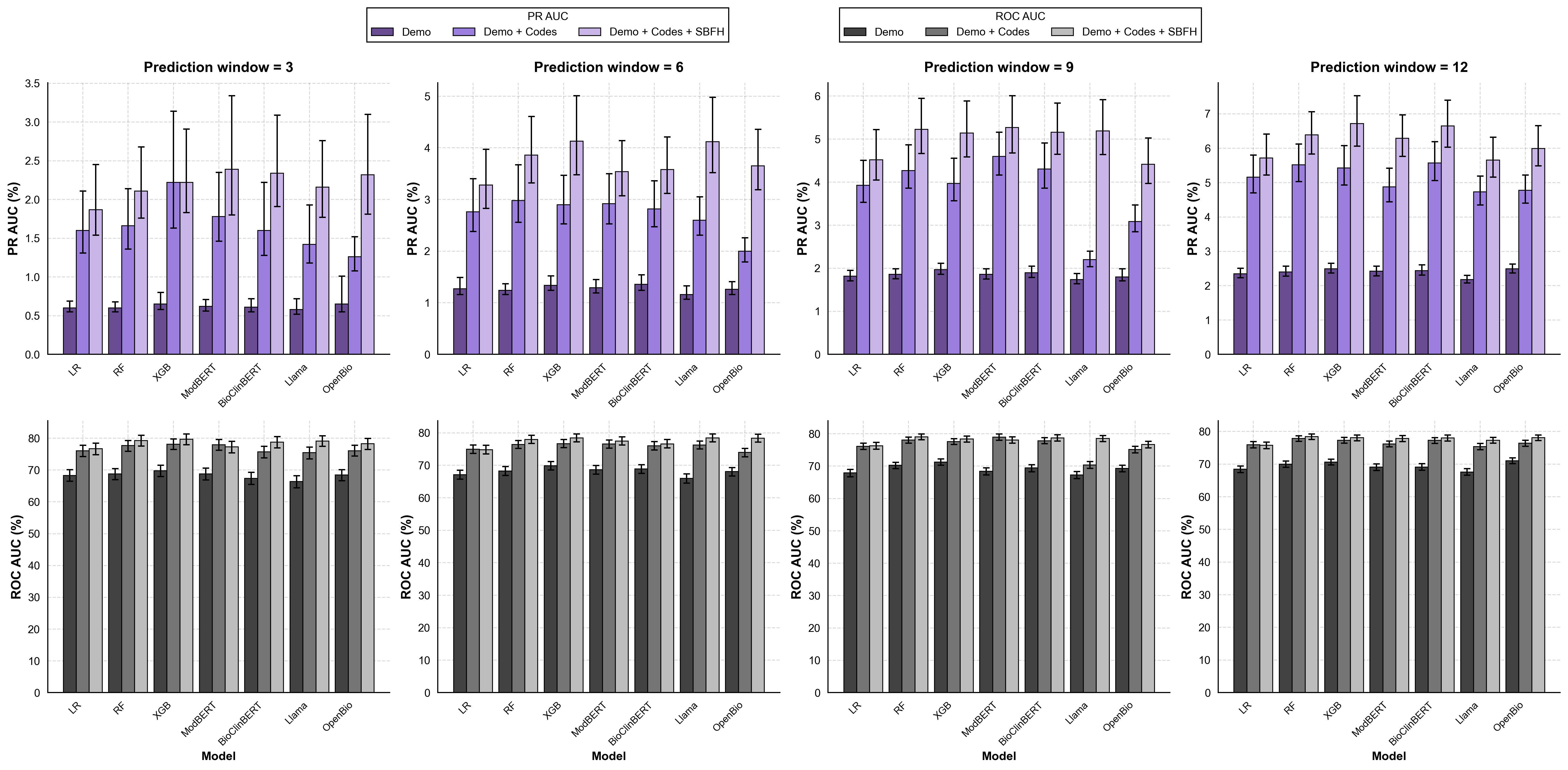}
    \caption{\textbf{Model performance by predictor set and prediction window.}\\
    PR-AUC, area under the precision-recall curve; ROC-AUC, area under the receiver operating characteristic curve; SD, standard deviation; Demo, demographics; Codes, clinical and utilization codes (diagnosis indicators and service-utilization variables derived from ICD-10 diagnostic codes and VA outpatient stop codes, excluding SBFH); SBFH, social and behavioral factors; LR, Elastic Net logistic regression; RF, random forest; XGB, XGBoost; ModernBERT, ModernBERT-large model; BioClinBERT, BioClinical-ModernBERT model; LLaMA, LLaMA-3.1-8B model; OpenBio, OpenBioLLM-8B model.}
    \label{fig:2}
\end{figure}
Performance generally improved as predictor sets expanded from demographics alone to demographics plus clinical and utilization codes (diagnosis indicators and service-utilization variables derived from ICD-10 and VA stop codes, excluding SBFH) and then to demographics, codes, and social and behavioral factors of health (SBFH) across models and prediction horizons (Figure. 2). For example, in Elastic Net Logistic Regression at the 3-month window, adding codes to demographics increased PR-AUC from 0.60\% to 1.60\%, and adding SBFH increased PR-AUC to 1.87\%. At 12 months, PR-AUC increased from 2.35\% with demographics alone to 5.16\% with the addition of codes, and a further jump to 6.67\% with the addition of SBFH. Performance gains from including SBFH were observed across most models and prediction windows and tended to be largest at longer horizons; for instance, for XGBoost, adding SBFH to demographics plus codes increased PR-AUC from 3.97\% to 5.14\% and from 5.43\% to 6.72\% at the 9- and 12-month windows, respectively (Supplementary Table \hyperref[supp_table_4]{4}). These improvements from adding SBFH represent relative gains of 15-30\% for traditional ML models. Overall, Supplementary Table \hyperref[supp_table_4]{4} demonstrates a consistent pattern: adding clinical codes boosts discrimination over demographics alone, and adding SBFH usually yields a further gain, particularly at longer prediction windows.

We also assessed predictor importance using the Kernel SHAP method for the classical machine-learning models (e.g., Elastic Net Logistic Regression, Random Forest, and XGBoost; Supplementary Figure \hyperref[supp_fig_2]{2}). Based on SHAP values, we identified predictors that increased model predictions toward a higher estimated risk of homelessness and those that decreased predictions, which we refer to as positive and negative predictors, respectively. Among the most influential positive predictors, older age (especially 70-79 and $\geq 80$ years), 0\% service-connected disability rating, and selected state-of-residence indicators were consistently selected across models and prediction windows. In contrast, being married, divorced, or separated, higher VISN (Veterans Integrated Service Network) counts, and indicators of emergency or urgent care use were among the most consistent negative predictors. Among diagnosis-related predictors, posttraumatic stress disorder (PTSD), sleep disorders, psychoses, and chronic obstructive pulmonary disease (COPD) frequently appeared as positive predictors. For substance use disorders, cannabis use disorder was a common positive predictor, whereas broader drug abuse diagnoses often had negative SHAP values. Social and behavioral factors (SBFH) showed heterogeneous patterns: government insurance status and documented housing or violence-related problems frequently appeared as positive predictors, particularly at shorter prediction windows, whereas coded employment or financial problems more often appeared with negative SHAP values. Patterns of service utilization characterized by zero mental health, emergency/urgent care, or substance use-related visits were also common negative predictors. These patterns reflect how the models use available codes in combination with other covariates rather than indicating causal risk or protective effects. Accordingly, SHAP values should be interpreted as ranking predictors by their contribution to model predictions for this homelessness-prediction task, not as evidence of causal risk or protective factors.

\subsection{Subgroup Analyses}
\begin{figure}[H]
    \centering
    \includegraphics[width=0.8\linewidth]{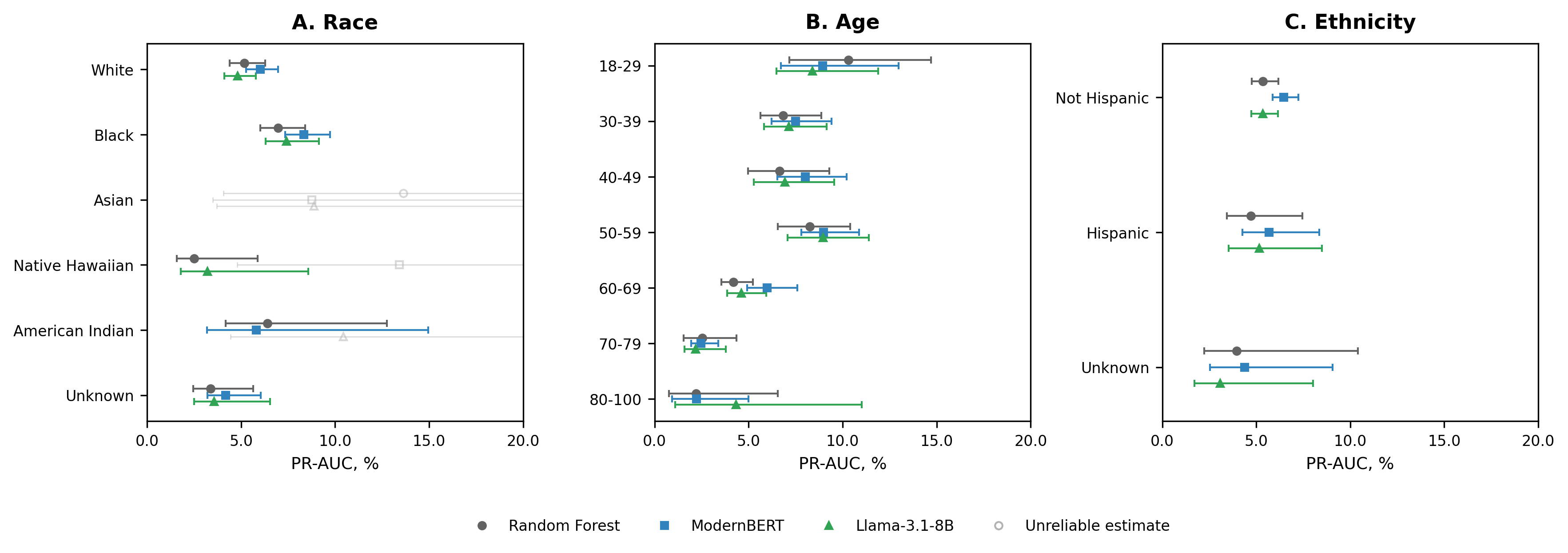}
    \caption{\textbf{Subgroup-specific performance of prediction models for homelessness risk for the 9-month prediction window.}\\
    PR-AUC precision-recall area under the curve, CI confidence interval, RF random forest, ModernBERT ModernBERT-base model, Llama Llama-3.1-8B model. Subgroup estimates labeled “Unreliable” did not meet pre-specified reliability criteria (fewer than 20 positive or 20 negative cases in the subgroup and/or bootstrap CI width > 0.12).}
    \label{fig:placeholder}
\end{figure}
To illustrate subgroup performance, we focused on the 9-month prediction window and three representative models corresponding to the best performers from each model class: Random Forest (ML), ModernBERT (MLM), and Llama-3.1-8B (LLM). Detailed subgroup results for all prediction windows and best-in-class models are reported in Supplementary Table \hyperref[supp_table_5]{5}. Across race groups, estimated discrimination was highest in Black and White veterans (Figure 3, panel A). For ModernBERT, PR-AUCs were 5.34\% (95\% CI, 4.75-6.11\%) in Black veterans and 4.92\% (95\% CI, 4.19-5.88\%) in White veterans, with similar patterns for Llama-3.1-8B and Random Forest. Performance estimates for American Indian, Asian, and Native Hawaiian veterans were lower and more variable; for instance, ModernBERT PR-AUCs ranged from approximately 2.5\% to 6.3\%, with wide confidence intervals extending up to about 15\% reflecting small sample sizes and few events. Likelihood ratio tests (LRTs) indicated significant heterogeneity in race-specific PR-AUCs for ModernBERT and Llama-3.1-8B, with borderline evidence for Random Forest (Supplementary Table \hyperref[supp_table_5]{5}). These patterns suggest modest but non-negligible racial performance differences, with the greatest uncertainty in smaller racial subgroups.

We also observed age-related heterogeneity in discrimination (Figure 3, panel B). PR-AUCs were highest in younger and mid-life adults and lower at the oldest ages. For example, ModernBERT achieved PR-AUCs of 8.92\% (95\% CI, 6.59-12.89\%) in veterans aged 18-29 years and 8.74\% (95\% CI, 7.06-11.09\%) in those aged 50-59 years, compared with 1.78\% (95\% CI, 0.55-6.13\%) in veterans aged 80-100 years; Llama-3.1-8B and Random Forest showed similar gradients, and LRTs rejected the null hypothesis of homogeneous performance across age groups for all three models (Supplementary Table \hyperref[supp_table_5]{5}).

In contrast, we observed only small differences in performance by ethnicity (Figure 3, panel C). For ModernBERT, PR-AUCs were 8.34\% (95\% CI, 3.96-20.12\%) in non-Hispanic veterans and 8.00\% (95\% CI, 4.03-18.94\%) in Hispanic veterans, with comparable estimates for Random Forest and Llama-3.1-8B. Veterans with unknown ethnicity had PR-AUCs around 7.31\% (95\% CI, 6.20-9.10\%), similar in magnitude to the other groups. Although one LRT for ModernBERT reached nominal statistical significance, the absolute PR-AUC gap across ethnic subgroups was small (on the order of 1-3 percentage points; Supplementary Table \hyperref[supp_table_5]{5}).

As a complementary fairness analysis, we examined both the maximum gap in PR-AUC across racial subgroups and the minimum (worst-group) PR-AUC (Supplementary Table \hyperref[supp_table_6]{6}). For the 12-month prediction window and our three primary models (Random Forest, ModernBERT, and Llama-3.1-8B), Llama-3.1-8B showed the smallest race gap (5.58 \%), followed closely by Random Forest (5.92 \%), whereas ModernBERT had the largest gap (9.24 \%). Random Forest achieved the highest worst-group PR-AUC (5.06\%), compared with 4.18\% for ModernBERT and 2.47\% for Llama-3.1-8B. These patterns highlight a trade-off between similarity of performance across race groups (smaller gap) and protection of the worst-performing group (higher worst-group PR-AUC), suggesting that Random Forest offers the strongest worst-group performance, whereas Llama-3.1-8B tends to minimize PR-AUC gaps between race groups. Across age and ethnicity, LLM and MLM models often exhibited gaps of similar or smaller magnitude than Random Forest and comparable worst-group performance, although no single model was uniformly most equitable across all subgroups and prediction windows.

\section{Discussion}

Using a national VA cohort, this study employed several advanced analytic methods to predict first-episode homelessness and revealed five major findings. First, across prediction windows, near-term prediction of first-episode homelessness was consistently more difficult than longer-horizon prediction: all models performed worst at 3 months (prevalence 0.32\%), and PR-AUCs increased as the prediction window and outcome prevalence rose (to 0.63\%, 0.92\%, and 1.19\% at 6, 9, and 12 months, respectively). The best-performing models, time-varying ModernBERT at 3 and 9 months and time-varying XGBoost at 6 and 12 months, achieved PR-AUCs of 2-7\%, which are well above the corresponding outcome prevalences in this extreme class-imbalance setting (for example, a PR-AUC of 2\% at 3 months is roughly six times the base rate). Incorporating time-varying features with clinically informed condition persistence rules improved PR-AUCs in all but one of the head-to-head comparisons with static representations, and ablating the condition persistence rules (while retaining time-varying structure) reduced performance in the majority of model-window combinations. These ablation findings suggest that clinically informed condition persistence rules are an important representational choice that typically improves performance, particularly at longer prediction horizons. Incorporating domain knowledge about the expected course of mental health conditions, physical comorbidities, and social risk yields more informative trajectories than purely naive time-varying representations and is likely to be important for other EHR-based prediction tasks as well. In other words, how well clinicians document changing symptoms and conditions in the data can greatly influence performance of these models. 

 Rather than treating the framework as a way to deal with model under-specification, we viewed it as an explicit, clinically interpretable feature-engineering step that encodes persistence and recurrence patterns not captured by raw visit-level indicators alone. Together, these findings underscore the value of representing how conditions and social risks remain active over time rather than as static ever/never indicators. Discrimination was broadly similar between the best time-varying tabular models and transformer encoders, with ModernBERT outperforming XGBoost at 3- and 9-month horizons and XGBoost performing best at 6 and 12 months, whereas large generative LLMs did not improve on either approach. Notably, gains were more consistently reflected in PR-AUC than ROC-AUC, as expected under extreme class imbalance (in our VA cohort, prevalence was 0.32–1.19\% across 3–12 months). ROC-AUC summarizes global ranking across the full score distribution, whereas PR-AUC is more sensitive to improvements in precision among the highest-risk patients. This divergence supports that homelessness onset can be precipitated by short-horizon destabilizing events and rapid changes in utilization and clinical status (e.g., acute emergency/urgent care use, substance-related encounters, psychiatric crises), where time-varying features capture more directly than static indicators, thereby improving the detection of homelessness even when ROC-AUC changes modestly.

 Second, expanding the predictor set from demographics alone to clinical codes and then to social and behavioral factors (SBFH) produced a consistent stepwise gain in discrimination across model classes and prediction horizons, with the largest absolute improvements in PR-AUC when SBFH were added at longer prediction windows. To our knowledge, no prior work in homelessness risk prediction, particularly among US veterans, has explicitly quantified the incremental value of SBFH beyond demographics and clinical comorbidity, even though prior studies have identified social factors such as low income, unemployment, legal problems, and prior housing instability as key correlates of homelessness risk\cite{o2016tailoring, montgomery2020demographic, tsai2015risk}. Our findings extend this literature by showing that routinely collected SBFH can yield consistent gains in PR-AUC, sensitivity, and PPV when incorporated into EHR-based risk models, supporting ongoing efforts to systematically capture social risk data in VA and other healthcare systems. At the same time, in some cases, effect sizes were small and confidence intervals overlapped in some settings, likely reflecting extreme outcome rarity and incomplete capture of social risks in structured EHR fields, underscoring the need for richer SBFH ascertainment (for example, via NLP and LLM-based extraction from clinical text) to fully realize the potential of social data for homelessness prevention\cite{yang2025predicting}. As various screeners, like the Assessing Circumstance and Offering Resources for Needs (ACORN) used in the VA\cite{russell2023implementing} or the various SBFH screeners that many states are mandating or incentivizing managed care organizations to use\cite{davidson2019screening} are implemented, further research is needed on determining which are most effective in capturing the important factors. 

Third, using Kernel SHAP, an additive feature-attribution method based on Shapley values, we ranked predictors across models and prediction windows. The most influential positive predictors were demographic and clinical markers of vulnerability, older age, lower income, and 0\% service-connected disability; serious mental illnesses, including PTSD; sleep disorders; cardiometabolic and pulmonary comorbidities; and substance use disorders, together with SBFH features such as housing or violence-related problems, neighborhood deprivation (ADI), and non-specific psychosocial needs. These patterns closely align with prior evidence that substance use disorders, mental illness, and income-related disadvantages are among the strongest and most consistent risk factors for homelessness among US veterans\cite{tsai2015risk, washington2010risk, tsai2019homelessness}. In contrast, marital status, more regular primary care or mental health contact, and the absence of recent emergency/urgent-care or substance-use visits frequently appeared as negative predictors, consistent with studies showing that veterans experiencing homelessness disproportionately rely on emergency departments and that engagement in ongoing outpatient care may reduce acute utilization\cite{o2018population, gundlapalli2017characteristics, tsai2013risk}. Some SBFH features with known adverse associations, such as employment/financial and legal problems, were negative predictors in certain windows, likely reflecting multivariable interactions, selection into intensive VA or legal-aid services, and documentation patterns rather than true protective effects. Taken together, these findings underscore that SHAP values highlight variables that are most useful for discrimination within our multivariable models, not necessarily causal risk or protective factors, and should be interpreted in conjunction with substantive knowledge of veteran homelessness and social factors of health.

Fourth, we evaluated model fairness by comparing discrimination across racial, age, and ethnic subgroups using PR-AUC, likelihood ratio tests for subgroup heterogeneity (with false discovery rate-adjusted P values), and worst-case (“minimum subgroup”) performance. Consistent with prior work on subgroup performance in clinical risk scores, we observed only modest race-related differences between the largest groups, Black and White veterans, with overlapping confidence intervals and similar PR-AUCs across models, whereas estimates for American Indian, Asian, and Native Hawaiian veterans were more variable and often unreliable because of very small sample sizes. Age-related heterogeneity was more pronounced: all three models performed best in mid-life adults and worse at the extremes of age, a pattern aligned with reports that one-size-fits-all models often underperform in underrepresented or clinically distinct subpopulations \cite{chen2023algorithmic}. In contrast, we found no statistically significant differences in PR-AUC by ethnicity, aside from lower and more uncertain estimates among veterans with unknown ethnicity, which may reflect differential documentation rather than true performance differences. Subgroup analyses also revealed that fairness is multidimensional: some models narrowed gaps in PR-AUC between race groups but offered lower performance for the worst-performing group, whereas others improved worst-group performance at the cost of larger gaps. Across race, age, and ethnicity, no single model was uniformly most equitable across all metrics and horizons, implying that model selection will necessarily depend on how health systems prioritize different fairness objectives, for example, minimizing disparities between groups versus maximizing protection for the most disadvantaged subgroup, rather than on overall model performance. Importantly, studies that have compared demographic groups on their rates of homelessness and utilization of VA homeless services suggest no clear disparities by race or sex\cite{montgomery2020housing}. But prediction models should be combined with clinician judgment and receive periodic evaluations to avoid any disparate impacts. 

Fifth, risk of future homelessness was highly concentrated within small high-risk tiers, such that a substantial share of events occurred among veterans in the top 1-5\% of predicted risk, implying that targeted outreach to a relatively small segment of the population could meaningfully increase the efficiency of prevention efforts. This has also been reported in previous studies\cite{elbogen2025identifying, tsai2024predicting} and underscore the value of having a strong prediction model that can be used to guide efficient use of outreach resources. Although understanding risk factors and identifying current homelessness has been an active area of research, our work extends prior studies by predicting future first-episode homelessness, explicitly modeling temporality, and quantifying the added value of SBFH across diverse modeling approaches in a national VA cohort. Because homelessness is a rare outcome, even well-calibrated models inevitably yield modest PPVs, and translating risk scores into practice raises challenges around logistics, generalizability, cost-effectiveness, and the risk-benefit balance of preventive interventions. From a programmatic perspective, these risk tiers could therefore be used to prioritize more intensive assessment and services, while keeping the overall screening burden manageable.

At the same time, low PPV and high false-positive rates carry important ethical, psychological, and resource implications. Labeling someone as “high risk for homelessness” may contribute to distress, mistrust, or stigma, and misdirect scarce housing and social-service resources if not paired with careful, person-centered assessment. These concerns underscore that homelessness risk scores should function as decision-support tools rather than standalone gatekeepers, embedded within transparent workflows that emphasize voluntary engagement, shared decision making, and clear benefit to veterans. Future work should include decision-analytic evaluation such as decision curve analysis and net-benefit frameworks to quantify whether models with modest PPV yield a favorable trade-off between correctly identified high-risk cases and unnecessary outreach under explicit assumptions about intervention burden, cost, and benefit. Prior evaluations of VA and other health-system risk models (for example, for suicide risk and high-cost utilization) suggest that, under reasonable assumptions about intervention burden and benefit, prediction tools can deliver positive net benefit even when PPV remains in the single or low double digits\cite{harris2025evaluating, kessler2023evaluation, vickers2016net}. Similar impact evaluations co-designed with veterans, clinicians, and housing partners will be essential to ensure that homelessness risk prediction is implemented in ways that are equitable, clinically useful, and aligned with prevention goals.

 Our study has several limitations. First, the VA population’s demographic composition and patterns of care differ from those of the overall US population, and VHA data do not fully capture community-based hospitalizations, which may limit generalizability to non-VA settings. Second, temporal generalizability is uncertain: our models were developed using pre-pandemic data, and secular changes such as the COVID-19 pandemic and recent transitions in VHA EHR systems may alter both homelessness risk and its documentation. Future work should evaluate similar approaches in post-pandemic cohorts to assess robustness over time. Third, we restricted our analysis to a one-year observation window before prediction; longer observation periods or alternative look-back windows could further improve performance and warrant exploration. Fourth, homelessness outcomes and SBFH were ascertained from structured EHR fields, which likely undercapture housing instability and social risks and may lead to outcome and predictor misclassification. Finally, we evaluated only a limited set of temporal granularities and condition persistence rules; although these rules improved discrimination in most settings, they remain approximations and may over-smooth some conditions. Refining and externally validating persistence choices is an important area for future work. 

 To our knowledge, this is one of the first large-scale national studies of predictive modeling for first-episode homelessness among US veterans using EHR data. This study identified five key findings regarding the prediction of first-episode homelessness. First, homelessness was rare and especially difficult to predict over short horizons, although the best time-varying models still clearly outperformed the base rate. Second, clinically informed longitudinal representations generally improved or matched performance relative to static or naive time-varying features. Third, expanding predictors from demographics to clinical diagnoses and routinely collected social and behavioral factors yielded consistent gains and, together with SHAP analyses, highlighted clinically and socially plausible patterns of vulnerability and engagement in care. Fourth, the risk of future homelessness was highly concentrated in small high-risk tiers, suggesting that targeted outreach to a small share of veterans could capture a substantial fraction of future events. Fifth, discrimination was broadly similar across major demographic subgroups, but the heterogeneity we observed by race and age underscores that risk scores should be deployed as equity-aware tools paired with explicit fairness objectives and ongoing monitoring rather than as one-size-fits-all solutions. Together, these findings suggest that EHR-based risk models that incorporate longitudinal trajectories and social risk information may help health systems support more targeted, prevention-oriented responses to homelessness among veterans.

\section{Methods}

\subsection{Data Source and Study Design}

We conducted a retrospective prognostic modeling study using inpatient and outpatient records from the U.S. Department of Veterans Affairs (VA) Corporate Data Warehouse (CDW). The CDW contains comprehensive clinical and administrative data from the Veterans Health Administration (VHA), the largest integrated healthcare system in the U.S., encompassing over 1,200 medical centers and clinics nationwide. Available information includes demographics, diagnoses, procedures, medications, clinical notes, and health services utilization, making it a robust resource for large-scale health research.

The study protocol was approved by the VA Bedford Healthcare System Institutional Review Board (IRB \#1652850-23), with a waiver of informed consent due to minimal risk. We adhered to the Transparent Reporting of a Multivariable Prediction Model for Individual Prognosis or Diagnosis (TRIPOD) guidelines, and the study was conducted in accordance with the principles of the Declaration of Helsinki.

The unit of analysis was the individual VA patient, with each patient contributing a single observation based on data aggregated over a 1-year observation window. We defined the observation window as January 1-December 31, 2016, and the index date as January 1, 2017. We predicted the first episode of homelessness over four cumulative risk windows: 3 months (Q1 2017), 6 months (Q1-Q2 2017), 9 months (Q1-Q3 2017), and 12 months (Q1-Q4 2017). We selected the 2016-2017 period to ensure sufficient follow-up time while utilizing contemporary data that predates COVID-19-related disruptions and follows the transition to ICD-10 coding.

Homelessness was defined using a composite measure used by the VA Homeless Programs Office and defined in previous work\cite{tsai2022developing}. Briefly, patients were classified as experiencing homelessness if they met any of the following criteria in CDW during the prediction window: (1) International Classification of Diseases, 10th Revision (ICD-10) diagnostic code Z59.0× in inpatient or outpatient records; (2) a positive response on part 2 of the annual Homeless Screening Clinical Reminder (HSCR); (3) a homelessness-related record in the Homeless Operations, Management, and Evaluation System (HOMES) (e.g., participation in programs such as Grant and Per Diem [GPD], Health Care for Homeless Veterans [HCHV]/Health Care for Re-Entry Veterans [HCMI], Domiciliary Care for Homeless Veterans [DCHV], or Compensated Work Therapy/Transitional Residence [CWT/TR]); or (4) relevant outpatient stop codes (504, 508, 511, 528, 529, 555, 556, 590) or inpatient specialty codes (28, 29, 37, 39) indicating receipt of homelessness-specific services. VA stop codes designate the type of care delivered during outpatient visits and the associated workload.

Eligible participants were U.S. veterans nationwide who met the following criteria: (1) aged 18-100 years; (2) had at least one VA health care encounter during the observation window. Patients were excluded if they had (1) any homelessness indicators in the five years before or during the observation window, or (2) conflicting demographic information. The cohort flow diagram is provided in Supplementary Figure \hyperref[supp_fig_1]{1}. 

\subsection{Predictor Construction}

\begin{table*}[!htbp]
\centering
\small
\renewcommand{\arraystretch}{1.2}

\begin{tabularx}{\textwidth}{>{\raggedright\arraybackslash}p{4cm} >{\raggedright\arraybackslash}X}
\toprule
\textbf{Predictor Type} & \textbf{Predictors} \\
\midrule
Demographics & Age, Body Mass Index (BMI), Combat Exposure, Education, Ethnicity, Gender, Income, Marital Status, Military Sexual Trauma (MST), Race, Rural/Urban, State, Service-connected disability, Veterans Integrated Service Network (VISN) \\
\midrule
Service Utilization & Outpatient Visits: Primary care, Mental health, Substance abuse, Speciality care, Rehabilitation, Diagnostic / Ancillary, \newline Inpatient Visits: Total Visits, Visit Days, \newline Emergency / Urgent-care Visits, \newline Veterans Integrated Service Network (VISN) count \\
\midrule
Mental Health Disorders & Anxiety Disorder, Bipolar Disorder, Dementia, Depression, Other Neurological Disorders, Posttraumatic Stress Disorder, Psychoses, Sleep Disorder \\
\midrule
Physical Health Disorders & Blood Loss Anemia, Cardiac Arrhythmia, Cardiovascular Disease, Chronic Pulmonary Disease, Cirrhosis, Coagulopathy, Congestive Heart Failure, Deficiency Anemia, Diabetes, Fluid and Electrolyte Disorders, Hepatitis, HIV, Hypertension, Hypothyroidism, Influenza, Liver Disease, Lymphoma, Metastatic Cancer, Obesity, Pain, Paralysis, Peptic Ulcer Disease, Peripheral Vascular Disorders, Pulmonary Circulation Disorders, Renal Failure, Rheumatoid Arthritis/collagen, Solid Tumor without Metastasis, Traumatic brain injury, Valvular Disease, Weight Loss \\
\midrule
Substance Abuse Disorders & Alcohol use disorder, Cannabis, Cocaine, Drug Abuse, Hallucinogen, Nicotine dependence, Opioid use disorder, Other stimulant \\
\midrule
Social and Behavioral Factors of Health & Area Deprivation Index (ADI), Employment or financial problems, Food insecurity, Housing problems, Insurance, Legal problems, Non-specific psychosocial needs, Social or familial problems, Violence problems \\
\bottomrule
\end{tabularx}

\caption{List of predictors considered in our study.}
\label{tab:predictors}
\end{table*}

Predictors were organized into six domains: demographics, service utilization, mental health disorders, physical health disorders, substance use disorders, and social and behavioral factors of health (SBFH). Our selection of variables was informed by epidemiological studies that have identified risk factors of homelessness among both Veterans and non-veterans\cite{tsai2015risk, nilsson2019individual}. All features were extracted during the observation window (January 1-December 31, 2016), except military sexual trauma (MST) and combat exposure, which were derived from any historical record before the index date. 

Patient-level demographic variables included age, body mass index (BMI), sex, gender, race, ethnicity, marital status, education, income, state of residence, and rural/urban status. Three indicators specific to military service were also included: combat exposure, MST, and service-connected disability status. Service use variables captured categorized counts of inpatient admissions, outpatient visits, emergency/urgent care visits, and unique Veterans Integrated Service Network (VISN) encounters: counts of unique VISNs visited in the observation window, which we included in all models to partially account for differences in healthcare utilization intensity that could otherwise confound associations between persistence-filled diagnoses and homelessness risk. VISNs are regional administrative networks that coordinate care across VA facilities. Major diagnoses were identified from ICD-10 codes, which were aggregated into clinically meaningful categories capturing mental health, physical health, and substance abuse disorders.

Structured ICD-10 and VA stop codes captured seven SBFH factors: employment or financial problems, food insecurity, housing instability, legal problems, non-specific psychosocial needs, social/familial problems, and exposure to violence. Neighborhood-level socioeconomic status was represented using the national Area Deprivation Index (ADI), linked via patient ZIP code. We assigned the maximum ADI observed during the observation window, based on the rationale that exposure to the most socioeconomically deprived environment may have the largest and most enduring impact on risk for homelessness. Insurance information was also added from the EHR database. 

In total, we used 79 predictors (Table 3): 14 demographic variables, 10 service utilization variables, 8 indicators of mental health disorders, 30 indicators of physical health disorders, 8 indicators of substance use disorders, and 9 SBFH. Collectively, mental health, physical health, and substance use disorder indicators constituted the diagnostic feature set. All diagnostic and SBFH variables were binary, while demographic and service utilization features were categorical (except combat exposure and MST, which were binary). Missing values on demographic variables were coded as “unknown”. For diagnostic and SBFH variables, absence of a code during a period was treated as no recorded evidence of that condition. Full variable-level definitions are provided in Supplementary Note \hyperref[supp_note_1]{1}.

Diagnostic and SBFH features were modeled using two complementary representations to examine the impact of temporal dynamics on prediction performance. The first, static, representation used binary indicators for conditions observed at any point during the 2016 observation window. The second, time-varying, representation aggregated each feature across multiple temporal periods and applied clinically informed condition persistence rules that specify how long diagnoses and social/behavioral problems remain active after they are last recorded (only within the 2016 observation window). We examined two temporal granularities, quarterly (Q1-Q4 2016) and half-year (H1-H2 2016), which was treated as a hyperparameter and selected using validation PR-AUC within each model and prediction horizon; selected values are reported in Supplementary Table \hyperref[supp_table_8]{8}. We selected half-year and quarterly aggregation because these intervals capture clinically meaningful changes in conditions and service use over time while avoiding the extreme sparsity and noise that would arise with finer-grained (e.g., monthly/weekly) windows. 

Time-varying features were assigned one of four condition persistence rules based on their clinical course and likelihood of recurrence: (1) chronic persistent: once observed, the feature remains active for all subsequent intervals, reflecting long-term or lifelong conditions; (2) recurrent time-limited: the feature remains active for T intervals after observation, capturing conditions that may remit or recur with treatment; (3) ever-history: the feature is marked active in all periods once recorded, representing diagnoses with enduring prognostic relevance regardless of current status; and (4) episodic: the feature is active only in the period in which it was recorded, suitable for acute or self-limited conditions. These persistence rules were specified a priori in consultation with VA clinicians and were intended as pragmatic approximations of underlying clinical courses rather than data-driven optimizations. For recurrent time-limited conditions, we used a default timeout of two quarters ($\approx$ 6 months) unless clinicians recommended a shorter horizon. The assignment of persistence modes and timeouts (T) for each predictor is provided in Supplementary Table \hyperref[supp_table_9]{9} (Fill strategy). To guard against over-correction, we additionally evaluated models using time-varying features without persistence and compared performance gains attributable to the framework in ablation analyses (Supplementary Table \hyperref[supp_table_2]{2}).

Persistence timeouts (\textit{T}) were defined in quarterly periods and converted proportionally for half-year aggregation (e.g., a two-quarter timeout equals one half-year period), with timeouts rounded to whole half-year intervals. This condition persistence framework was developed through iterative consultation with VA clinicians specializing in homelessness prevention, primary care, and psychiatry. Because the observation window covered only one year, T values necessarily spanned relatively short horizons and were designed to approximate, rather than precisely model, long-term temporal patterns, providing a structured way to evaluate whether incorporating clinically informed condition persistence improves predictive performance. An illustrative example of the data representation and model inputs is shown in Supplementary Note \hyperref[supp_note_3]{3}. 

\subsection{Statistical Analyses}

We evaluated predictive performance using three classes of models: classical machine learning (ML), masked language models (MLMs), and large language models (LLMs). Classical ML models included Elastic Net logistic regression, Random Forest, and XGBoost, with hyperparameters selected via grid search on validation data. For MLMs, ModernBERT and BioClinical-ModernBERT were fine-tuned for binary sequence classification using a standard classification head. LLMs included LLaMA-3.1-8B and OpenBioLLM-8B, both fine-tuned using Low-Rank Adaptation (LoRA)\cite{hu2022lora} with a modified language-modeling head. We replaced the standard vocabulary prediction layer with a 2-class classification head that outputs “Yes” (homeless) or “No” (not homeless). LoRA fine-tunes only a small subset (<1\%) of parameters while keeping the base model otherwise frozen, reducing computational requirements. During training, only the final predicted token contributed to the loss (answer-only training), focusing optimization on the classification task. Proprietary models such as GPT-4 were excluded because VHA privacy restrictions prohibit data transfer outside the VINCI environment. Full model architectures and hyperparameters are provided in Supplementary Note \hyperref[supp_note_2]{2}.

Tabular models used categorical demographic and utilization variables and binary diagnosis and social and behavioral factors of health (SBFH) features, represented either statically or aggregated across quarterly/half-year intervals using the time-varying representation described above. MLMs and LLMs received an equivalent text rendering that grouped features by clinical domain and time (Q1-Q4 or H1-H2) and produced binary Yes/No outputs, aligning information content across modeling approaches.

Predictive performance was assessed on the held-out test set using the area under the receiver operating characteristic curve (ROC-AUC), area under the precision-recall curve (PR-AUC), sensitivity, specificity, positive predictive value (PPV), and observed-to-expected (O/E) ratios. The O/E ratio represents the observed incidence in a risk group divided by the expected incidence if the same number of patients were selected at random from the population. Because homelessness is a rare event, we calculated each metric for multiple risk-group sizes. For a given model, a risk-group size P corresponds to the top P fraction of patients by predicted risk. Following prior work and our cohort characteristics, we considered risk groups corresponding to the top 0.5\%, 1\%, 5\%, 10\%, 25\%, 50\%, and 75\% of patients by predicted risk.

Models were evaluated under varying predictor and temporal representations to assess their contributions to performance. We compared three predictor configurations: (1) demographics only; (2) demographics plus clinical and utilization codes (diagnosis indicators and service‐utilization variables derived from ICD-10 diagnostic codes and VA outpatient stop codes); and (3) demographics plus clinical/utilization codes and SBFH, to quantify the incremental value of each predictor class.

To evaluate generalizability and fairness, we conducted subgroup analyses stratified by race, age, and ethnicity for the 9-month prediction window. We pre-specified reliability criteria requiring at least 20 positive and 20 negative cases in the subgroup and a bootstrap CI width $\leq $0.12; estimates not meeting these criteria were flagged as unreliable in figures and supplementary tables. To formally assess heterogeneity in performance across subgroups, we used likelihood-ratio tests comparing models with and without subgroup-specific performance parameters and controlled for multiple testing using the Benjamini-Hochberg false-discovery rate procedure, reporting FDR-adjusted q values. We additionally summarized fairness trade-offs using the maximum gap in PR-AUC across racial subgroups and the minimum (worst-group) PR-AUC.

Training data were split at the patient level into 92\% training, 3\% validation, and 5\% test sets to prevent data leakage. To address severe class imbalance while preserving realistic evaluation, we applied stratified downsampling only to the training set (matched on gender, age group, and race) to balance outcome classes; validation and test sets retained the original prevalence distribution. Given the large cohort size (n = 4,276,403) and event prevalence across prediction windows (0.32-1.19\%), the training set contained tens of thousands of positive cases, and the validation and test sets contained several hundred to a few thousand positives each, sufficient for hyperparameter tuning and independent evaluation. This allocation maximized training data for model optimization while maintaining adequately powered subsets for unbiased performance assessment (Supplementary Table \hyperref[supp_table_7]{7}). Random seeds were fixed to ensure reproducibility, and all models were trained, validated, and tested on identical splits to ensure comparability.

Stratified bootstrapping with 2,000 iterations was used to estimate 95\% confidence intervals (CIs) for all metrics. In each iteration, the test set was resampled with replacement while preserving the original prevalence of homelessness. The 95\% CI was defined as the 2.5th and 97.5th percentiles of the bootstrapped metric distribution. For each model, the hyperparameter configuration yielding the highest PR-AUC on validation data was selected, prioritizing performance on this rare-event outcome. 

\section{Data Availability}

The VHA EHR data used in this study are available under restricted access because of veterans’ privacy and data security regulations. Access may be obtained with appropriate approvals through the VA Informatics and Computing Infrastructure (VINCI; contact: VINCI@va.gov). Individuals who wish to use these data for research purposes must meet the research credentialing requirements outlined by the VA Office of Research and Development.

\section{Code Availability}

Code for the prediction models is available from the corresponding author upon reasonable request. Relevant software tools and libraries include: PEFT (for Low-Rank Adaptation fine-tuning)\cite{PEFT} , LLaMA-3.1-8B\cite{2024MetallamaLlama318BHugging}, OpenBioLLM-8B\cite{AadityaLlama3OpenBioLLM8BHugging}, ModernBERT\cite{warner2025smarter}, BioClinical-ModernBERT\cite{sounack2025bioclinical}, scikit-learn\cite{pedregosa2011scikit}, XGBoost\cite{chen2016xgboost}, PyTorch\cite{paszke2019pytorch}, and HuggingFace Transformers\cite{Transformers}. All natural language prompts used for MLM and LLM model training are provided in the Supplementary Material (Supplementary Note \hyperref[supp_note_2]{2}). Analyses were conducted using Python 3.10.

\section{Acknowledgements}

Research reported in this study was supported by the National Center on Homelessness Among Veterans (NCHAV) and by the National Institutes of Health (NIH) under award number 1R01NR020868. This study was also in part supported by NIH under award numbers R01DA056470-A1 and 1R01AG080670-01, and by the U.S. Department of Veterans Affairs (VA) Health Systems Research. We thank Joel Reisman for guidance on data access and interpretation and for helpful discussions. The funder played no role in study design, data collection, analysis and interpretation of data, or the writing of this manuscript.

\section{Author contributions}

Rohan Pandey, Hong Yu, and Jack Tsai conceived and designed the study. Rohan Pandey and Haijuan Yan performed data collection. Rohan Pandey implemented the code, conducted experiments, and performed analyses. Rohan Pandey drafted the manuscript. Hong Yu and Jack Tsai contributed to manuscript revision and provided critical feedback. Hong Yu and Jack Tsai supervised the study. All authors contributed to interpretation of results, approved the final manuscript, and take responsibility for the submission.

\section{Competing interests 
}
The authors declare no competing interests.

\bibliography{sample}

\clearpage
\input{appendix}

\end{document}

%% file: appendix.tex
\section*{Supplementary Figure 1. Cohort flow diagram}\phantomsection
\label{supp_fig_1}
a) Construction of the cohort and b) the study timeline; M = \{3,6,9,12\} (in months)

\begin{figure}[!htbp]
    \centering
    \includegraphics[width=\linewidth]{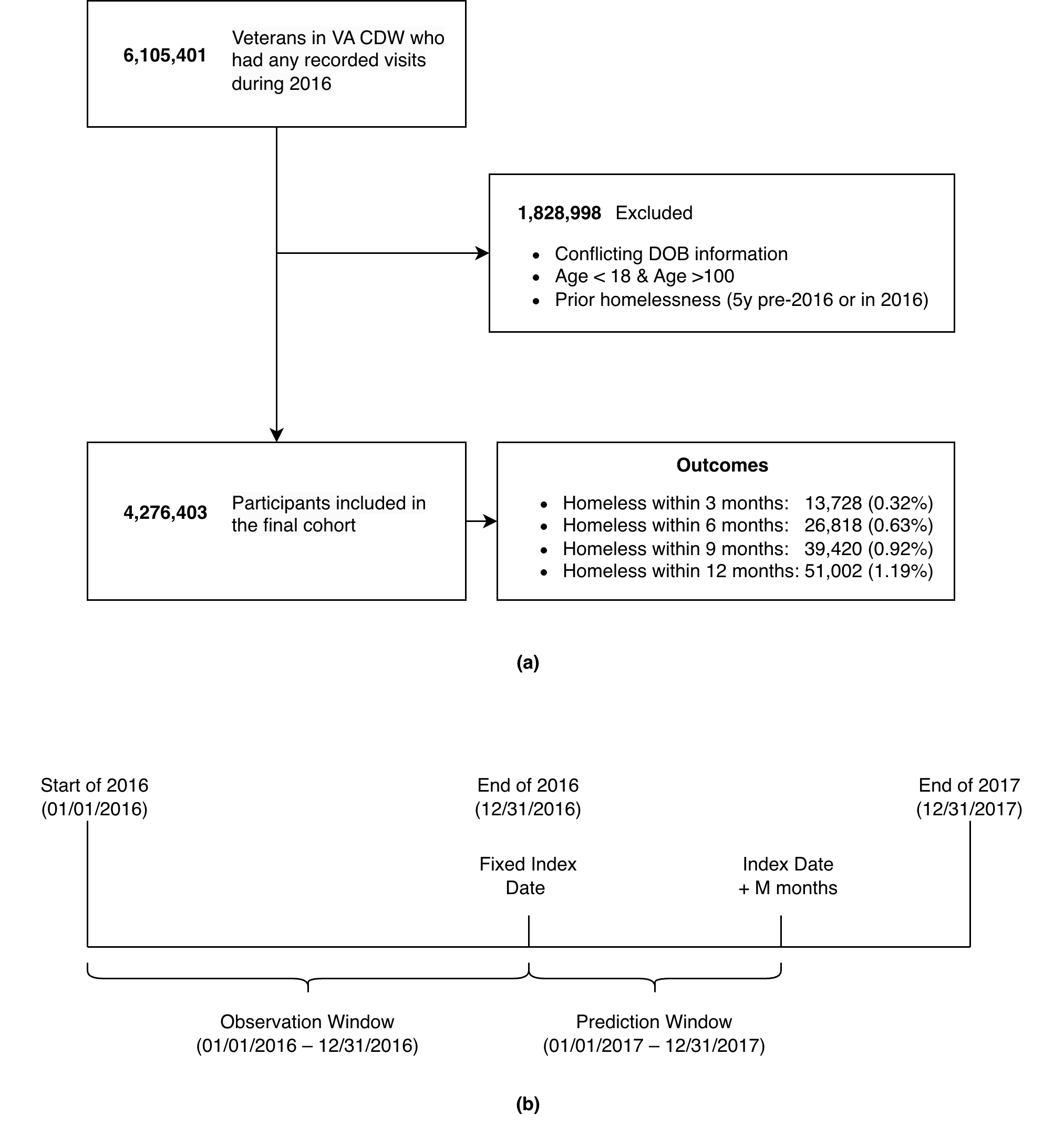}
\end{figure}

\clearpage
\section*{Supplementary Table 1. Cohort characteristics}\phantomsection
\label{supp_table_1}

{
\scriptsize
\renewcommand{\arraystretch}{0.9}

\begin{center}
\begin{longtable}{>{\raggedright\arraybackslash}p{1.5cm} >{\raggedright\arraybackslash}p{2.0cm} >{\raggedright\arraybackslash}p{1.4cm} >{\raggedright\arraybackslash}p{2.0cm} >{\raggedright\arraybackslash}p{2.0cm} >{\raggedright\arraybackslash}p{2.0cm} >{\raggedright\arraybackslash}p{2.0cm} >{\raggedright\arraybackslash}p{2.0cm}}
\toprule
\textbf{Predictor Type} & \textbf{Predictor} & \textbf{Level} &
\makecell[l]{\textbf{Total}\\\textbf{(n = 4,276,403)}} &
\makecell[l]{\textbf{3 Months Homeless}\\\textbf{(n = 13,728; 0.32\%)}} &
\makecell[l]{\textbf{6 Months Homeless}\\\textbf{(n = 26,818; 0.63\%)}} &
\makecell[l]{\textbf{9 Months Homeless}\\\textbf{(n = 39,420; 0.92\%)}} &
\makecell[l]{\textbf{12 Months Homeless}\\\textbf{(n = 51,002; 1.19\%)}} \\
\midrule
\endfirsthead

\midrule
\multicolumn{8}{c}{{Supplementary Table 1 -- continued}} \\
\midrule
\endhead

\midrule
\multicolumn{8}{r}{{Continued on next page}} \\
\endfoot

\bottomrule
\endlastfoot
\multirow{135}{*}{Demographics} & \multirow{7}{=}{Age} & 18-29 & 142204 (3.33\%) & 1146 (8.35\%) & 2225 (8.3\%) & 3260 (8.27\%) & 4183 (8.2\%) \\
 &  & 30-39 & 339880 (7.95\%) & 2301 (16.76\%) & 4468 (16.66\%) & 6443 (16.34\%) & 8161 (16.0\%) \\
 &  & 40-49 & 372185 (8.7\%) & 1860 (13.55\%) & 3574 (13.33\%) & 5201 (13.19\%) & 6737 (13.21\%) \\
 &  & 50-59 & 580750 (13.58\%) & 3192 (23.25\%) & 6113 (22.79\%) & 8961 (22.73\%) & 11591 (22.73\%) \\
 &  & 60-69 & 1280756 (29.95\%) & 3459 (25.2\%) & 6836 (25.49\%) & 10207 (25.89\%) & 13325 (26.13\%) \\
 &  & 70-79 & 921838 (21.56\%) & 1251 (9.11\%) & 2525 (9.42\%) & 3778 (9.58\%) & 4996 (9.8\%) \\
 &  & 80-100 & 638790 (14.94\%) & 519 (3.78\%) & 1077 (4.02\%) & 1570 (3.98\%) & 2009 (3.94\%) \\
\cmidrule(l){2-8}
 & \multirow{5}{=}{Body Mass Index (BMI)} & Unknown & 1187477 (27.77\%) & 3757 (27.37\%) & 7296 (27.21\%) & 10591 (26.87\%) & 13617 (26.7\%) \\
 &  & Obese & 1412355 (33.03\%) & 4123 (30.03\%) & 8163 (30.44\%) & 12154 (30.83\%) & 15867 (31.11\%) \\
 &  & Overweight & 1101176 (25.75\%) & 3319 (24.18\%) & 6574 (24.51\%) & 9679 (24.55\%) & 12516 (24.54\%) \\
 &  & Normal & 547678 (12.81\%) & 2388 (17.4\%) & 4515 (16.84\%) & 6603 (16.75\%) & 8493 (16.65\%) \\
 &  & Underweight & 27717 (0.65\%) & 141 (1.03\%) & 270 (1.01\%) & 393 (1.0\%) & 509 (1.0\%) \\
\cmidrule(l){2-8}
 & \multirow{1}{=}{Combat Exposure} &  & 102679 (2.4\%) & 556 (4.05\%) & 1067 (3.98\%) & 1590 (4.03\%) & 2003 (3.93\%) \\
\cmidrule(l){2-8}
 & \multirow{5}{=}{Education} & Completed High School & 1636116 (38.26\%) & 4709 (34.3\%) & 9412 (35.1\%) & 13914 (35.3\%) & 17911 (35.12\%) \\
 &  & Unknown & 1156717 (27.05\%) & 5708 (41.58\%) & 10996 (41.0\%) & 16091 (40.82\%) & 20838 (40.86\%) \\
 &  & Completed College & 1073517 (25.1\%) & 2527 (18.41\%) & 4867 (18.15\%) & 7199 (18.26\%) & 9361 (18.35\%) \\
 &  & Completed Graduate School & 377028 (8.82\%) & 674 (4.91\%) & 1333 (4.97\%) & 1910 (4.85\%) & 2488 (4.88\%) \\
 &  & Attended Vocational School & 33025 (0.77\%) & 110 (0.8\%) & 210 (0.78\%) & 306 (0.78\%) & 404 (0.79\%) \\
\cmidrule(l){2-8}
 & \multirow{3}{=}{Ethnicity} & Not Hispanic & 3887177 (90.9\%) & 12416 (90.44\%) & 24162 (90.1\%) & 35581 (90.26\%) & 45917 (90.03\%) \\
 &  & Hispanic & 266998 (6.24\%) & 1010 (7.36\%) & 2059 (7.68\%) & 3023 (7.67\%) & 4059 (7.96\%) \\
 &  & Unknown & 122228 (2.86\%) & 302 (2.2\%) & 597 (2.23\%) & 816 (2.07\%) & 1026 (2.01\%) \\
\cmidrule(l){2-8}
 & \multirow{3}{=}{Gender} & Male & 3936231 (92.05\%) & 12001 (87.42\%) & 23444 (87.42\%) & 34380 (87.21\%) & 44443 (87.14\%) \\
 &  & Female & 324152 (7.58\%) & 1635 (11.91\%) & 3215 (11.99\%) & 4829 (12.25\%) & 6296 (12.34\%) \\
 &  & Unknown & 16020 (0.37\%) & 92 (0.67\%) & 159 (0.59\%) & 211 (0.54\%) & 263 (0.52\%) \\
\cmidrule(l){2-8}
 & \multirow{14}{=}{Income} & 7,500 & 249116 (5.83\%) & 932 (6.79\%) & 1806 (6.73\%) & 2670 (6.77\%) & 3403 (6.67\%) \\
 &  & 17,500 & 190481 (4.45\%) & 613 (4.47\%) & 1233 (4.6\%) & 1796 (4.56\%) & 2319 (4.55\%) \\
 &  & 25,000 & 361004 (8.44\%) & 1144 (8.33\%) & 2260 (8.43\%) & 3351 (8.5\%) & 4382 (8.59\%) \\
 &  & 35,000 & 466936 (10.92\%) & 1436 (10.46\%) & 2734 (10.19\%) & 4085 (10.36\%) & 5257 (10.31\%) \\
 &  & 45,000 & 486317 (11.37\%) & 1325 (9.65\%) & 2692 (10.04\%) & 3924 (9.95\%) & 5114 (10.03\%) \\
 &  & 55,000 & 195326 (4.57\%) & 701 (5.11\%) & 1406 (5.24\%) & 2087 (5.29\%) & 2644 (5.18\%) \\
 &  & 65,000 & 408420 (9.55\%) & 1100 (8.01\%) & 2099 (7.83\%) & 3127 (7.93\%) & 4032 (7.91\%) \\
 &  & 75,000 & 369208 (8.63\%) & 840 (6.12\%) & 1602 (5.97\%) & 2341 (5.94\%) & 3087 (6.05\%) \\
 &  & 85,000 & 155571 (3.64\%) & 339 (2.47\%) & 661 (2.46\%) & 969 (2.46\%) & 1239 (2.43\%) \\
 &  & 95,000 & 193460 (4.52\%) & 425 (3.1\%) & 858 (3.2\%) & 1217 (3.09\%) & 1564 (3.07\%) \\
 &  & 112,500 & 202596 (4.74\%) & 395 (2.88\%) & 814 (3.04\%) & 1191 (3.02\%) & 1519 (2.98\%) \\
 &  & 137,500 & 53525 (1.25\%) & 92 (0.67\%) & 198 (0.74\%) & 304 (0.77\%) & 393 (0.77\%) \\
 &  & 175,000 & 161797 (3.78\%) & 270 (1.97\%) & 532 (1.98\%) & 752 (1.91\%) & 979 (1.92\%) \\
 &  & Unknown & 782646 (18.3\%) & 4116 (29.98\%) & 7923 (29.54\%) & 11606 (29.44\%) & 15070 (29.55\%) \\
\cmidrule(l){2-8}
 & \multirow{6}{=}{Marital Status} & Married & 2469391 (57.74\%) & 3953 (28.8\%) & 7886 (29.41\%) & 11508 (29.19\%) & 15124 (29.65\%) \\
 &  & Divorced & 904106 (21.14\%) & 4954 (36.09\%) & 9625 (35.89\%) & 14163 (35.93\%) & 18186 (35.66\%) \\
 &  & Not Married & 475057 (11.11\%) & 3215 (23.42\%) & 6156 (22.95\%) & 9025 (22.89\%) & 11538 (22.62\%) \\
 &  & Widowed & 288709 (6.75\%) & 614 (4.47\%) & 1197 (4.46\%) & 1834 (4.65\%) & 2414 (4.73\%) \\
 &  & Separated & 116933 (2.73\%) & 945 (6.88\%) & 1860 (6.94\%) & 2752 (6.98\%) & 3562 (6.98\%) \\
 &  & Unknown & 22207 (0.52\%) & 47 (0.34\%) & 94 (0.35\%) & 138 (0.35\%) & 178 (0.35\%) \\
\cmidrule(l){2-8}
 & Military Sexual Trauma (MST) &  & 33931 (0.79\%) & 284 (2.07\%) & 536 (2.0\%) & 754 (1.91\%) & 994 (1.95\%) \\
\cmidrule(l){2-8}
 & \multirow{6}{=}{Race} & White & 3196183 (74.74\%) & 8263 (60.19\%) & 16310 (60.82\%) & 23972 (60.81\%) & 30970 (60.72\%) \\
 &  & Black & 679256 (15.88\%) & 4214 (30.7\%) & 7997 (29.82\%) & 11827 (30.0\%) & 15370 (30.14\%) \\
 &  & Unknown & 289755 (6.78\%) & 822 (5.99\%) & 1674 (6.24\%) & 2423 (6.15\%) & 3078 (6.04\%) \\
 &  & Asian & 42507 (0.99\%) & 116 (0.84\%) & 227 (0.85\%) & 312 (0.79\%) & 414 (0.81\%) \\
 &  & Native Hawaii & 37427 (0.88\%) & 139 (1.01\%) & 245 (0.91\%) & 354 (0.9\%) & 482 (0.95\%) \\
 &  & American Indian & 31275 (0.73\%) & 174 (1.27\%) & 365 (1.36\%) & 532 (1.35\%) & 688 (1.35\%) \\
\cmidrule(l){2-8}
 & \multirow{3}{=}{Rural/urban} & Urban & 2669610 (62.43\%) & 10051 (73.22\%) & 19633 (73.21\%) & 28906 (73.33\%) & 37397 (73.32\%) \\
 &  & Rural & 1598419 (37.38\%) & 3663 (26.68\%) & 7143 (26.64\%) & 10448 (26.5\%) & 13510 (26.49\%) \\
 &  & Unknown & 8374 (0.2\%) & 14 (0.1\%) & 42 (0.16\%) & 66 (0.17\%) & 95 (0.19\%) \\
\cmidrule(l){2-8}
 & \multirow{52}{=}{State} & Texas & 347078 (8.12\%) & 1168 (8.51\%) & 2323 (8.66\%) & 3457 (8.77\%) & 4440 (8.71\%) \\
 &  & Florida & 344221 (8.05\%) & 938 (6.83\%) & 1816 (6.77\%) & 2745 (6.96\%) & 3609 (7.08\%) \\
 &  & California & 279602 (6.54\%) & 1253 (9.13\%) & 2491 (9.29\%) & 3806 (9.65\%) & 4936 (9.68\%) \\
 &  & Unknown & 212773 (4.98\%) & 621 (4.52\%) & 1241 (4.63\%) & 1877 (4.76\%) & 2581 (5.06\%) \\
 &  & Ohio & 170848 (4.0\%) & 503 (3.66\%) & 984 (3.67\%) & 1505 (3.82\%) & 1939 (3.8\%) \\
 &  & North Carolina & 169954 (3.97\%) & 453 (3.3\%) & 855 (3.19\%) & 1295 (3.29\%) & 1655 (3.24\%) \\
 &  & Georgia & 150705 (3.52\%) & 759 (5.53\%) & 1390 (5.18\%) & 1990 (5.05\%) & 2578 (5.05\%) \\
 &  & Pennsylvania & 144747 (3.38\%) & 414 (3.02\%) & 799 (2.98\%) & 1189 (3.02\%) & 1510 (2.96\%) \\
 &  & New York & 139224 (3.26\%) & 487 (3.55\%) & 917 (3.42\%) & 1361 (3.45\%) & 1734 (3.4\%) \\
 &  & Illinois & 117953 (2.76\%) & 413 (3.01\%) & 775 (2.89\%) & 1128 (2.86\%) & 1402 (2.75\%) \\
 &  & Tennessee & 111183 (2.6\%) & 334 (2.43\%) & 601 (2.24\%) & 836 (2.12\%) & 1098 (2.15\%) \\
 &  & Virginia & 110521 (2.58\%) & 276 (2.01\%) & 564 (2.1\%) & 864 (2.19\%) & 1142 (2.24\%) \\
 &  & Michigan & 107271 (2.51\%) & 344 (2.51\%) & 660 (2.46\%) & 946 (2.4\%) & 1258 (2.47\%) \\
 &  & South Carolina & 105475 (2.47\%) & 278 (2.03\%) & 541 (2.02\%) & 808 (2.05\%) & 1046 (2.05\%) \\
 &  & Arizona & 105156 (2.46\%) & 365 (2.66\%) & 733 (2.73\%) & 1068 (2.71\%) & 1389 (2.72\%) \\
 &  & Missouri & 99668 (2.33\%) & 220 (1.6\%) & 476 (1.77\%) & 705 (1.79\%) & 956 (1.87\%) \\
 &  & Indiana & 95681 (2.24\%) & 262 (1.91\%) & 514 (1.92\%) & 759 (1.93\%) & 977 (1.92\%) \\
 &  & Washington & 87701 (2.05\%) & 340 (2.48\%) & 654 (2.44\%) & 915 (2.32\%) & 1191 (2.34\%) \\
 &  & Wisconsin & 86173 (2.02\%) & 214 (1.56\%) & 397 (1.48\%) & 541 (1.37\%) & 702 (1.38\%) \\
 &  & Alabama & 84011 (1.96\%) & 273 (1.99\%) & 528 (1.97\%) & 724 (1.84\%) & 924 (1.81\%) \\
 &  & Minnesota & 78824 (1.84\%) & 116 (0.84\%) & 235 (0.88\%) & 331 (0.84\%) & 418 (0.82\%) \\
 &  & Kentucky & 77327 (1.81\%) & 182 (1.33\%) & 345 (1.29\%) & 536 (1.36\%) & 702 (1.38\%) \\
 &  & Oklahoma & 70552 (1.65\%) & 221 (1.61\%) & 411 (1.53\%) & 588 (1.49\%) & 759 (1.49\%) \\
 &  & Oregon & 66454 (1.55\%) & 521 (3.8\%) & 1013 (3.78\%) & 1273 (3.23\%) & 1475 (2.89\%) \\
 &  & Arkansas & 66055 (1.54\%) & 155 (1.13\%) & 325 (1.21\%) & 516 (1.31\%) & 647 (1.27\%) \\
 &  & Colorado & 63300 (1.48\%) & 209 (1.52\%) & 397 (1.48\%) & 598 (1.52\%) & 755 (1.48\%) \\
 &  & Louisiana & 61675 (1.44\%) & 226 (1.65\%) & 468 (1.75\%) & 653 (1.66\%) & 876 (1.72\%) \\
 &  & Nevada & 52287 (1.22\%) & 231 (1.68\%) & 454 (1.69\%) & 672 (1.7\%) & 876 (1.72\%) \\
 &  & Maryland & 50707 (1.19\%) & 168 (1.22\%) & 347 (1.29\%) & 506 (1.28\%) & 675 (1.32\%) \\
 &  & Mississippi & 50562 (1.18\%) & 148 (1.08\%) & 256 (0.95\%) & 382 (0.97\%) & 484 (0.95\%) \\
 &  & Massachusetts & 49617 (1.16\%) & 189 (1.38\%) & 428 (1.6\%) & 587 (1.49\%) & 768 (1.51\%) \\
 &  & New Jersey & 47907 (1.12\%) & 157 (1.14\%) & 313 (1.17\%) & 455 (1.15\%) & 572 (1.12\%) \\
 &  & Iowa & 46185 (1.08\%) & 120 (0.87\%) & 259 (0.97\%) & 389 (0.99\%) & 471 (0.92\%) \\
 &  & West Virginia & 42326 (0.99\%) & 109 (0.79\%) & 218 (0.81\%) & 301 (0.76\%) & 376 (0.74\%) \\
 &  & Kansas & 38515 (0.9\%) & 104 (0.76\%) & 195 (0.73\%) & 282 (0.72\%) & 382 (0.75\%) \\
 &  & New Mexico & 36755 (0.86\%) & 85 (0.62\%) & 186 (0.69\%) & 268 (0.68\%) & 367 (0.72\%) \\
 &  & Idaho & 34911 (0.82\%) & 82 (0.6\%) & 158 (0.59\%) & 256 (0.65\%) & 345 (0.68\%) \\
 &  & Nebraska & 32154 (0.75\%) & 55 (0.4\%) & 136 (0.51\%) & 218 (0.55\%) & 282 (0.55\%) \\
 &  & Connecticut & 31737 (0.74\%) & 86 (0.63\%) & 178 (0.66\%) & 284 (0.72\%) & 380 (0.75\%) \\
 &  & Montana & 26790 (0.63\%) & 79 (0.58\%) & 142 (0.53\%) & 208 (0.53\%) & 272 (0.53\%) \\
 &  & Utah & 26016 (0.61\%) & 77 (0.56\%) & 149 (0.56\%) & 230 (0.58\%) & 291 (0.57\%) \\
 &  & Maine & 22395 (0.52\%) & 59 (0.43\%) & 115 (0.43\%) & 162 (0.41\%) & 205 (0.4\%) \\
 &  & South Dakota & 21384 (0.5\%) & 42 (0.31\%) & 91 (0.34\%) & 154 (0.39\%) & 208 (0.41\%) \\
 &  & New Hampshire & 19668 (0.46\%) & 79 (0.58\%) & 142 (0.53\%) & 201 (0.51\%) & 246 (0.48\%) \\
 &  & Hawaii & 17112 (0.4\%) & 56 (0.41\%) & 113 (0.42\%) & 151 (0.38\%) & 194 (0.38\%) \\
 &  & Wyoming & 13787 (0.32\%) & 43 (0.31\%) & 76 (0.28\%) & 98 (0.25\%) & 115 (0.23\%) \\
 &  & North Dakota & 13753 (0.32\%) & 37 (0.27\%) & 74 (0.28\%) & 98 (0.25\%) & 136 (0.27\%) \\
 &  & Rhode Island & 12636 (0.3\%) & 29 (0.21\%) & 61 (0.23\%) & 93 (0.24\%) & 135 (0.26\%) \\
 &  & Delaware & 11172 (0.26\%) & 39 (0.28\%) & 82 (0.31\%) & 119 (0.3\%) & 159 (0.31\%) \\
 &  & Alaska & 10696 (0.25\%) & 34 (0.25\%) & 67 (0.25\%) & 96 (0.24\%) & 114 (0.22\%) \\
 &  & Vermont & 9554 (0.22\%) & 39 (0.28\%) & 60 (0.22\%) & 93 (0.24\%) & 123 (0.24\%) \\
 &  & District Of Columbia & 3645 (0.09\%) & 36 (0.26\%) & 65 (0.24\%) & 103 (0.26\%) & 127 (0.25\%) \\
\cmidrule(l){2-8}
 & \multirow{11}{=}{Service connected disability} & 0\% & 1481706 (34.65\%) & 4291 (31.26\%) & 8289 (30.91\%) & 12137 (30.79\%) & 15595 (30.58\%) \\
 &  & 10\% & 267544 (6.26\%) & 828 (6.03\%) & 1681 (6.27\%) & 2462 (6.25\%) & 3187 (6.25\%) \\
 &  & 20\% & 131359 (3.07\%) & 328 (2.39\%) & 685 (2.55\%) & 1023 (2.6\%) & 1352 (2.65\%) \\
 &  & 30\% & 204318 (4.78\%) & 679 (4.95\%) & 1290 (4.81\%) & 1920 (4.87\%) & 2517 (4.94\%) \\
 &  & 40\% & 130930 (3.06\%) & 337 (2.45\%) & 672 (2.51\%) & 1020 (2.59\%) & 1299 (2.55\%) \\
 &  & 50\% & 114961 (2.69\%) & 331 (2.41\%) & 644 (2.4\%) & 945 (2.4\%) & 1204 (2.36\%) \\
 &  & 60\% & 187984 (4.4\%) & 495 (3.61\%) & 968 (3.61\%) & 1413 (3.58\%) & 1845 (3.62\%) \\
 &  & 70\% & 256197 (5.99\%) & 1058 (7.71\%) & 2099 (7.83\%) & 3088 (7.83\%) & 3978 (7.8\%) \\
 &  & 80\% & 282927 (6.62\%) & 1036 (7.55\%) & 1971 (7.35\%) & 2880 (7.31\%) & 3749 (7.35\%) \\
 &  & 90\% & 298501 (6.98\%) & 1042 (7.59\%) & 2020 (7.53\%) & 3013 (7.64\%) & 3903 (7.65\%) \\
 &  & 100\% & 919976 (21.51\%) & 3303 (24.06\%) & 6499 (24.23\%) & 9519 (24.15\%) & 12373 (24.26\%) \\
\cmidrule(l){2-8}
 & \multirow{18}{=}{VISN} & 8 & 409694 (9.58\%) & 1022 (7.44\%) & 2038 (7.6\%) & 3137 (7.96\%) & 4346 (8.52\%) \\
 &  & 10 & 344367 (8.05\%) & 1043 (7.6\%) & 2023 (7.54\%) & 3004 (7.62\%) & 3878 (7.6\%) \\
 &  & 22 & 319139 (7.46\%) & 1385 (10.09\%) & 2783 (10.38\%) & 4049 (10.27\%) & 5252 (10.3\%) \\
 &  & 7 & 305471 (7.14\%) & 1217 (8.87\%) & 2278 (8.49\%) & 3256 (8.26\%) & 4219 (8.27\%) \\
 &  & 16 & 289020 (6.76\%) & 840 (6.12\%) & 1679 (6.26\%) & 2528 (6.41\%) & 3334 (6.54\%) \\
 &  & 17 & 271647 (6.35\%) & 868 (6.32\%) & 1711 (6.38\%) & 2506 (6.36\%) & 3163 (6.2\%) \\
 &  & 6 & 257847 (6.03\%) & 690 (5.03\%) & 1342 (5.0\%) & 2037 (5.17\%) & 2625 (5.15\%) \\
 &  & 21 & 225793 (5.28\%) & 898 (6.54\%) & 1761 (6.57\%) & 2724 (6.91\%) & 3517 (6.9\%) \\
 &  & 23 & 214793 (5.02\%) & 410 (2.99\%) & 861 (3.21\%) & 1308 (3.32\%) & 1693 (3.32\%) \\
 &  & 19 & 208984 (4.89\%) & 608 (4.43\%) & 1164 (4.34\%) & 1769 (4.49\%) & 2245 (4.4\%) \\
 &  & 20 & 202901 (4.74\%) & 1023 (7.45\%) & 1978 (7.38\%) & 2635 (6.68\%) & 3254 (6.38\%) \\
 &  & 9 & 197768 (4.62\%) & 530 (3.86\%) & 989 (3.69\%) & 1440 (3.65\%) & 1900 (3.73\%) \\
 &  & 12 & 197604 (4.62\%) & 635 (4.63\%) & 1173 (4.37\%) & 1680 (4.26\%) & 2126 (4.17\%) \\
 &  & 2 & 188982 (4.42\%) & 659 (4.8\%) & 1239 (4.62\%) & 1846 (4.68\%) & 2374 (4.65\%) \\
 &  & 4 & 184479 (4.31\%) & 566 (4.12\%) & 1101 (4.11\%) & 1582 (4.01\%) & 1969 (3.86\%) \\
 &  & 15 & 170893 (4.0\%) & 406 (2.96\%) & 823 (3.07\%) & 1225 (3.11\%) & 1611 (3.16\%) \\
 &  & 1 & 158642 (3.71\%) & 522 (3.8\%) & 1058 (3.95\%) & 1524 (3.87\%) & 1976 (3.87\%) \\
 &  & 5 & 128379 (3.0\%) & 406 (2.96\%) & 817 (3.05\%) & 1170 (2.97\%) & 1520 (2.98\%) \\
\multirow{51}{*}{Service Utilization} & \multirow{6}{=}{Outpatient Visits: Primary care} & 0 & 208605 (4.88\%) & 1559 (11.36\%) & 2864 (10.68\%) & 4023 (10.21\%) & 5000 (9.8\%) \\
 &  & 1 & 902376 (21.1\%) & 2149 (15.65\%) & 4176 (15.57\%) & 6109 (15.5\%) & 7907 (15.5\%) \\
 &  & 2 & 847575 (19.82\%) & 2056 (14.98\%) & 4075 (15.2\%) & 6033 (15.3\%) & 7879 (15.45\%) \\
 &  & 3 & 623847 (14.59\%) & 1767 (12.87\%) & 3493 (13.02\%) & 5152 (13.07\%) & 6708 (13.15\%) \\
 &  & 4 & 439139 (10.27\%) & 1311 (9.55\%) & 2589 (9.65\%) & 3886 (9.86\%) & 5130 (10.06\%) \\
 &  & 5+ & 1254861 (29.34\%) & 4886 (35.59\%) & 9621 (35.88\%) & 14217 (36.07\%) & 18378 (36.03\%) \\
\cmidrule(l){2-8}
 & \multirow{6}{=}{Outpatient Visits: Mental health} & 0 & 3015033 (70.5\%) & 5752 (41.9\%) & 11531 (43.0\%) & 17272 (43.82\%) & 22665 (44.44\%) \\
 &  & 1 & 264321 (6.18\%) & 1404 (10.23\%) & 2737 (10.21\%) & 3988 (10.12\%) & 5067 (9.93\%) \\
 &  & 2 & 198294 (4.64\%) & 955 (6.96\%) & 1857 (6.92\%) & 2770 (7.03\%) & 3640 (7.14\%) \\
 &  & 3 & 160804 (3.76\%) & 730 (5.32\%) & 1483 (5.53\%) & 2189 (5.55\%) & 2899 (5.68\%) \\
 &  & 4 & 128044 (2.99\%) & 641 (4.67\%) & 1303 (4.86\%) & 1868 (4.74\%) & 2410 (4.73\%) \\
 &  & 5+ & 509907 (11.92\%) & 4246 (30.93\%) & 7907 (29.48\%) & 11333 (28.75\%) & 14321 (28.08\%) \\
\cmidrule(l){2-8}
 & {Outpatient Visits: Substance abuse} & 0 & 4193235 (98.06\%) & 11972 (87.21\%) & 23709 (88.41\%) & 35129 (89.11\%) & 45721 (89.65\%) \\
 &  & 1 & 26372 (0.62\%) & 420 (3.06\%) & 809 (3.02\%) & 1133 (2.87\%) & 1386 (2.72\%) \\
 &  & 2 & 9769 (0.23\%) & 191 (1.39\%) & 340 (1.27\%) & 471 (1.19\%) & 591 (1.16\%) \\
 &  & 3+ & 47027 (1.1\%) & 1145 (8.34\%) & 1960 (7.31\%) & 2687 (6.82\%) & 3304 (6.48\%) \\
\cmidrule(l){2-8}
 & \multirow{6}{=}{Outpatient Visits: Specialty care} & 0 & 1376548 (32.19\%) & 4740 (34.53\%) & 9167 (34.18\%) & 13293 (33.72\%) & 16929 (33.19\%) \\
 &  & 1 & 645905 (15.1\%) & 1967 (14.33\%) & 3790 (14.13\%) & 5578 (14.15\%) & 7306 (14.32\%) \\
 &  & 2 & 450014 (10.52\%) & 1256 (9.15\%) & 2568 (9.58\%) & 3804 (9.65\%) & 4953 (9.71\%) \\
 &  & 3 & 325704 (7.62\%) & 933 (6.8\%) & 1840 (6.86\%) & 2715 (6.89\%) & 3593 (7.04\%) \\
 &  & 4 & 248658 (5.81\%) & 808 (5.89\%) & 1542 (5.75\%) & 2292 (5.81\%) & 2989 (5.86\%) \\
 &  & 5+ & 1229574 (28.75\%) & 4024 (29.31\%) & 7911 (29.5\%) & 11738 (29.78\%) & 15232 (29.87\%) \\
\cmidrule(l){2-8}
 & \multirow{6}{=}{Outpatient Visits: Rehabilitation} & 0 & 2705599 (63.27\%) & 8331 (60.69\%) & 16445 (61.32\%) & 24139 (61.24\%) & 31366 (61.5\%) \\
 &  & 1 & 548830 (12.83\%) & 1751 (12.75\%) & 3454 (12.88\%) & 5198 (13.19\%) & 6733 (13.2\%) \\
 &  & 2 & 326780 (7.64\%) & 959 (6.99\%) & 1858 (6.93\%) & 2696 (6.84\%) & 3469 (6.8\%) \\
 &  & 3 & 194497 (4.55\%) & 569 (4.14\%) & 1115 (4.16\%) & 1648 (4.18\%) & 2114 (4.14\%) \\
 &  & 4 & 120933 (2.83\%) & 392 (2.86\%) & 738 (2.75\%) & 1085 (2.75\%) & 1401 (2.75\%) \\
 &  & 5+ & 379764 (8.88\%) & 1726 (12.57\%) & 3208 (11.96\%) & 4654 (11.81\%) & 5919 (11.61\%) \\
\cmidrule(l){2-8}
 & \multirow{6}{=}{Outpatient Visits: Diagnostic / Ancillary} & 0 & 287421 (6.72\%) & 1134 (8.26\%) & 2119 (7.9\%) & 3003 (7.62\%) & 3738 (7.33\%) \\
 &  & 1 & 635550 (14.86\%) & 1395 (10.16\%) & 2720 (10.14\%) & 4027 (10.22\%) & 5211 (10.22\%) \\
 &  & 2 & 579045 (13.54\%) & 1324 (9.64\%) & 2729 (10.18\%) & 4015 (10.19\%) & 5278 (10.35\%) \\
 &  & 3 & 457623 (10.7\%) & 1186 (8.64\%) & 2407 (8.98\%) & 3563 (9.04\%) & 4683 (9.18\%) \\
 &  & 4 & 354861 (8.3\%) & 1046 (7.62\%) & 2063 (7.69\%) & 3060 (7.76\%) & 4019 (7.88\%) \\
 &  & 5+ & 1961903 (45.88\%) & 7643 (55.67\%) & 14780 (55.11\%) & 21752 (55.18\%) & 28073 (55.04\%) \\
\cmidrule(l){2-8}
 & \multirow{4}{=}{Emergency / Urgent-care} & 0 & 3353468 (78.42\%) & 8046 (58.61\%) & 15979 (59.58\%) & 23708 (60.14\%) & 30991 (60.76\%) \\
 &  & 1 & 520258 (12.17\%) & 2713 (19.76\%) & 5208 (19.42\%) & 7539 (19.12\%) & 9646 (18.91\%) \\
 &  & 2 & 204184 (4.77\%) & 1291 (9.4\%) & 2493 (9.3\%) & 3650 (9.26\%) & 4640 (9.1\%) \\
 &  & 3+ & 198493 (4.64\%) & 1678 (12.22\%) & 3138 (11.7\%) & 4523 (11.47\%) & 5725 (11.23\%) \\
\cmidrule(l){2-8}
 & \multirow{4}{=}{Inpatient Visits: Total} & 0 & 3950804 (92.39\%) & 11407 (83.09\%) & 22567 (84.15\%) & 33475 (84.92\%) & 43602 (85.49\%) \\
 &  & 1 & 108068 (2.53\%) & 786 (5.73\%) & 1469 (5.48\%) & 2071 (5.25\%) & 2590 (5.08\%) \\
 &  & 2 & 73840 (1.73\%) & 597 (4.35\%) & 1037 (3.87\%) & 1451 (3.68\%) & 1796 (3.52\%) \\
 &  & 3+ & 143691 (3.36\%) & 938 (6.83\%) & 1745 (6.51\%) & 2423 (6.15\%) & 3014 (5.91\%) \\
\cmidrule(l){2-8}
 & \multirow{4}{=}{Inpatient Visits: No. of Days} & 0 & 3950834 (92.39\%) & 11407 (83.09\%) & 22567 (84.15\%) & 33475 (84.92\%) & 43602 (85.49\%) \\
 &  & 1 & 35603 (0.83\%) & 160 (1.17\%) & 335 (1.25\%) & 485 (1.23\%) & 622 (1.22\%) \\
 &  & 2 & 40262 (0.94\%) & 246 (1.79\%) & 445 (1.66\%) & 657 (1.67\%) & 829 (1.63\%) \\
 &  & 3+ & 249704 (5.84\%) & 1915 (13.95\%) & 3471 (12.94\%) & 4803 (12.18\%) & 5949 (11.66\%) \\
\cmidrule(l){2-8}
 & \multirow{5}{=}{VISN count} & 1 & 2034356 (47.57\%) & 4516 (32.9\%) & 8772 (32.71\%) & 12789 (32.44\%) & 16500 (32.35\%) \\
 &  & 2 & 1282822 (30.0\%) & 4208 (30.65\%) & 8326 (31.05\%) & 12241 (31.05\%) & 15763 (30.91\%) \\
 &  & 3 & 587032 (13.73\%) & 2597 (18.92\%) & 5032 (18.76\%) & 7494 (19.01\%) & 9763 (19.14\%) \\
 &  & 4 & 232504 (5.44\%) & 1304 (9.5\%) & 2536 (9.46\%) & 3729 (9.46\%) & 4857 (9.52\%) \\
 &  & 5+ & 139689 (3.27\%) & 1103 (8.03\%) & 2152 (8.02\%) & 3167 (8.03\%) & 4119 (8.08\%) \\
\midrule\
\multirow{8}{*}{\makecell[l]{Mental\\Health\\Disorders}} & \multirow{1}{=}{Anxiety Disorder} &  & 369175 (8.63\%) & 2320 (16.9\%) & 4498 (16.77\%) & 6462 (16.39\%) & 8313 (16.3\%) \\
\cmidrule(l){2-8}
 & \multirow{1}{=}{Bipolar Disorder} &  & 79218 (1.85\%) & 829 (6.04\%) & 1619 (6.04\%) & 2331 (5.91\%) & 2979 (5.84\%) \\
\cmidrule(l){2-8}
 & \multirow{1}{=}{Dementia} &  & 125731 (2.94\%) & 262 (1.91\%) & 522 (1.95\%) & 732 (1.86\%) & 917 (1.8\%) \\
\cmidrule(l){2-8}
 & \multirow{1}{=}{Depression} &  & 727811 (17.02\%) & 4628 (33.71\%) & 8954 (33.39\%) & 12939 (32.82\%) & 16591 (32.53\%) \\
\cmidrule(l){2-8}
 & Other Neurological Disorders &  & 168145 (3.93\%) & 661 (4.81\%) & 1272 (4.74\%) & 1815 (4.6\%) & 2282 (4.47\%) \\
\cmidrule(l){2-8}
 & Posttraumatic Stress Disorder &  & 612738 (14.33\%) & 3348 (24.39\%) & 6423 (23.95\%) & 9358 (23.74\%) & 11940 (23.41\%) \\
\cmidrule(l){2-8}
 & \multirow{1}{=}{Psychoses} &  & 67160 (1.57\%) & 741 (5.4\%) & 1408 (5.25\%) & 2060 (5.23\%) & 2587 (5.07\%) \\
\cmidrule(l){2-8}
 & \multirow{1}{=}{Sleep Disorder} &  & 728343 (17.03\%) & 2453 (17.87\%) & 4766 (17.77\%) & 6976 (17.7\%) & 9095 (17.83\%) \\
\midrule\
\multirow{30}{*}{\makecell[l]{Physical\\ Health\\ Disorders}} & \multirow{1}{=}{Blood Loss Anemia} &  & 13449 (0.31\%) & 52 (0.38\%) & 103 (0.38\%) & 150 (0.38\%) & 190 (0.37\%) \\
\cmidrule(l){2-8}
 & \multirow{1}{=}{Cardiac Arrhythmia} &  & 441826 (10.33\%) & 1000 (7.28\%) & 1974 (7.36\%) & 2869 (7.28\%) & 3692 (7.24\%) \\
\cmidrule(l){2-8}
 & Cardiovascular Disease &  & 41516 (0.97\%) & 140 (1.02\%) & 275 (1.03\%) & 385 (0.98\%) & 507 (0.99\%) \\
\cmidrule(l){2-8}
 & {Chronic Pulmonary Disease} &  & 484324 (11.33\%) & 1547 (11.27\%) & 3014 (11.24\%) & 4415 (11.2\%) & 5642 (11.06\%) \\
\cmidrule(l){2-8}
 & \multirow{1}{=}{Cirrhosis} &  & 36720 (0.86\%) & 195 (1.42\%) & 406 (1.51\%) & 586 (1.49\%) & 724 (1.42\%) \\
\cmidrule(l){2-8}
 & \multirow{1}{=}{Coagulopathy} &  & 54368 (1.27\%) & 200 (1.46\%) & 361 (1.35\%) & 504 (1.28\%) & 647 (1.27\%) \\
\cmidrule(l){2-8}
 & Congestive Heart Failure &  & 202163 (4.73\%) & 611 (4.45\%) & 1223 (4.56\%) & 1787 (4.53\%) & 2234 (4.38\%) \\
\cmidrule(l){2-8}
 & \multirow{1}{=}{Deficiency Anemia} &  & 114764 (2.68\%) & 355 (2.59\%) & 689 (2.57\%) & 1007 (2.55\%) & 1300 (2.55\%) \\
\cmidrule(l){2-8}
 & \multirow{1}{=}{Diabetes} &  & 1132896 (26.49\%) & 2685 (19.56\%) & 5259 (19.61\%) & 7886 (20.01\%) & 10375 (20.34\%) \\
\cmidrule(l){2-8}
 & Fluid and Electrolyte Disorders &  & 142822 (3.34\%) & 679 (4.95\%) & 1313 (4.9\%) & 1845 (4.68\%) & 2321 (4.55\%) \\
\cmidrule(l){2-8}
 & \multirow{1}{=}{Hepatitis} &  & 86619 (2.03\%) & 701 (5.11\%) & 1356 (5.06\%) & 2006 (5.09\%) & 2543 (4.99\%) \\
\cmidrule(l){2-8}
 & \multirow{1}{=}{HIV} &  & 16553 (0.39\%) & 138 (1.01\%) & 260 (0.97\%) & 404 (1.02\%) & 501 (0.98\%) \\
\cmidrule(l){2-8}
 & \multirow{1}{=}{Hypertension} &  & 1971448 (46.1\%) & 4966 (36.17\%) & 9703 (36.18\%) & 14281 (36.23\%) & 18605 (36.48\%) \\
\cmidrule(l){2-8}
 & \multirow{1}{=}{Hypothyroidism} &  & 269726 (6.31\%) & 572 (4.17\%) & 1103 (4.11\%) & 1678 (4.26\%) & 2156 (4.23\%) \\
\cmidrule(l){2-8}
 & \multirow{1}{=}{Influenza} &  & 7221 (0.17\%) & 30 (0.22\%) & 64 (0.24\%) & 110 (0.28\%) & 149 (0.29\%) \\
\cmidrule(l){2-8}
 & \multirow{1}{=}{Liver Disease} &  & 62564 (1.46\%) & 289 (2.11\%) & 568 (2.12\%) & 804 (2.04\%) & 1045 (2.05\%) \\
\cmidrule(l){2-8}
 & \multirow{1}{=}{Lymphoma} &  & 25603 (0.6\%) & 41 (0.3\%) & 77 (0.29\%) & 114 (0.29\%) & 159 (0.31\%) \\
\cmidrule(l){2-8}
 & \multirow{1}{=}{Metastatic Cancer} &  & 30586 (0.72\%) & 62 (0.45\%) & 118 (0.44\%) & 168 (0.43\%) & 202 (0.4\%) \\
\cmidrule(l){2-8}
 & \multirow{1}{=}{Obesity} &  & 535709 (12.53\%) & 1642 (11.96\%) & 3270 (12.19\%) & 4890 (12.4\%) & 6329 (12.41\%) \\
\cmidrule(l){2-8}
 & \multirow{1}{=}{Pain} &  & 2285963 (53.46\%) & 8152 (59.38\%) & 15986 (59.61\%) & 23504 (59.62\%) & 30432 (59.67\%) \\
\cmidrule(l){2-8}
 & \multirow{1}{=}{Paralysis} &  & 23737 (0.56\%) & 101 (0.74\%) & 176 (0.66\%) & 261 (0.66\%) & 320 (0.63\%) \\
\cmidrule(l){2-8}
 & {Peptic Ulcer Disease} &  & 10804 (0.25\%) & 40 (0.29\%) & 81 (0.3\%) & 121 (0.31\%) & 152 (0.3\%) \\
\cmidrule(l){2-8}
 & Peripheral Vascular Disorders &  & 188758 (4.41\%) & 475 (3.46\%) & 939 (3.5\%) & 1350 (3.42\%) & 1762 (3.45\%) \\
\cmidrule(l){2-8}
 & Pulmonary Circulation Disorders &  & 34361 (0.8\%) & 107 (0.78\%) & 224 (0.84\%) & 324 (0.82\%) & 410 (0.8\%) \\
\cmidrule(l){2-8}
 & \multirow{1}{=}{Renal Failure} &  & 240999 (5.64\%) & 619 (4.51\%) & 1189 (4.43\%) & 1692 (4.29\%) & 2140 (4.2\%) \\
\cmidrule(l){2-8}
 & {Rheumatoid Arthritis/collagen} &  & 62803 (1.47\%) & 166 (1.21\%) & 329 (1.23\%) & 481 (1.22\%) & 614 (1.2\%) \\
\cmidrule(l){2-8}
 & {Solid Tumor without Metastasis} &  & 237679 (5.56\%) & 484 (3.53\%) & 989 (3.69\%) & 1480 (3.75\%) & 1915 (3.75\%) \\
\cmidrule(l){2-8}
 & {Traumatic brain injury} &  & 53858 (1.26\%) & 366 (2.67\%) & 729 (2.72\%) & 1069 (2.71\%) & 1331 (2.61\%) \\
\cmidrule(l){2-8}
 & \multirow{1}{=}{Valvular Disease} &  & 86468 (2.02\%) & 182 (1.33\%) & 349 (1.3\%) & 514 (1.3\%) & 654 (1.28\%) \\
\cmidrule(l){2-8}
 & \multirow{1}{=}{Weight Loss} &  & 61445 (1.44\%) & 276 (2.01\%) & 520 (1.94\%) & 749 (1.9\%) & 954 (1.87\%) \\
\midrule\
\multirow{8}{*}{\makecell[l]{Substance\\ Abuse\\ Disorders}} & {Alcohol use disorder} &  & 263595 (6.16\%) & 2753 (20.05\%) & 5105 (19.04\%) & 7290 (18.49\%) & 9158 (17.96\%) \\
\cmidrule(l){2-8}
 & \multirow{1}{=}{Cannabis} &  & 58342 (1.36\%) & 1006 (7.33\%) & 1920 (7.16\%) & 2708 (6.87\%) & 3423 (6.71\%) \\
\cmidrule(l){2-8}
 & \multirow{1}{=}{Cocaine} &  & 23833 (0.56\%) & 702 (5.11\%) & 1249 (4.66\%) & 1747 (4.43\%) & 2184 (4.28\%) \\
\cmidrule(l){2-8}
 & \multirow{1}{=}{Drug Abuse} &  & 96809 (2.26\%) & 1915 (13.95\%) & 3517 (13.11\%) & 4966 (12.6\%) & 6213 (12.18\%) \\
\cmidrule(l){2-8}
 & \multirow{1}{=}{Hallucinogen} &  & 591 (0.01\%) & 16 (0.12\%) & 31 (0.12\%) & 43 (0.11\%) & 53 (0.1\%) \\
\cmidrule(l){2-8}
 & {Nicotine dependence} &  & 315904 (7.39\%) & 2047 (14.91\%) & 3912 (14.59\%) & 5659 (14.36\%) & 7194 (14.11\%) \\
\cmidrule(l){2-8}
 & {Opioid use disorder} &  & 33303 (0.78\%) & 654 (4.76\%) & 1206 (4.5\%) & 1695 (4.3\%) & 2066 (4.05\%) \\
\cmidrule(l){2-8}
 & \multirow{1}{=}{Other stimulant} &  & 8220 (0.19\%) & 294 (2.14\%) & 541 (2.02\%) & 733 (1.86\%) & 907 (1.78\%) \\
\midrule\
\multirow{22}{*}{\makecell[l]{Social and \\Behavioral \\Factors of Health}} & \multirow{11}{=}{ADI} & 0-9.9 & 132341 (3.09\%) & 428 (3.12\%) & 884 (3.3\%) & 1330 (3.37\%) & 1686 (3.31\%) \\
 &  & 10-19.9 & 243070 (5.68\%) & 820 (5.97\%) & 1606 (5.99\%) & 2369 (6.01\%) & 3070 (6.02\%) \\
 &  & 20-29.9 & 342748 (8.01\%) & 974 (7.09\%) & 1976 (7.37\%) & 2918 (7.4\%) & 3697 (7.25\%) \\
 &  & 30-39.9 & 454757 (10.63\%) & 1269 (9.24\%) & 2454 (9.15\%) & 3594 (9.12\%) & 4684 (9.18\%) \\
 &  & 40-49.9 & 501180 (11.72\%) & 1361 (9.91\%) & 2767 (10.32\%) & 4007 (10.16\%) & 5137 (10.07\%) \\
 &  & 50-59.9 & 524171 (12.26\%) & 1544 (11.25\%) & 2941 (10.97\%) & 4265 (10.82\%) & 5462 (10.71\%) \\
 &  & 60-69.9 & 520226 (12.17\%) & 1492 (10.87\%) & 2947 (10.99\%) & 4395 (11.15\%) & 5672 (11.12\%) \\
 &  & 70-79.9 & 507166 (11.86\%) & 1614 (11.76\%) & 3084 (11.5\%) & 4565 (11.58\%) & 5873 (11.52\%) \\
 &  & 80-89.9 & 488948 (11.43\%) & 1651 (12.03\%) & 3211 (11.97\%) & 4743 (12.03\%) & 6168 (12.09\%) \\
 &  & 90-100 & 482767 (11.29\%) & 2070 (15.08\%) & 4055 (15.12\%) & 5994 (15.21\%) & 7967 (15.62\%) \\
 &  & Unknown & 79029 (1.85\%) & 505 (3.68\%) & 893 (3.33\%) & 1240 (3.15\%) & 1586 (3.11\%) \\
\cmidrule(l){2-8}
 & {Employment or financial problems} &  & 71035 (1.66\%) & 1808 (13.17\%) & 3150 (11.75\%) & 4281 (10.86\%) & 5181 (10.16\%) \\
\cmidrule(l){2-8}
 & \multirow{1}{=}{Food insecurity} &  & 178 (0.0\%) & 12 (0.09\%) & 17 (0.06\%) & 24 (0.06\%) & 29 (0.06\%) \\
\cmidrule(l){2-8}
 & \multirow{1}{=}{Housing problems} &  & 16388 (0.38\%) & 746 (5.43\%) & 1380 (5.15\%) & 1804 (4.58\%) & 2129 (4.17\%) \\
\cmidrule(l){2-8}
 & \multirow{4}{=}{Insurance} & Unknown & 1965156 (45.95\%) & 8529 (62.13\%) & 16416 (61.21\%) & 24105 (61.15\%) & 31068 (60.92\%) \\
 &  & Government & 1155865 (27.03\%) & 2699 (19.66\%) & 5337 (19.9\%) & 7890 (20.02\%) & 10247 (20.09\%) \\
 &  & Private & 600385 (14.04\%) & 960 (6.99\%) & 1965 (7.33\%) & 2845 (7.22\%) & 3734 (7.32\%) \\
 &  & Specialized & 554997 (12.98\%) & 1540 (11.22\%) & 3100 (11.56\%) & 4580 (11.62\%) & 5953 (11.67\%) \\
\cmidrule(l){2-8}
 & \multirow{1}{=}{Legal problems} &  & 24419 (0.57\%) & 905 (6.59\%) & 1532 (5.71\%) & 1990 (5.05\%) & 2363 (4.63\%) \\
\cmidrule(l){2-8}
 & {Non-specific psychosocial needs} &  & 364068 (8.51\%) & 2872 (20.92\%) & 5286 (19.71\%) & 7279 (18.47\%) & 9098 (17.84\%) \\
\cmidrule(l){2-8}
 & {Social or familial problems} &  & 54496 (1.27\%) & 339 (2.47\%) & 631 (2.35\%) & 876 (2.22\%) & 1186 (2.33\%) \\
\cmidrule(l){2-8}
 & {Violence problems} &  & 347251 (8.12\%) & 2912 (21.21\%) & 5353 (19.96\%) & 7373 (18.7\%) & 9102 (17.85\%) \\

\end{longtable}
\end{center}
}

\clearpage

\section*{Supplementary Table 2. Ablation of the condition persistence framework}\phantomsection
\label{supp_table_2}

\begin{table*}[!ht]
\scriptsize
\renewcommand{\arraystretch}{0.95}
\setlength{\tabcolsep}{4pt}
\centering

\begin{tabularx}{\textwidth}{l X l l l}
\toprule
\textbf{Model Class} & \textbf{Model} & \textbf{Input Representation} & \textbf{PR-AUC, \%} & \textbf{ROC AUC, \%} \\
\midrule

\multicolumn{5}{c}{\textbf{Prediction Window = 3 Months}} \\
\midrule
\multirow{6}{*}{Machine Learning} 
 & \multirow{2}{*}{Elastic Net LR} & Time Varying & 1.87 (1.54, 2.45) & 76.71 (74.79, 78.46) \\
 & & TV w/o Fill Strategy & 1.75 (1.45, 2.22) & 76.71 (74.82, 78.51) \\
\cmidrule(lr){2-5}
 & \multirow{2}{*}{Random Forest} & Time Varying & 2.11 (1.76, 2.68) & 79.27 (77.54, 80.94) \\
 & & TV w/o Fill Strategy & 1.87 (1.54, 2.44) & 78.76 (77.03, 80.39) \\
\cmidrule(lr){2-5}
 & \multirow{2}{*}{XGBoost} & Time Varying & 2.22 (1.83, 2.91) & 79.73 (77.95, 81.37) \\
 & & TV w/o Fill Strategy & 2.22 (1.81, 2.98) & 79.72 (77.99, 81.34) \\
\midrule
\multirow{4}{*}{\makecell[l]{Masked Language\\Models}} 
 & \multirow{2}{*}{ModernBERT} & Time Varying & 2.39 (1.8, 3.34) & 77.28 (75.39, 79.00) \\
 & & TV w/o Fill Strategy & 2.03 (1.64, 2.74) & 79.36 (76.57, 80.11) \\
\cmidrule(lr){2-5}
 & \multirow{2}{*}{BioClinical ModernBERT} & Time Varying & 2.34 (1.91, 3.09) & 78.78 (76.97, 80.53) \\
 & & TV w/o Fill Strategy & 2.61 (2.01, 3.55) & 78.94 (77.03, 80.59) \\
\midrule
\multirow{4}{*}{\makecell[l]{Large Language\\Models}} 
 & \multirow{2}{*}{Llama-3.1-8B} & Time Varying & 2.16 (1.77, 2.76) & 79.12 (77.43, 80.79) \\
 & & TV w/o Fill Strategy & 1.93 (1.59, 2.51) & 78.49 (76.74, 80.17) \\
\cmidrule(lr){2-5}
 & \multirow{2}{*}{OpenBioLLM-8B} & Time Varying & 2.32 (1.81, 3.10) & 78.31 (76.49, 79.95) \\
 & & TV w/o Fill Strategy & 2.36 (1.81, 3.10) & 77.87 (76.01, 79.62) \\
\midrule

\multicolumn{5}{c}{\textbf{Prediction Window = 6 Months}} \\
\midrule
\multirow{6}{*}{Machine Learning} 
 & \multirow{2}{*}{Elastic Net LR} & Time Varying & 3.28 (2.83, 3.97) & 74.83 (73.46, 76.18) \\
 & & TV w/o Fill Strategy & 3.14 (2.73, 3.76) & 74.66 (73.32, 76.01) \\
\cmidrule(lr){2-5}
 & \multirow{2}{*}{Random Forest} & Time Varying & 3.86 (3.32, 4.61) & 77.98 (76.77, 79.24) \\
 & & TV w/o Fill Strategy & 3.86 (3.32, 4.60) & 78.02 (76.82, 79.26) \\
\cmidrule(lr){2-5}
 & \multirow{2}{*}{XGBoost} & Time Varying & 4.13 (3.48, 5.01) & 78.45 (77.20, 79.67) \\
 & & TV w/o Fill Strategy & 3.97 (3.38, 4.77) & 78.47 (77.25, 79.68) \\
\midrule
\multirow{4}{*}{\makecell[l]{Masked Language\\Models}} 
 & \multirow{2}{*}{ModernBERT} & Time Varying & 3.54 (3.07, 4.14) & 77.49 (76.23, 78.73) \\
 & & TV w/o Fill Strategy & 3.43 (3.01, 4.00) & 77.21 (75.95, 78.47) \\
\cmidrule(lr){2-5}
 & \multirow{2}{*}{BioClinical ModernBERT} & Time Varying & 3.58 (3.12, 4.21) & 76.60 (75.33, 77.93) \\
 & & TV w/o Fill Strategy & 3.53 (3.05, 4.20) & 77.21 (75.93, 78.46) \\
\midrule
\multirow{4}{*}{\makecell[l]{Large Language\\Models}} 
 & \multirow{2}{*}{Llama-3.1-8B} & Time Varying & 4.12 (3.52, 4.98) & 78.43 (77.21, 79.66) \\
 & & TV w/o Fill Strategy & 3.49 (3.06, 4.11) & 76.85 (75.56, 78.12) \\
\cmidrule(lr){2-5}
 & \multirow{2}{*}{OpenBioLLM-8B} & Time Varying & 3.65 (3.19, 4.36) & 78.33 (77.11, 79.54) \\
 & & TV w/o Fill Strategy & 2.76 (2.44, 3.25) & 75.31 (73.98, 76.67) \\
\midrule 

\multicolumn{5}{c}{\textbf{Prediction Window = 9 Months}} \\
\midrule
\multirow{6}{*}{Machine Learning} 
 & \multirow{2}{*}{Elastic Net LR} & Time Varying & 4.52 (4.05, 5.22) & 76.30 (75.22, 77.32) \\
 & & TV w/o Fill Strategy & 4.41 (3.95, 5.07) & 76.18 (75.09, 77.19) \\
\cmidrule(lr){2-5}
 & \multirow{2}{*}{Random Forest} & Time Varying & 5.23 (4.67, 5.95) & 79.03 (78.11, 79.95) \\
 & & TV w/o Fill Strategy & 5.21 (4.67, 5.90) & 78.94 (78.02, 79.84) \\
\cmidrule(lr){2-5}
 & \multirow{2}{*}{XGBoost} & Time Varying & 5.14 (4.59, 5.89) & 78.37 (77.39, 79.31) \\
 & & TV w/o Fill Strategy & 5.08 (4.53, 5.90) & 78.14 (77.15, 79.06) \\
\midrule
\multirow{4}{*}{\makecell[l]{Masked Language\\Models}} 
 & \multirow{2}{*}{ModernBERT} & Time Varying & 5.27 (4.68, 6.01) & 78.08 (77.08, 79.03) \\
 & & TV w/o Fill Strategy & 5.04 (4.39, 5.70) & 78.18 (76.22, 78.22) \\
\cmidrule(lr){2-5}
 & \multirow{2}{*}{BioClinical ModernBERT} & Time Varying & 5.16 (4.65, 5.84) & 78.72 (77.75, 79.68) \\
 & & TV w/o Fill Strategy & 4.92 (4.41, 5.60) & 76.94 (75.89, 77.97) \\
\midrule
\multirow{4}{*}{\makecell[l]{Large Language\\Models}} 
 & \multirow{2}{*}{Llama-3.1-8B} & Time Varying & 5.19 (4.64, 5.92) & 78.56 (77.60, 79.46) \\
 & & TV w/o Fill Strategy & 5.01 (4.51, 5.68) & 79.10 (78.11, 80.02) \\
\cmidrule(lr){2-5}
 & \multirow{2}{*}{OpenBioLLM-8B} & Time Varying & 4.42 (3.97, 5.03) & 76.68 (75.64, 77.68) \\
 & & TV w/o Fill Strategy & 4.62 (4.17, 5.23) & 79.24 (78.28, 80.15) \\
\midrule

\multicolumn{5}{c}{\textbf{Prediction Window = 12 Months}} \\
\midrule
\multirow{6}{*}{Machine Learning} 
 & \multirow{2}{*}{Elastic Net LR} & Time Varying & 5.72 (5.22, 6.41) & 75.75 (74.74, 76.72) \\
 & & TV w/o Fill Strategy & 5.57 (5.11, 6.21) & 75.59 (74.60, 76.54) \\
\cmidrule(lr){2-5}
 & \multirow{2}{*}{Random Forest} & Time Varying & 6.39 (5.83, 7.06) & 78.39 (77.50, 79.24) \\
 & & TV w/o Fill Strategy & 6.48 (5.92, 7.18) & 78.56 (77.70, 79.41) \\
\cmidrule(lr){2-5}
 & \multirow{2}{*}{XGBoost} & Time Varying & 6.72 (6.06, 7.53) & 78.05 (77.14, 78.92) \\
 & & TV w/o Fill Strategy & 6.58 (5.95, 7.35) & 78.02 (77.12, 78.90) \\
\midrule
\multirow{4}{*}{\makecell[l]{Masked Language\\Models}} 
 & \multirow{2}{*}{ModernBERT} & Time Varying & 6.29 (5.76, 6.97) & 77.84 (76.90, 78.75) \\
 & & TV w/o Fill Strategy & 6.30 (5.78, 6.96) & 78.09 (77.14, 78.97) \\
\cmidrule(lr){2-5}
 & \multirow{2}{*}{BioClinical ModernBERT} & Time Varying & 6.65 (6.03, 7.40) & 77.99 (77.07, 78.86) \\
 & & TV w/o Fill Strategy & 6.39 (5.85, 7.12) & 77.91 (76.98, 78.83) \\
\midrule
\multirow{4}{*}{\makecell[l]{Large Language\\Models}} 
 & \multirow{2}{*}{Llama-3.1-8B} & Time Varying & 5.66 (5.16, 6.32) & 77.30 (76.38, 78.15) \\
 & & TV w/o Fill Strategy & 6.19 (5.61, 6.92) & 76.96 (76.04, 77.89) \\
\cmidrule(lr){2-5}
 & \multirow{2}{*}{OpenBioLLM-8B} & Time Varying & 5.99 (5.49, 6.66) & 78.06 (77.18, 78.93) \\
 & & TV w/o Fill Strategy & 4.37 (4.04, 4.77) & 75.39 (74.49, 76.26) \\
\bottomrule
\end{tabularx}
\end{table*}

\clearpage
\section*{Supplementary Table 3. Performance of time-varying models by risk tier (top 1\% and 5\%) and prediction window}\phantomsection
\label{supp_table_3}

\begin{table*}[!ht]
\scriptsize
\renewcommand{\arraystretch}{0.9}
\setlength{\tabcolsep}{3pt}
\centering

\begin{tabularx}{\textwidth}{l X l l l l l}
\toprule
\textbf{Model Class} & \textbf{Model} & \makecell[c]{\textbf{Risk Group}\\\textbf{Size, P}} & \makecell[c]{\textbf{Sensitivity}\\\textbf{@ top-P, \%}} & \makecell[c]{\textbf{Specificity}\\\textbf{@ top-P, \%}} & \makecell[c]{\textbf{PPV}\\\textbf{@ top-P, \%}} & \makecell[c]{\textbf{O/E Ratio}\\\textbf{@ top-P}} \\
\midrule

\multicolumn{7}{c}{\textbf{Prediction Window = 3 Months}} \\
\midrule
\multirow{6}{*}{Machine Learning} 
 & \multirow{2}{=}{Elastic Net \\ Logistic Regression} & 0.01 & 12.68 (10.35, 15.16) & 99.04 (99.03, 99.05) & 4.07 (3.32, 4.86) & 12.68 (10.35, 15.15) \\
 & & 0.05 & 30.76 (27.26, 34.11) & 95.08 (95.07, 95.09) & 1.97 (1.75, 2.19) & 6.15 (5.45, 6.82) \\ \cmidrule(lr){2-7}
 & \multirow{2}{*}{Random Forest} & 0.01 & 13.41 (10.93, 16.03) & 99.04 (99.03, 99.05) & 4.30 (3.51, 5.14) & 13.41 (10.93, 16.03) \\
 & & 0.05 & 33.09 (29.59, 36.59) & 95.09 (95.08, 95.10) & 2.12 (1.90, 2.35) & 6.62 (5.92, 7.32) \\ \cmidrule(lr){2-7}
 & \multirow{2}{*}{XGBoost} & 0.01 & 12.83 (10.64, 15.45) & 99.04 (99.03, 99.05) & 4.12 (3.41, 4.96) & 12.83 (10.64, 15.45) \\
 & & 0.05 & 32.36 (29.01, 35.71) & 95.09 (95.08, 95.10) & 2.08 (1.86, 2.29) & 6.47 (5.80, 7.14) \\
\midrule
\multirow{4}{*}{\makecell[l]{Masked Language\\Models}} 
 & \multirow{2}{*}{ModernBERT} & 0.01 & 12.24 (9.77, 14.58) & 99.04 (99.03, 99.04) & 3.93 (3.13, 4.68) & 12.25 (9.76, 14.58) \\
 & & 0.05 & 30.17 (26.96, 33.53) & 95.08 (95.07, 95.09) & 1.94 (1.73, 2.15) & 6.04 (5.39, 6.70) \\ \cmidrule(lr){2-7}
 & \multirow{2}{*}{BioClinical ModernBERT} & 0.01 & 14.72 (12.39, 17.35) & 99.04 (99.04, 99.05) & 4.72 (3.97, 5.56) & 14.72 (12.39, 17.34) \\
 & & 0.05 & 32.80 (29.30, 36.30) & 95.09 (95.08, 95.10) & 2.10 (1.88, 2.33) & 6.56 (5.86, 7.26) \\
\midrule
\multirow{4}{*}{\makecell[l]{Large Language\\Models}} 
 & \multirow{2}{*}{Llama-3.1-8B} & 0.01 & 13.85 (11.37, 16.33) & 99.04 (99.03, 99.05) & 4.44 (3.65, 5.24) & 13.85 (11.37, 16.32) \\
 & & 0.05 & 32.51 (29.15, 36.15) & 95.09 (95.08, 95.10) & 2.09 (1.87, 2.32) & 6.50 (5.83, 7.23) \\ \cmidrule(lr){2-7}
 & \multirow{2}{*}{OpenBioLLM-8B} & 0.01 & 13.41 (11.37, 16.62) & 99.04 (99.03, 99.05) & 4.30 (3.65, 5.33) & 13.41 (11.37, 16.61) \\
 & & 0.05 & 31.92 (28.57, 35.57) & 95.09 (95.08, 95.10) & 2.05 (1.83, 2.28) & 6.38 (5.71, 7.11) \\
\midrule 

\multicolumn{7}{c}{\textbf{Prediction Window = 6 Months}} \\
\midrule
\multirow{6}{*}{Machine Learning} 
 & \multirow{2}{=}{Elastic Net \\ Logistic Regression} & 0.01 & 11.78 (10.14, 13.50) & 99.07 (99.06, 99.08) & 7.39 (6.36, 8.46) & 11.78 (10.14, 13.49) \\
 & & 0.05 & 27.59 (25.21, 29.90) & 95.14 (95.13, 95.16) & 3.46 (3.16, 3.75) & 5.52 (5.04, 5.98) \\ \cmidrule(lr){2-7}
 & \multirow{2}{*}{Random Forest} & 0.01 & 12.53 (10.81, 14.24) & 99.07 (99.06, 99.08) & 7.86 (6.78, 8.93) & 12.53 (10.81, 14.24) \\
 & & 0.05 & 29.53 (27.22, 31.99) & 95.15 (95.14, 95.17) & 3.70 (3.41, 4.01) & 5.91 (5.44, 6.40) \\ \cmidrule(lr){2-7}
 & \multirow{2}{*}{XGBoost} & 0.01 & 12.53 (10.81, 14.32) & 99.07 (99.06, 99.08) & 7.86 (6.78, 8.98) & 12.53 (10.81, 14.31) \\
 & & 0.05 & 29.31 (26.92, 31.69) & 95.15 (95.14, 95.17) & 3.68 (3.38, 3.98) & 5.86 (5.38, 6.34) \\
\midrule
\multirow{4}{*}{\makecell[l]{Masked Language\\Models}} 
 & \multirow{2}{*}{ModernBERT} & 0.01 & 12.53 (10.81, 14.32) & 99.07 (99.06, 99.08) & 7.86 (6.78, 8.98) & 12.53 (10.81, 14.31) \\
 & & 0.05 & 30.05 (27.67, 32.44) & 95.16 (95.14, 95.17) & 3.77 (3.47, 4.07) & 6.01 (5.53, 6.49) \\ \cmidrule(lr){2-7}
 & \multirow{2}{*}{BioClinical ModernBERT} & 0.01 & 13.27 (11.41, 14.91) & 99.08 (99.07, 99.09) & 8.33 (7.15, 9.35) & 13.27 (11.40, 14.91) \\
 & & 0.05 & 29.53 (27.07, 31.99) & 95.15 (95.14, 95.17) & 3.70 (3.39, 4.01) & 5.91 (5.41, 6.40) \\
\midrule
\multirow{4}{*}{\makecell[l]{Large Language\\Models}} 
 & \multirow{2}{*}{Llama-3.1-8B} & 0.01 & 12.83 (11.04, 14.47) & 99.07 (99.06, 99.08) & 8.04 (6.92, 9.07) & 12.83 (11.03, 14.46) \\
 & & 0.05 & 30.95 (28.49, 33.41) & 95.16 (95.15, 95.18) & 3.88 (3.57, 4.19) & 6.19 (5.70, 6.68) \\ \cmidrule(lr){2-7}
 & \multirow{2}{*}{OpenBioLLM-8B} & 0.01 & 11.63 (10.37, 14.10) & 99.07 (99.06, 99.08) & 7.30 (6.50, 8.84) & 11.63 (10.36, 14.09) \\
 & & 0.05 & 30.28 (28.11, 32.96) & 95.16 (95.15, 95.18) & 3.80 (3.53, 4.13) & 6.06 (5.62, 6.59) \\
\midrule 

\multicolumn{7}{c}{\textbf{Prediction Window = 9 Months}} \\
\midrule
\multirow{6}{*}{Machine Learning} 
 & \multirow{2}{=}{Elastic Net \\ Logistic Regression} & 0.01 & 10.71 (9.34, 12.02) & 99.09 (99.08, 99.10) & 9.87 (8.60, 11.08) & 10.71 (9.33, 12.02) \\
 & & 0.05 & 26.69 (24.71, 28.61) & 95.20 (95.18, 95.22) & 4.92 (4.55, 5.27) & 5.34 (4.94, 5.72) \\ \cmidrule(lr){2-7}
 & \multirow{2}{*}{Random Forest} & 0.01 & 12.23 (10.86, 13.65) & 99.10 (99.09, 99.12) & 11.27 (10.00, 12.58) & 12.23 (10.85, 13.64) \\
 & & 0.05 & 29.17 (27.19, 31.05) & 95.22 (95.21, 95.24) & 5.38 (5.01, 5.72) & 5.83 (5.44, 6.21) \\ \cmidrule(lr){2-7}
 & \multirow{2}{*}{XGBoost} & 0.01 & 11.97 (10.55, 13.39) & 99.10 (99.09, 99.11) & 11.04 (9.72, 12.34) & 11.97 (10.55, 13.39) \\
 & & 0.05 & 28.92 (26.89, 30.95) & 95.22 (95.20, 95.24) & 5.33 (4.96, 5.71) & 5.78 (5.38, 6.19) \\
\midrule
\multirow{4}{*}{\makecell[l]{Masked Language\\Models}} 
 & \multirow{2}{*}{ModernBERT} & 0.01 & 12.08 (10.65, 13.50) & 99.10 (99.09, 99.12) & 11.13 (9.82, 12.44) & 12.08 (10.65, 13.49) \\
 & & 0.05 & 29.98 (27.90, 32.01) & 95.23 (95.21, 95.25) & 5.53 (5.14, 5.90) & 6.00 (5.58, 6.40) \\ \cmidrule(lr){2-7}
 & \multirow{2}{*}{BioClinical ModernBERT} & 0.01 & 11.52 (10.25, 12.99) & 99.10 (99.09, 99.11) & 10.62 (9.44, 11.97) & 11.52 (10.24, 12.98) \\
 & & 0.05 & 30.44 (28.36, 32.32) & 95.24 (95.22, 95.25) & 5.61 (5.23, 5.96) & 6.09 (5.67, 6.46) \\
\midrule
\multirow{4}{*}{\makecell[l]{Large Language\\Models}} 
 & \multirow{2}{*}{Llama-3.1-8B} & 0.01 & 12.38 (11.06, 13.90) & 99.11 (99.09, 99.12) & 11.41 (10.19, 12.81) & 12.38 (11.06, 13.90) \\
 & & 0.05 & 29.07 (27.09, 30.95) & 95.22 (95.21, 95.24) & 5.36 (4.99, 5.71) & 5.81 (5.42, 6.19) \\ \cmidrule(lr){2-7}
 & \multirow{2}{*}{OpenBioLLM-8B} & 0.01 & 10.10 (9.28, 12.03) & 99.08 (99.08, 99.10) & 9.31 (8.56, 11.08) & 10.10 (9.28, 12.02) \\
 & & 0.05 & 27.45 (26.18, 30.19) & 95.21 (95.20, 95.23) & 5.06 (4.83, 5.56) & 5.49 (5.24, 6.04) \\
\midrule 

\multicolumn{7}{c}{\textbf{Prediction Window = 12 Months}} \\
\midrule
\multirow{6}{*}{Machine Learning} 
 & \multirow{2}{=}{Elastic Net \\ Logistic Regression} & 0.01 & 10.90 (9.76, 11.96) & 99.12 (99.11, 99.13) & 13.00 (11.64, 14.26) & 10.90 (9.76, 11.96) \\
 & & 0.05 & 26.39 (24.78, 28.00) & 95.26 (95.24, 95.28) & 6.30 (5.91, 6.68) & 5.28 (4.96, 5.60) \\ \cmidrule(lr){2-7}
 & \multirow{2}{*}{Random Forest} & 0.01 & 11.29 (10.12, 12.47) & 99.12 (99.11, 99.14) & 13.47 (12.06, 14.87) & 11.30 (10.11, 12.47) \\
 & & 0.05 & 28.12 (26.47, 29.84) & 95.28 (95.26, 95.30) & 6.71 (6.31, 7.12) & 5.62 (5.29, 5.97) \\ \cmidrule(lr){2-7}
 & \multirow{2}{*}{XGBoost} & 0.01 & 11.57 (10.43, 12.78) & 99.13 (99.11, 99.14) & 13.80 (12.44, 15.24) & 11.57 (10.43, 12.78) \\
 & & 0.05 & 27.88 (26.20, 29.57) & 95.28 (95.26, 95.30) & 6.65 (6.25, 7.05) & 5.58 (5.24, 5.91) \\
\midrule
\multirow{4}{*}{\makecell[l]{Masked Language\\Models}} 
 & \multirow{2}{*}{ModernBERT} & 0.01 & 11.10 (10.00, 12.24) & 99.12 (99.11, 99.14) & 13.24 (11.92, 14.59) & 11.10 (10.00, 12.23) \\
 & & 0.05 & 29.69 (27.88, 31.29) & 95.30 (95.28, 95.32) & 7.08 (6.65, 7.46) & 5.94 (5.58, 6.26) \\ \cmidrule(lr){2-7}
 & \multirow{2}{*}{BioClinical ModernBERT} & 0.01 & 11.41 (10.24, 12.59) & 99.13 (99.11, 99.14) & 13.61 (12.20, 15.01) & 11.41 (10.23, 12.58) \\
 & & 0.05 & 29.18 (27.41, 30.78) & 95.29 (95.27, 95.31) & 6.96 (6.54, 7.34) & 5.84 (5.48, 6.16) \\
\midrule
\multirow{4}{*}{\makecell[l]{Large Language\\Models}} 
 & \multirow{2}{*}{Llama-3.1-8B} & 0.01 & 9.76 (8.71, 10.90) & 99.11 (99.09, 99.12) & 11.65 (10.38, 13.00) & 9.77 (8.70, 10.90) \\
 & & 0.05 & 26.67 (25.14, 28.47) & 95.26 (95.24, 95.28) & 6.36 (6.00, 6.79) & 5.33 (5.03, 5.69) \\ \cmidrule(lr){2-7}
 & \multirow{2}{*}{OpenBioLLM-8B} & 0.01 & 10.63 (10.04, 12.39) & 99.12 (99.11, 99.14) & 12.68 (11.97, 14.77) & 10.63 (10.04, 12.39) \\
 & & 0.05 & 28.00 (26.71, 30.04) & 95.28 (95.26, 95.30) & 6.68 (6.37, 7.16) & 5.60 (5.34, 6.01) \\
\bottomrule
\end{tabularx}
\end{table*}

\clearpage
\section*{Supplementary Table 4. Model performance by predictor set and prediction window}\phantomsection
\label{supp_table_4}

{
\scriptsize
\renewcommand{\arraystretch}{0.92}
\setlength{\tabcolsep}{4pt}

\begin{xltabular}{\textwidth}{l X l l l}
\toprule
\textbf{Model Class} & \textbf{Model} & \textbf{Feature Group} & \textbf{PR-AUC, \%} & \textbf{ROC AUC, \%} \\
\midrule
\endfirsthead

\midrule
\multicolumn{5}{c}{{Supplementary Table 4 -- continued}} \\
\midrule
\endhead

\midrule
\multicolumn{5}{r}{\textit{Continued on next page}} \\
\endfoot

\bottomrule
\endlastfoot

\multicolumn{5}{c}{\textbf{Prediction Window = 3 Months}} \\
\midrule
\multirow{9}{*}{Machine Learning}
 & \multirow{3}{=}{Elastic Net \\ Logistic Regression} & Demographic & 0.6 (0.55, 0.69) & 68.31 (66.44, 70.08) \\
 & & Demographic + Code & 1.6 (1.31, 2.11) & 76.06 (74.28, 77.81) \\
 & & Demographic + Code + SBFH & 1.87 (1.54, 2.45) & 76.71 (74.79, 78.46) \\ \cmidrule(lr){2-5}
 & \multirow{3}{*}{Random Forest} & Demographic & 0.6 (0.55, 0.68) & 68.77 (66.99, 70.45) \\
 & & Demographic + Code & 1.66 (1.36, 2.14) & 77.67 (75.93, 79.31) \\
 & & Demographic + Code + SBFH & 2.11 (1.76, 2.68) & 79.27 (77.54, 80.94) \\ \cmidrule(lr){2-5}
 & \multirow{3}{*}{XGBoost} & Demographic & 0.65 (0.58, 0.80) & 69.79 (68.00, 71.49) \\
 & & Demographic + Code & 2.22 (1.63, 3.14) & 78.15 (76.46, 79.75) \\
 & & Demographic + Code + SBFH & 2.22 (1.83, 2.91) & 79.73 (77.95, 81.37) \\
\midrule
\multirow{6}{*}{\makecell[l]{Masked Language\\Models}}
 & \multirow{3}{*}{ModernBERT} & Demographic & 0.62 (0.56, 0.71) & 68.80 (66.93, 70.61) \\
 & & Demographic + Code & 1.78 (1.46, 2.35) & 77.97 (76.20, 79.64) \\
 & & Demographic + Code + SBFH & 2.39 (1.80, 3.34) & 77.28 (75.39, 79.00) \\ \cmidrule(lr){2-5}
 & \multirow{3}{*}{BioClinical ModernBERT} & Demographic & 0.61 (0.55, 0.72) & 67.40 (65.49, 69.29) \\
 & & Demographic + Code & 1.60 (1.28, 2.22) & 75.71 (73.87, 77.43) \\
 & & Demographic + Code + SBFH & 2.34 (1.91, 3.09) & 78.78 (76.97, 80.53) \\
\midrule
\multirow{6}{*}{\makecell[l]{Large Language\\Models}}
 & \multirow{3}{*}{Llama-3.1-8B} & Demographic & 0.58 (0.52, 0.72) & 66.36 (64.43, 68.22) \\
 & & Demographic + Code & 1.42 (1.18, 1.93) & 75.44 (73.54, 77.18) \\
 & & Demographic + Code + SBFH & 2.16 (1.77, 2.76) & 79.12 (77.43, 80.79) \\ \cmidrule(lr){2-5}
 & \multirow{3}{*}{OpenBioLLM-8B} & Demographic & 0.65 (0.55, 1.01) & 68.42 (66.66, 70.13) \\
 & & Demographic + Code & 1.26 (1.08, 1.52) & 76.10 (74.40, 77.83) \\
 & & Demographic + Code + SBFH & 2.32 (1.81, 3.10) & 78.31 (76.49, 79.95) \\
\midrule 

\multicolumn{5}{c}{\textbf{Prediction Window = 6 Months}} \\
\midrule
\multirow{9}{*}{Machine Learning}
 & \multirow{3}{=}{Elastic Net \\ Logistic Regression} & Demographic & 1.27 (1.16, 1.49) & 67.15 (65.73, 68.51) \\
 & & Demographic + Code & 2.76 (2.38, 3.40) & 74.96 (73.67, 76.27) \\
 & & Demographic + Code + SBFH & 3.28 (2.83, 3.97) & 74.83 (73.46, 76.18) \\ \cmidrule(lr){2-5}
 & \multirow{3}{*}{Random Forest} & Demographic & 1.24 (1.16, 1.37) & 68.22 (66.88, 69.60) \\
 & & Demographic + Code & 2.98 (2.56, 3.67) & 76.41 (75.20, 77.65) \\
 & & Demographic + Code + SBFH & 3.86 (3.32, 4.61) & 77.98 (76.77, 79.24) \\ \cmidrule(lr){2-5}
 & \multirow{3}{*}{XGBoost} & Demographic & 1.34 (1.24, 1.52) & 69.87 (68.57, 71.17) \\
 & & Demographic + Code & 2.90 (2.53, 3.47) & 76.69 (75.47, 77.91) \\
 & & Demographic + Code + SBFH & 4.13 (3.48, 5.01) & 78.45 (77.20, 79.67) \\
\midrule
\multirow{6}{*}{\makecell[l]{Masked Language\\Models}}
 & \multirow{3}{*}{ModernBERT} & Demographic & 1.29 (1.19, 1.45) & 68.66 (67.30, 69.94) \\
 & & Demographic + Code & 2.92 (2.53, 3.50) & 76.54 (75.31, 77.78) \\
 & & Demographic + Code + SBFH & 3.54 (3.07, 4.14) & 77.49 (76.23, 78.73) \\ \cmidrule(lr){2-5}
 & \multirow{3}{*}{BioClinical ModernBERT} & Demographic & 1.36 (1.24, 1.54) & 68.87 (67.54, 70.16) \\
 & & Demographic + Code & 2.82 (2.47, 3.36) & 76.02 (74.75, 77.32) \\
 & & Demographic + Code + SBFH & 3.58 (3.12, 4.21) & 76.60 (75.33, 77.93) \\
\midrule
\multirow{6}{*}{\makecell[l]{Large Language\\Models}}
 & \multirow{3}{*}{Llama-3.1-8B} & Demographic & 1.16 (1.07, 1.33) & 65.95 (64.53, 67.36) \\
 & & Demographic + Code & 2.60 (2.31, 3.05) & 76.28 (75.04, 77.50) \\
 & & Demographic + Code + SBFH & 4.12 (3.52, 4.98) & 78.43 (77.21, 79.66) \\ \cmidrule(lr){2-5}
 & \multirow{3}{*}{OpenBioLLM-8B} & Demographic & 1.26 (1.16, 1.41) & 68.05 (66.71, 69.31) \\
 & & Demographic + Code & 2.00 (1.79, 2.26) & 73.97 (72.65, 75.19) \\
 & & Demographic + Code + SBFH & 3.65 (3.19, 4.36) & 78.33 (77.11, 79.54) \\
\midrule 

\multicolumn{5}{c}{\textbf{Prediction Window = 9 Months}} \\
\midrule
\multirow{9}{*}{Machine Learning}
 & \multirow{3}{=}{Elastic Net \\ Logistic Regression} & Demographic & 1.82 (1.71, 1.95) & 67.88 (66.71, 69.00) \\
 & & Demographic + Code & 3.93 (3.53, 4.51) & 76.11 (75.11, 77.10) \\
 & & Demographic + Code + SBFH & 4.52 (4.05, 5.22) & 76.30 (75.22, 77.32) \\ \cmidrule(lr){2-5}
 & \multirow{3}{*}{Random Forest} & Demographic & 1.86 (1.76, 1.99) & 70.28 (69.24, 71.20) \\
 & & Demographic + Code & 4.27 (3.86, 4.87) & 78.03 (77.11, 78.96) \\
 & & Demographic + Code + SBFH & 5.23 (4.67, 5.95) & 79.03 (78.11, 79.95) \\ \cmidrule(lr){2-5}
 & \multirow{3}{*}{XGBoost} & Demographic & 1.97 (1.86, 2.12) & 71.27 (70.24, 72.23) \\
 & & Demographic + Code & 3.97 (3.57, 4.56) & 77.60 (76.68, 78.50) \\
 & & Demographic + Code + SBFH & 5.14 (4.59, 5.89) & 78.37 (77.39, 79.31) \\
\midrule
\multirow{6}{*}{\makecell[l]{Masked Language\\Models}}
 & \multirow{3}{*}{ModernBERT} & Demographic & 1.86 (1.75, 1.99) & 68.41 (67.29, 69.47) \\
 & & Demographic + Code & 4.60 (4.17, 5.16) & 78.93 (77.95, 79.89) \\
 & & Demographic + Code + SBFH & 5.27 (4.68, 6.01) & 78.08 (77.08, 79.03) \\ \cmidrule(lr){2-5}
 & \multirow{3}{*}{BioClinical ModernBERT} & Demographic & 1.90 (1.79, 2.05) & 69.43 (68.28, 70.45) \\
 & & Demographic + Code & 4.31 (3.86, 4.91) & 77.84 (76.89, 78.79) \\
 & & Demographic + Code + SBFH & 5.16 (4.65, 5.84) & 78.72 (77.75, 79.68) \\
\midrule
\multirow{6}{*}{\makecell[l]{Large Language\\Models}}
 & \multirow{3}{*}{Llama-3.1-8B} & Demographic & 1.74 (1.64, 1.88) & 67.27 (66.13, 68.32) \\
 & & Demographic + Code & 2.20 (2.04, 2.40) & 70.35 (69.31, 71.44) \\
 & & Demographic + Code + SBFH & 5.19 (4.64, 5.92) & 78.56 (77.60, 79.46) \\ \cmidrule(lr){2-5}
 & \multirow{3}{*}{OpenBioLLM-8B} & Demographic & 1.80 (1.71, 1.99) & 69.31 (68.27, 70.29) \\
 & & Demographic + Code & 3.09 (2.85, 3.47) & 75.15 (74.12, 76.14) \\
 & & Demographic + Code + SBFH & 4.42 (3.97, 5.03) & 76.68 (75.64, 77.68) \\
\midrule 

\multicolumn{5}{c}{\textbf{Prediction Window = 12 Months}} \\
\midrule
\multirow{9}{*}{Machine Learning}
 & \multirow{3}{=}{Elastic Net \\ Logistic Regression} & Demographic & 2.35 (2.23, 2.51) & 68.41 (67.38, 69.44) \\
 & & Demographic + Code & 5.16 (4.70, 5.80) & 75.94 (74.95, 76.86) \\
 & & Demographic + Code + SBFH & 5.72 (5.22, 6.41) & 75.75 (74.74, 76.72) \\ \cmidrule(lr){2-5}
 & \multirow{3}{*}{Random Forest} & Demographic & 2.40 (2.28, 2.57) & 69.99 (69.04, 70.94) \\
 & & Demographic + Code & 5.52 (5.03, 6.12) & 77.86 (76.97, 78.69) \\
 & & Demographic + Code + SBFH & 6.39 (5.83, 7.06) & 78.39 (77.50, 79.24) \\ \cmidrule(lr){2-5}
 & \multirow{3}{*}{XGBoost} & Demographic & 2.49 (2.37, 2.65) & 70.63 (69.72, 71.53) \\
 & & Demographic + Code & 5.43 (4.93, 6.08) & 77.31 (76.43, 78.14) \\
 & & Demographic + Code + SBFH & 6.72 (6.06, 7.53) & 78.05 (77.14, 78.92) \\
\midrule
\multirow{6}{*}{\makecell[l]{Masked Language\\Models}}
 & \multirow{3}{*}{ModernBERT} & Demographic & 2.42 (2.29, 2.57) & 69.05 (68.05, 70.08) \\
 & & Demographic + Code & 4.88 (4.44, 5.42) & 76.18 (75.26, 77.06) \\
 & & Demographic + Code + SBFH & 6.29 (5.76, 6.97) & 77.84 (76.90, 78.75) \\ \cmidrule(lr){2-5}
 & \multirow{3}{*}{BioClinical ModernBERT} & Demographic & 2.44 (2.31, 2.61) & 69.11 (68.13, 70.12) \\
 & & Demographic + Code & 5.57 (5.06, 6.19) & 77.27 (76.33, 78.12) \\
 & & Demographic + Code + SBFH & 6.65 (6.03, 7.40) & 77.99 (77.07, 78.86) \\
\midrule
\multirow{6}{*}{\makecell[l]{Large Language\\Models}}
 & \multirow{3}{*}{Llama-3.1-8B} & Demographic & 2.18 (2.08, 2.30) & 67.59 (66.60, 68.58) \\
 & & Demographic + Code & 4.73 (4.35, 5.19) & 75.39 (74.41, 76.30) \\
 & & Demographic + Code + SBFH & 5.66 (5.16, 6.32) & 77.30 (76.38, 78.15) \\ \cmidrule(lr){2-5}
 & \multirow{3}{*}{OpenBioLLM-8B} & Demographic & 2.49 (2.37, 2.63) & 71.03 (70.12, 71.91) \\
 & & Demographic + Code & 4.78 (4.40, 5.22) & 76.40 (75.47, 77.30) \\
 & & Demographic + Code + SBFH & 5.99 (5.49, 6.66) & 78.06 (77.18, 78.93) \\
\end{xltabular}
}

\clearpage
\section*{Supplementary Figure 2. Predictor importance estimated using SHAP values}\phantomsection
\label{supp_fig_2}

SHAP (SHapley Additive exPlanations) values for top 100 positive (A) and negative (B) predictors across all three models (Elastic Net Logistic Regression [ENL], Random Forest [RF], and XGBoost) and four prediction windows (3, 6, 9, and 12 months). Columns represent predictors shown "as is" with time-stamped variants preserved (e.g., PTSD\_2016\_Q1 and PTSD\_2016\_H1 are displayed as separate features), reflecting the actual feature space used by each model-window combination. Predictors are grouped by category with visual separators and prefixed as follows: [D] Demographics, [D-MH] Mental Health Disorders, [D-PH] Physical Health Disorders, [D-SA] Substance Abuse Disorders, [U] Service Utilization, [P] Procedures, and [S] Social and Behavioral Factors of Health.

\begin{figure}[!htbp]
    \centering
    \includegraphics[width=1\linewidth]{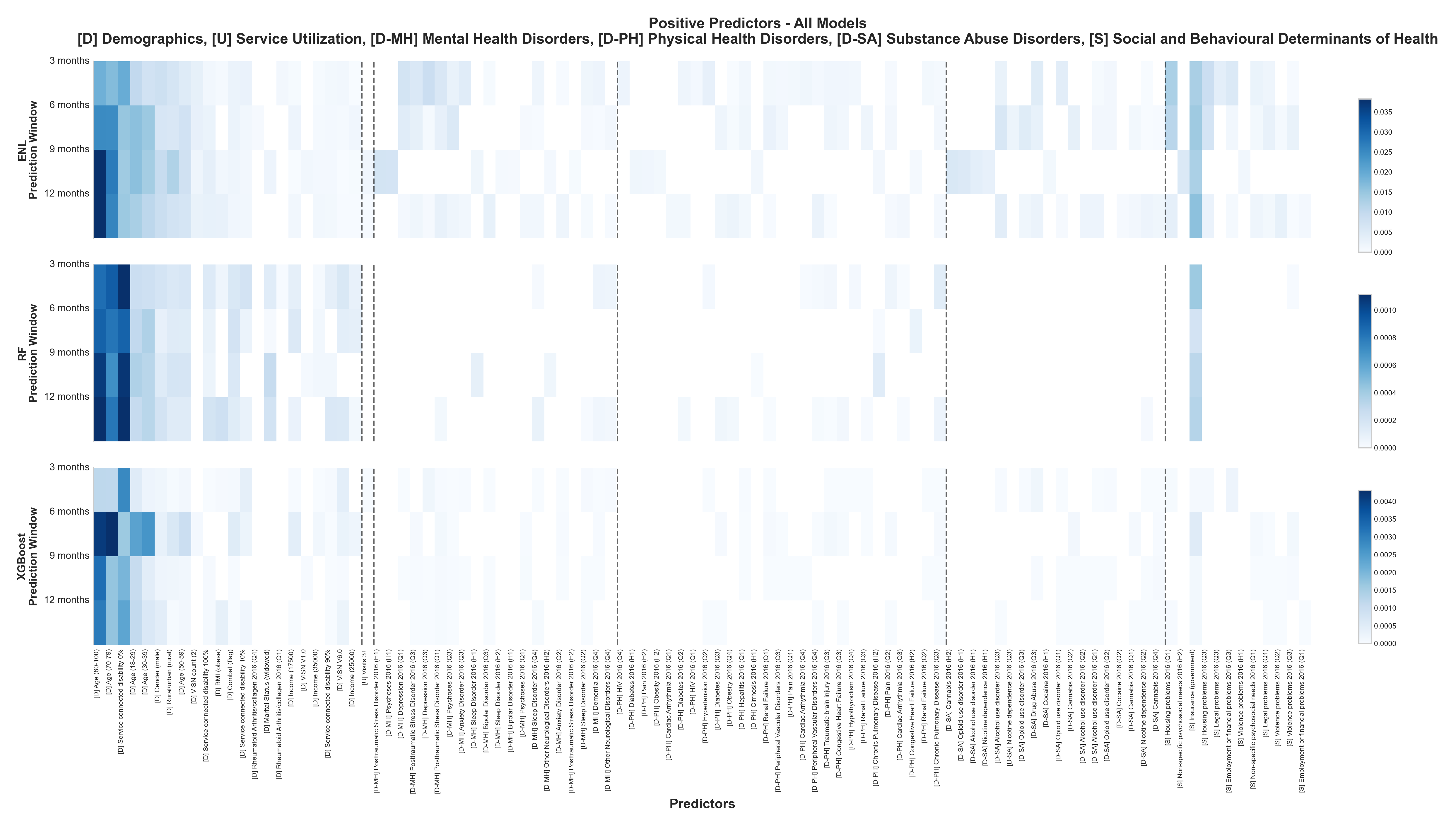}
    (A)
\end{figure}

\begin{figure}[!htbp]
    \centering
    \includegraphics[width=1\linewidth]{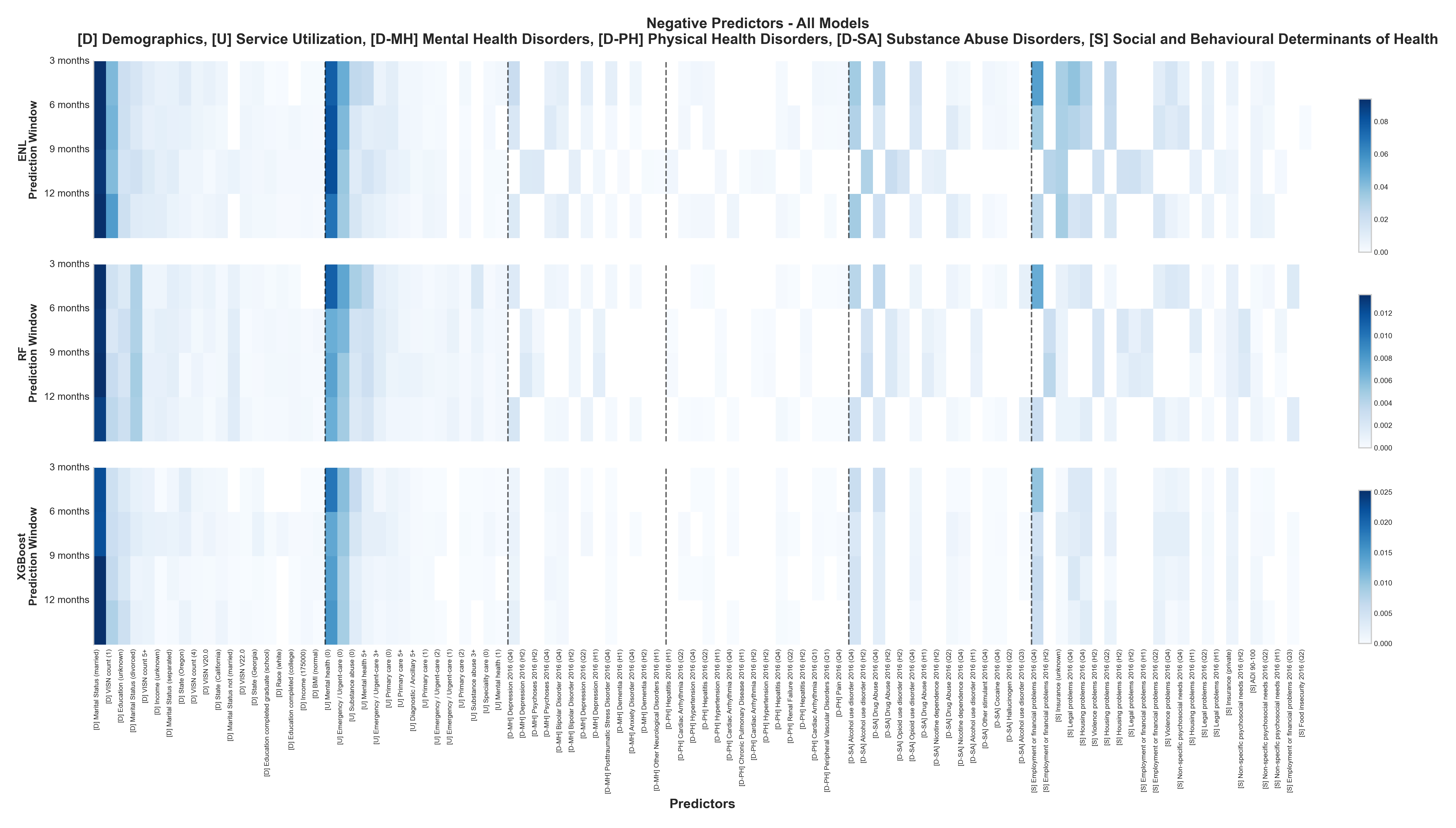}
    (B)
\end{figure}

\clearpage
\section*{Supplementary Table 5. Subgroup-specific PR-AUC by prediction window, model, and demographic subgroup}\phantomsection
\label{supp_table_5}

PR-AUC for the best models from each class at each of the prediction windows (3-, 6-, 9-, and 12-months), stratified by age, race, and ethnicity. * Estimates did not meet pre-specified reliability criteria (fewer than 20 positive or negative cases, or 95\% CI width > 0.12).

{
\scriptsize
\renewcommand{\arraystretch}{1.1}
\setlength{\tabcolsep}{3pt}

\begin{xltabular}{\textwidth}{l X l l l l l}
\toprule
\textbf{Subgroup} & \textbf{Level} & \textbf{N} & \textbf{Events} & 
\makecell{\textbf{Model 1}\\\textbf{PR-AUC, \%}} & 
\makecell{\textbf{Model 2}\\\textbf{PR-AUC, \%}} & 
\makecell{\textbf{Model 3}\\\textbf{PR-AUC, \%}} \\
\midrule
\endfirsthead

\midrule
\multicolumn{7}{c}{{Supplementary Table 5 -- continued}} \\
\midrule
\endhead

\midrule
\multicolumn{7}{r}{{Continued on next page}} \\
\endfoot

\bottomrule
\endlastfoot

\multicolumn{7}{c}{\textbf{Prediction Window = 3 Months}} \\
\midrule
 & & & & \textbf{XGBoost} & \textbf{ModernBERT} & \textbf{Llama} \\
\midrule
\multirow{7}{*}{Age}
 & 18--29 & 7195 & 63 & 4.78 (2.98, 9.38) & 5.52 (2.69, 12.59) & 4.15 (2.79, 7.36) \\
 & 30--39 & 17206 & 114 & 4.06 (2.56, 7.74) & 4.63 (2.45, 8.94) & 4.01 (2.63, 7.05) \\
 & 40--49 & 18605 & 83 & 2.76 (1.81, 4.69) & 2.75 (1.56, 6.35) & 2.54 (1.65, 4.35) \\
 & 50--59 & 29066 & 178 & 3.19 (2.30, 5.34) & 3.00 (2.13, 4.88) & 2.98 (2.11, 5.01) \\
 & 60--69 & 63966 & 155 & 1.96 (1.24, 3.59) & 2.21 (1.09, 4.31) & 1.72 (1.10, 3.26) \\
 & 70--79 & 45844 & 62 & 0.63 (0.43, 1.14) & 0.68 (0.35, 2.21) & 0.50 (0.35, 0.84) \\
 & 80--100 & 31939 & 31 & 0.52 (0.28, 1.14) & 0.36 (0.21, 0.77) & 0.41 (0.22, 0.94) \\
\midrule
\multirow{3}{*}{Ethnicity}
 & Hispanic & 13344 & 52 & 3.80 (1.67, 9.24) & 4.26 (1.38, 9.48) & 3.95 (1.76, 10.34) \\
 & Not Hispanic & 194335 & 621 & 2.18 (1.78, 2.88) & 2.32 (1.73, 3.27) & 2.08 (1.70, 2.71) \\
 & Unknown & 6142 & 13 & 2.68 (0.46, 16.35)* & 7.43 (0.39, 30.51)* & 4.08 (0.49, 19.97)* \\
\midrule
\multirow{5}{*}{Race}
 & American Indian & 1555 & 9 & 11.31 (2.41, 39.16)* & 25.56 (6.36, 56.06)* & 24.49 (4.66, 54.27)* \\
 & Black & 33902 & 223 & 3.21 (2.28, 5.14) & 2.78 (1.81, 4.42) & 2.87 (2.01, 4.64) \\
 & Native Hawaiian & 1796 & 7 & 8.83 (2.05, 34.73)* & 6.35 (1.90, 23.47)* & 9.43 (2.28, 31.64)* \\
 & Unknown & 14526 & 39 & 1.92 (0.86, 6.34) & 1.58 (0.67, 5.52) & 1.86 (0.83, 6.72) \\
 & White & 159932 & 408 & 1.98 (1.52, 2.86) & 2.49 (1.63, 3.98) & 1.95 (1.52, 2.69) \\
\midrule 

\multicolumn{7}{c}{\textbf{Prediction Window = 6 Months}} \\
\midrule
 & & & & \textbf{XGBoost} & \textbf{Clin. ModernBERT} & \textbf{Llama} \\
\midrule
\multirow{7}{*}{Age}
 & 18--29 & 7137 & 107 & 7.43 (4.78, 12.41) & 6.07 (4.17, 10.03) & 7.56 (4.56, 12.59) \\
 & 30--39 & 16885 & 220 & 6.13 (4.32, 9.27) & 5.25 (3.93, 7.78) & 5.25 (3.84, 7.96) \\
 & 40--49 & 18571 & 183 & 8.15 (5.26, 12.47) & 6.50 (4.24, 10.71) & 6.81 (4.46, 10.54) \\
 & 50--59 & 29023 & 307 & 4.29 (3.40, 6.02) & 3.54 (2.84, 4.73) & 4.08 (3.14, 5.95) \\
 & 60--69 & 64276 & 339 & 4.36 (3.12, 6.27) & 3.22 (2.49, 4.32) & 3.98 (2.92, 6.03) \\
 & 70--79 & 45969 & 125 & 2.12 (1.47, 3.84) & 3.05 (1.74, 6.18) & 3.81 (1.81, 8.17) \\
 & 80--100 & 31960 & 60 & 2.59 (1.05, 7.36) & 0.93 (0.51, 2.19) & 1.04 (0.67, 1.92) \\
\midrule
\multirow{3}{*}{Ethnicity}
 & Hispanic & 13339 & 93 & 5.19 (3.06, 9.79) & 5.14 (2.98, 10.07) & 5.35 (3.07, 10.63) \\
 & Not Hispanic & 194320 & 1217 & 4.14 (3.47, 5.09) & 3.55 (3.07, 4.20) & 4.10 (3.52, 5.04) \\
 & Unknown & 6162 & 31 & 3.16 (1.50, 8.45) & 4.06 (0.89, 13.99) & 5.17 (1.07, 15.33) \\
\midrule
\multirow{6}{*}{Race}
 & American Indian & 1617 & 17 & 10.89 (2.10, 28.08)* & 11.07 (2.12, 28.36)* & 4.87 (2.03, 16.20)* \\
 & Asian & 2075 & 12 & 6.94 (1.78, 27.70)* & 11.59 (1.38, 32.42)* & 7.95 (1.92, 29.16)* \\
 & Black & 33879 & 409 & 5.10 (3.99, 7.08) & 4.09 (3.33, 5.32) & 4.53 (3.68, 6.08) \\
 & Native Hawaiian & 1954 & 10 & 13.46 (1.02, 37.51)* & 3.58 (0.77, 16.22)* & 8.65 (1.01, 33.57)* \\
 & Unknown & 14460 & 91 & 3.31 (2.30, 5.38) & 3.80 (2.43, 7.30)* & 4.12 (2.53, 8.32)* \\
 & White & 159836 & 802 & 4.12 (3.36, 5.27) & 3.56 (2.96, 4.41) & 4.23 (3.38, 5.56) \\
\midrule 

\multicolumn{7}{c}{\textbf{Prediction Window = 9 Months}} \\
\midrule
 & & & & \textbf{Random Forest} & \textbf{ModernBERT} & \textbf{Llama} \\
\midrule
\multirow{7}{*}{Age}
 & 18--29 & 7105 & 174 & 10.30 (7.16, 14.70) & 8.92 (6.59, 12.89) & 8.38 (6.48, 11.89) \\
 & 30--39 & 16922 & 335 & 6.83 (5.62, 8.86) & 7.39 (5.92, 9.59) & 7.15 (5.83, 9.16) \\
 & 40--49 & 18618 & 268 & 6.63 (4.96, 9.29) & 6.75 (5.23, 9.48) & 6.94 (5.27, 9.54) \\
 & 50--59 & 29241 & 436 & 8.24 (6.55, 10.40) & 8.74 (7.06, 11.09) & 8.96 (7.06, 11.40) \\
 & 60--69 & 63979 & 513 & 4.19 (3.54, 5.23) & 4.64 (3.83, 5.95) & 4.62 (3.86, 5.94) \\
 & 70--79 & 45875 & 178 & 2.53 (1.55, 4.35) & 2.21 (1.48, 4.21) & 2.19 (1.58, 3.78) \\
 & 80--100 & 32081 & 67 & 2.20 (0.76, 6.55) & 1.78 (0.55, 6.13) & 4.34 (1.09, 11.02) \\
\midrule
\multirow{3}{*}{Ethnicity}
 & Hispanic & 13148 & 163 & 6.40 (4.18, 12.77) & 8.00 (4.03, 18.94) & 10.44 (4.45, 21.34) \\
 & Not Hispanic & 194631 & 1773 & 13.63 (4.06, 31.10) & 8.34 (3.96, 20.12) & 8.87 (3.71, 23.71) \\
 & Unknown & 6042 & 35 & 6.97 (6.03, 8.40) & 7.31 (6.20, 9.10) & 7.42 (6.31, 9.15) \\
\midrule
\multirow{6}{*}{Race}
 & American Indian & 1682 & 32 & 4.71 (3.43, 7.45) & 5.01 (3.54, 7.83)* & 5.15 (3.53, 8.49)* \\
 & Asian & 2133 & 21 & 2.50 (1.58, 5.89)* & 2.54 (1.62, 5.86)* & 3.21 (1.79, 8.58)* \\
 & Black & 34121 & 623 & 5.35 (4.76, 6.17) & 5.34 (4.75, 6.11) & 5.34 (4.74, 6.15) \\
 & Native Hawaiian & 1807 & 21 & 3.97 (2.23, 10.42) & 6.27 (2.05, 14.87) & 3.08 (1.71, 8.03) \\
 & Unknown & 14488 & 128 & 3.38 (2.45, 5.65) & 3.37 (2.39, 5.49)* & 3.58 (2.50, 6.54) \\
 & White & 159590 & 1146 & 5.17 (4.40, 6.27) & 4.92 (4.19, 5.88) & 4.81 (4.11, 5.78) \\
\midrule 

\multicolumn{7}{c}{\textbf{Prediction Window = 12 Months}} \\
\midrule
 & & & & \textbf{XGBoost} & \textbf{Clin. ModernBERT} & \textbf{OpenBio} \\
\midrule
\multirow{7}{*}{Age}
 & 18--29 & 7138 & 202 & 8.77 (6.29, 12.37) & 8.97 (6.90, 12.55) & 9.57 (7.19, 13.31) \\
 & 30--39 & 16921 & 397 & 9.95 (7.80, 12.61) & 8.12 (6.67, 10.22) & 9.73 (7.95, 12.28) \\
 & 40--49 & 18743 & 341 & 8.77 (7.01, 11.35) & 8.47 (6.85, 11.03) & 9.14 (7.30, 11.77) \\
 & 50--59 & 28788 & 582 & 9.74 (8.27, 11.93) & 8.92 (7.72, 10.89) & 9.40 (8.08, 11.29) \\
 & 60--69 & 63785 & 660 & 6.17 (5.08, 7.83) & 6.74 (5.39, 8.54) & 6.39 (5.18, 7.99) \\
 & 70--79 & 46107 & 259 & 2.92 (2.15, 4.59) & 2.67 (2.03, 3.79) & 3.03 (2.29, 4.30) \\
 & 80--100 & 32339 & 109 & 2.67 (1.29, 5.29) & 2.96 (1.12, 6.30) & 2.18 (1.31, 4.67) \\
\midrule
\multirow{3}{*}{Ethnicity}
 & Hispanic & 13434 & 198 & 6.04 (4.14, 9.13) & 5.33 (3.95, 8.20) & 4.75 (3.72, 6.74) \\
 & Not Hispanic & 194328 & 2298 & 6.90 (6.18, 7.77) & 6.91 (6.20, 7.76) & 6.21 (5.62, 6.94) \\
 & Unknown & 6059 & 54 & 4.02 (2.49, 8.43) & 3.73 (2.28, 7.62) & 3.82 (2.44, 7.07) \\
\midrule
\multirow{6}{*}{Race}
 & American Indian & 1535 & 27 & 5.96 (2.99, 15.77)* & 5.43 (3.44, 12.51)* & 6.23 (3.01, 16.54)* \\
 & Asian & 2076 & 14 & 7.72 (3.18, 18.81)* & 5.74 (2.55, 17.22)* & 5.71 (2.14, 20.34)* \\
 & Black & 33963 & 811 & 8.69 (7.46, 10.34) & 8.84 (7.69, 10.54) & 9.03 (7.90, 10.58) \\
 & Native Hawaiian & 1862 & 22 & 12.81 (5.70, 29.87)* & 10.42 (4.87, 24.27)* & 11.19 (5.51, 26.90)* \\
 & Unknown & 14447 & 139 & 4.55 (3.26, 7.41) & 3.90 (3.08, 5.60) & 3.97 (2.96, 5.93) \\
 & White & 159938 & 1537 & 6.54 (5.66, 7.70) & 6.47 (5.64, 7.48) & 5.71 (5.01, 6.61) \\
\end{xltabular}
}
  
\clearpage
\section*{Supplementary Table 6. Fairness summary: maximum PR-AUC gap and worst-group PR-AUC across demographic subgroups and prediction windows}\phantomsection
\label{supp_table_6}

The Likelihood Ratio Test evaluates whether the relationship between model predictions and outcomes varies across demographic subgroups. We apply the Benjamini-Hochberg procedure to control the FDR when testing heterogeneity across multiple subgroup variables (typically 5 per model-time window). The q-FDR value is the FDR-adjusted p-value; q-FDR < 0.05 indicates statistically significant heterogeneity after accounting for multiple comparisons.

{
\scriptsize
\renewcommand{\arraystretch}{1.2}
\setlength{\tabcolsep}{3pt}

\begin{xltabular}{\textwidth}{l l X c c c}
\toprule
\textbf{Subgroup} & \textbf{Model Class} & \textbf{Model} & 
\makecell[c]{\textbf{PR-AUC Gap}\\\textbf{(max--min), \%}} & 
\makecell[c]{\textbf{Worst-Group}\\\textbf{PR-AUC, \%}} & 
\textbf{LRT q-FDR} \\
\midrule
\endfirsthead

\midrule
\multicolumn{6}{c}{{Supplementary Table 6 -- continued}} \\
\midrule
\endhead

\midrule
\multicolumn{6}{r}{\textit{Continued on next page}} \\
\endfoot

\bottomrule
\endlastfoot

\multicolumn{6}{c}{\textbf{Prediction Window = 3 Months}} \\
\midrule
\multirow{7}{*}{Age} & \multirow{3}{*}{Machine Learning} & Elastic Net Logistic Regression & 4.30\% & 0.41\% & 0.398 \\
 & & Random Forest & 4.25\% & 0.61\% & 0.661 \\
 & & XGBoost & 4.26\% & 0.52\% & 0.848 \\ \cmidrule(lr){2-6}
 & \multirow{2}{*}{\makecell[l]{Masked Language\\Models}} & ModernBERT & 5.16\% & 0.36\% & 0.786 \\
 & & ClinicalModernBERT & 4.68\% & 0.56\% & 0.757 \\ \cmidrule(lr){2-6}
 & \multirow{2}{*}{\makecell[l]{Large Language\\Models}} & Llama-3.1-8B & 3.74\% & 0.41\% & 0.982 \\
 & & OpenBioLLM-8B & 4.46\% & 0.58\% & 0.647 \\
\midrule
\multirow{7}{*}{Ethnicity} & \multirow{3}{*}{Machine Learning} & Elastic Net Logistic Regression & 1.98\% & 1.81\% & 0.823 \\
 & & Random Forest & 1.08\% & 1.86\% & 0.810 \\
 & & XGBoost & 1.61\% & 2.18\% & 0.939 \\ \cmidrule(lr){2-6}
 & \multirow{2}{*}{\makecell[l]{Masked Language\\Models}} & ModernBERT & 5.10\% & 2.32\% & 0.786 \\
 & & ClinicalModernBERT & 1.42\% & 2.30\% & 0.921 \\ \cmidrule(lr){2-6}
 & \multirow{2}{*}{\makecell[l]{Large Language\\Models}} & Llama-3.1-8B & 2.00\% & 2.08\% & 0.982 \\
 & & OpenBioLLM-8B & 2.10\% & 1.91\% & 0.818 \\
\midrule
\multirow{7}{*}{Race} & \multirow{3}{*}{Machine Learning} & Elastic Net Logistic Regression & 15.34\% & 1.60\% & 0.010 \\
 & & Random Forest & 12.77\% & 1.95\% & 0.018 \\
 & & XGBoost & 9.39\% & 1.92\% & 0.228 \\ \cmidrule(lr){2-6}
 & \multirow{2}{*}{\makecell[l]{Masked Language\\Models}} & ModernBERT & 23.99\% & 1.58\% & 0.033 \\
 & & ClinicalModernBERT & 13.59\% & 2.16\% & 0.001 \\ \cmidrule(lr){2-6}
 & \multirow{2}{*}{\makecell[l]{Large Language\\Models}} & Llama-3.1-8B & 22.63\% & 1.86\% & 0.038 \\
 & & OpenBioLLM-8B & 17.92\% & 1.66\% & 0.211 \\
\midrule 

\multicolumn{6}{c}{\textbf{Prediction Window = 6 Months}} \\
\midrule
\multirow{7}{*}{Age} & \multirow{3}{*}{Machine Learning} & Elastic Net Logistic Regression & 6.84\% & 0.95\% & $<0.001$ \\
 & & Random Forest & 6.95\% & 1.32\% & 0.013 \\
 & & XGBoost & 6.03\% & 2.12\% & 0.016 \\ \cmidrule(lr){2-6}
 & \multirow{2}{*}{\makecell[l]{Masked Language\\Models}} & ModernBERT & 6.15\% & 0.85\% & 0.002 \\
 & & ClinicalModernBERT & 5.57\% & 0.93\% & $<0.001$ \\ \cmidrule(lr){2-6}
 & \multirow{2}{*}{\makecell[l]{Large Language\\Models}} & Llama-3.1-8B & 6.52\% & 1.04\% & $<0.001$ \\
 & & OpenBioLLM-8B & 5.33\% & 1.22\% & 0.032 \\
\midrule
\multirow{7}{*}{Ethnicity} & \multirow{3}{*}{Machine Learning} & Elastic Net Logistic Regression & 2.21\% & 2.08\% & 0.810 \\
 & & Random Forest & 2.69\% & 2.20\% & 0.869 \\
 & & XGBoost & 2.04\% & 3.16\% & 0.935 \\ \cmidrule(lr){2-6}
 & \multirow{2}{*}{\makecell[l]{Masked Language\\Models}} & ModernBERT & 2.00\% & 3.34\% & 0.503 \\
 & & ClinicalModernBERT & 1.58\% & 3.55\% & 0.309 \\ \cmidrule(lr){2-6}
 & \multirow{2}{*}{\makecell[l]{Large Language\\Models}} & Llama-3.1-8B & 1.25\% & 4.10\% & 0.427 \\
 & & OpenBioLLM-8B & 3.41\% & 2.40\% & 0.538 \\
\midrule
\multirow{7}{*}{Race} & \multirow{3}{*}{Machine Learning} & Elastic Net Logistic Regression & 2.37\% & 2.93\% & $<0.001$ \\
 & & Random Forest & 10.27\% & 3.09\% & 0.008 \\
 & & XGBoost & 10.15\% & 3.31\% & 0.043 \\ \cmidrule(lr){2-6}
 & \multirow{2}{*}{\makecell[l]{Masked Language\\Models}} & ModernBERT & 7.81\% & 3.46\% & $<0.001$ \\
 & & ClinicalModernBERT & 8.03\% & 3.56\% & $<0.001$ \\ \cmidrule(lr){2-6}
 & \multirow{2}{*}{\makecell[l]{Large Language\\Models}} & Llama-3.1-8B & 4.53\% & 4.12\% & 0.427 \\
 & & OpenBioLLM-8B & 9.74\% & 3.03\% & 0.062 \\
\midrule 

\multicolumn{6}{c}{\textbf{Prediction Window = 9 Months}} \\
\midrule
\multirow{7}{*}{Age} & \multirow{3}{*}{Machine Learning} & Elastic Net Logistic Regression & 6.62\% & 1.48\% & $<0.001$ \\
 & & Random Forest & 8.10\% & 2.20\% & 0.004 \\
 & & XGBoost & 7.14\% & 1.97\% & 0.012 \\ \cmidrule(lr){2-6}
 & \multirow{2}{*}{\makecell[l]{Masked Language\\Models}} & ModernBERT & 7.14\% & 1.78\% & $<0.001$ \\
 & & ClinicalModernBERT & 7.22\% & 1.24\% & $<0.001$ \\ \cmidrule(lr){2-6}
 & \multirow{2}{*}{\makecell[l]{Large Language\\Models}} & Llama-3.1-8B & 6.77\% & 2.19\% & 0.009 \\
 & & OpenBioLLM-8B & 6.01\% & 1.91\% & 0.011 \\
\midrule
\multirow{7}{*}{Ethnicity} & \multirow{3}{*}{Machine Learning} & Elastic Net Logistic Regression & 0.06\% & 4.50\% & 0.113 \\
 & & Random Forest & 1.38\% & 3.97\% & 0.147 \\
 & & XGBoost & 2.70\% & 4.75\% & 0.110 \\ \cmidrule(lr){2-6}
 & \multirow{2}{*}{\makecell[l]{Masked Language\\Models}} & ModernBERT & 1.27\% & 5.01\% & 0.041 \\
 & & ClinicalModernBERT & 1.51\% & 4.56\% & 0.110 \\ \cmidrule(lr){2-6}
 & \multirow{2}{*}{\makecell[l]{Large Language\\Models}} & Llama-3.1-8B & 2.26\% & 3.08\% & 0.366 \\
 & & OpenBioLLM-8B & 0.95\% & 3.61\% & 0.003 \\
\midrule
\multirow{7}{*}{Race} & \multirow{3}{*}{Machine Learning} & Elastic Net Logistic Regression & 7.51\% & 2.56\% & $<0.001$ \\
 & & Random Forest & 11.12\% & 2.50\% & 0.067 \\
 & & XGBoost & 9.27\% & 2.98\% & 0.001 \\ \cmidrule(lr){2-6}
 & \multirow{2}{*}{\makecell[l]{Masked Language\\Models}} & ModernBERT & 5.80\% & 2.54\% & $<0.001$ \\
 & & ClinicalModernBERT & 8.11\% & 3.22\% & $<0.001$ \\ \cmidrule(lr){2-6}
 & \multirow{2}{*}{\makecell[l]{Large Language\\Models}} & Llama-3.1-8B & 7.23\% & 3.21\% & 0.011 \\
 & & OpenBioLLM-8B & 6.50\% & 1.99\% & 0.001 \\
\midrule 

\multicolumn{6}{c}{\textbf{Prediction Window = 12 Months}} \\
\midrule
\multirow{7}{*}{Age} & \multirow{3}{*}{Machine Learning} & Elastic Net Logistic Regression & 7.51\% & 1.21\% & $<0.001$ \\
 & & Random Forest & 7.18\% & 1.80\% & $<0.001$ \\
 & & XGBoost & 7.28\% & 2.67\% & $<0.001$ \\ \cmidrule(lr){2-6}
 & \multirow{2}{*}{\makecell[l]{Masked Language\\Models}} & ModernBERT & 6.74\% & 2.24\% & $<0.001$ \\
 & & ClinicalModernBERT & 6.30\% & 2.67\% & $<0.001$ \\ \cmidrule(lr){2-6}
 & \multirow{2}{*}{\makecell[l]{Large Language\\Models}} & Llama-3.1-8B & 5.70\% & 4.16\% & $<0.001$ \\
 & & OpenBioLLM-8B & 7.22\% & 3.97\% & 0.187 \\
\midrule
\multirow{7}{*}{Ethnicity} & \multirow{3}{*}{Machine Learning} & Elastic Net Logistic Regression & 2.09\% & 3.77\% & 0.055 \\
 & & Random Forest & 2.62\% & 3.94\% & 0.304 \\
 & & XGBoost & 2.88\% & 4.02\% & 0.116 \\ \cmidrule(lr){2-6}
 & \multirow{2}{*}{\makecell[l]{Masked Language\\Models}} & ModernBERT & 2.07\% & 4.39\% & 0.083 \\
 & & ClinicalModernBERT & 3.18\% & 3.73\% & 0.019 \\ \cmidrule(lr){2-6}
 & \multirow{2}{*}{\makecell[l]{Large Language\\Models}} & Llama-3.1-8B & 1.31\% & 4.37\% & 0.189 \\
 & & OpenBioLLM-8B & 2.39\% & 3.82\% & 0.183 \\
\midrule
\multirow{7}{*}{Race} & \multirow{3}{*}{Machine Learning} & Elastic Net Logistic Regression & 9.51\% & 3.94\% & $<0.001$ \\
 & & Random Forest & 5.92\% & 5.06\% & $<0.001$ \\
 & & XGBoost & 8.26\% & 4.55\% & $<0.001$ \\ \cmidrule(lr){2-6}
 & \multirow{2}{*}{\makecell[l]{Masked Language\\Models}} & ModernBERT & 9.24\% & 4.18\% & $<0.001$ \\
 & & ClinicalModernBERT & 6.52\% & 3.90\% & $<0.001$ \\ \cmidrule(lr){2-6}
 & \multirow{2}{*}{\makecell[l]{Large Language\\Models}} & Llama-3.1-8B & 5.58\% & 2.47\% & $<0.001$ \\
 & & OpenBioLLM-8B & 7.55\% & 2.18\% & 0.183 \\
\end{xltabular}
}

\clearpage
\section*{Supplementary Note 1. ICD codes and categorization of predictors}\phantomsection
\label{supp_note_1}
Diagnosis-based predictors were defined using ICD-10-CM codes extracted from the VA Corporate Data Warehouse and grouped into clinically coherent domains. Mental health conditions (e.g., depressive disorders, PTSD, psychotic disorders), substance use disorders (alcohol, opioids, stimulants, cannabis, other drugs), and physical health conditions (e.g., cardiovascular, metabolic, respiratory, neurologic) were categorized using code lists developed with clinical input and aligned to prior VA and homelessness literature. Social and behavioral factors of health (SBFH) predictors were derived from ICD-10-CM Z-codes and related flags capturing housing instability, financial and employment problems, interpersonal violence, and other social needs.

\subsection*{Physical Health Disorders}

\begin{xltabular}{\textwidth}{l X}
\toprule
\textbf{Predictor} & \textbf{ICD-10 Codes} \\
\midrule
\endfirsthead
\midrule
\multicolumn{2}{c}{{\tablename\ -- continued}} \\
\midrule
\endhead
\midrule
\multicolumn{2}{r}{{Continued on next page}} \\
\endfoot
\bottomrule
\endlastfoot
AIDS/HIV & B20, B21, B22, B24 \\ \addlinespace
Blood Loss Anemia & D50.0 \\ \addlinespace
Cardiac Arrhythmia & I44.1, I44.2, I44.3, I45.6, I45.9, I47, I48, I49, R00.0, R00.1, R00.8, T82.1, Z45.0, Z95.0 \\ \addlinespace
Cardiovascular Disease & I48, I49, I70 \\ \addlinespace
Chronic Pulmonary Disease & I27.8, I27.9, J40, J41, J42, J43, J44, J45, J46, J47, J60, J61, J62, J63, J64, J65, J66, J67, J68.4, J70.1, J70.3 \\ \addlinespace
Cirrhosis & K70.3, K71.7, K74, K76.1 \\ \addlinespace
Coagulopathy & D65, D66, D67, D68, D69.1, D69.3, D69.4, D69.5, D69.6 \\ \addlinespace
Congestive Heart Failure & I09.9, I11.0, I13.0, I13.2, I25.5, I42.0, I42.5, I42.6, I42.7, I42.8, I42.9, I43, I50, P29.0 \\ \addlinespace
Deficiency Anemia & D50.8, D50.9, D51, D52, D53 \\ \addlinespace
Diabetes & E10.2, E10.3, E10.4, E10.5, E10.6, E10.7, E10.8, E11.2, E11.3, E11.4, E11.5, E11.6, E11.7, E11.8, E12.2, E12.3, E12.4, E12.5, E12.6, E12.7, E12.8, E13.2, E13.3, E13.4, E13.5, E13.6, E13.7, E13.8, E14.2, E14.3, E14.4, E14.5, E14.6, E14.7, E14.8, E10.0, E10.1, E10.9, E11.0, E11.1, E11.9, E12.0, E12.1, E12.9, E13.0, E13.1, E13.9, E14.0, E14.1, E14.9 \\ \addlinespace
Fluid and Electrolyte Disorders & E22.2, E86, E87 \\ \addlinespace
Hepatitis & B17.1, B18.2, B19.2, Z22.52, B15.0, B15.9, B16, B16.0, B16.1, B16.2, B16.9, B17.0, B17.1, B17.10, B17.11, B17.2, B17.8, B17.9, B18.0, B18.1, B18.2, B18.8, B18.9, B19, B19.0, B19.1, B19.10, B19.11, B19.2, B19.20, B19.21, B19.9, B25.1 \\ \addlinespace
Hypertension & I11, I12, I13, I15, I10 \\ \addlinespace
Hypothyroidism & E00, E01, E02, E03, E89.0 \\ \addlinespace
Influenza & J09.X1, J09.X2, J09.X3, J09.X9, J10.00, J10.01, J10.08, J10.1, J10.2, J10.81, J10.82, J10.83, J10.89, J11.00, J11.08, J11.1, J11.2, J11.81, J11.82, J11.83, J11.89 \\ \addlinespace
Liver Disease & B18, I85, I86.4, I98.2, K70, K71.1, K71.3, K71.4, K71.5, K71.7, K72, K73, K74, K76.0, K76.2, K76.3, K76.4, K76.5, K76.6, K76.7, K76.8, K76.9, Z94.4 \\ \addlinespace
Lymphoma & C81, C82, C83, C84, C85, C88, C96, C90.0, C90.2 \\ \addlinespace
Metastatic Cancer & C77, C78, C79, C80 \\ \addlinespace
Obesity & E66 \\ \addlinespace
Pain & A02.\%, A18.\%, A39.\%, A52.\%, A54.\%, A59.\%, A69.\%, B0\%, B02.\%, B06.\%, B1\%, C0\%, C1\%, D0\%, D1\%, D57.\%, D86.\%, E08.\%, E10.\%, E11.\%, E13.\%, F45.\%, G25.\%, G43\%, G43.\%, G44.\%, G50.\%, G52.\%, G54.\%, G56.\%, G57.\%, G58.\%, G59\%, G60.\%, G61.\%, G62.\%, G63\%, G72.\%, G89.\%, G90.\%, G96.\%, G99.\%, H46.\%, H47.\%, H57.\%, I20.\%, I77.\%, K40.\%, K41.\%, K42.\%, K43.\%, K44.\%, K45.\%, K46.\%, K58.\%, K80.\%, L40.\%, L97.\%, L98.\%, M00.\%, M01\%, M02.\%, M05.\%, M06.\%, M07.\%, M08.\%, M10.\%, M11.\%, M12.\%, M13.\%, M14.\%, M15.\%, M16.\%, M17.\%, M18.\%, M19.\%, M1A\%, M20.\%, M21.\%, M22.\%, M23.\%, M24.\%, M25.\%, M26.\%, M27.\%, M32.\%, M33.\%, M34.\%, M35.\%, M43.\%, M45.\%, M46.\%, M47.\%, M48.\%, M49.\%, M50.\%, M51.\%, M53.\%, M54.\%, M60.\%, M62.\%, M65.\%, M66.\%, M67.\%, M70.\%, M71.\%, M72.\%, M75.\%, M76.\%, M77.\%, M79.\%, M80.\%, M84.\%, M86.\%, M87.\%, M89.\%, M90.\%, M94.\%, M95.\%, M96.\%, M99.\%, N20.\%, N21.\%, N22\%, N30.\%, N41.\%, N42.\%, N45.\%, N50.\%, N51\%, N53.\%, N71.\%, N72\%, N73.\%, N80.\%, N94.\%, O26.\%, Q66.\%, Q67.\%, Q74.\%, Q76.\%, R07.\%, R10.\%, R25.\%, R51\%, R52\%, R68.\%, S00.\%, S02.\%, S03.\%, S05.\%, S06.\%, S09.\%, S10.\%, S12.\%, S13.\%, S14.\%, S16.\%, S19.\%, S20.\%, S22.\%, S23.\%, S24.\%, S29.\%, S30.\%, S32.\%, S33.\%, S34.\%, S39.\%, S30.\%, S32.\%, S33.\%, S34.\%, S39.\%, S40.\%, S42.\%, S43.\%, S46.\%, S49.\%, S50.\%, S52.\%, S53.\%, S56.\%, S59.\%, S60.\%, S62.\%, S63.\%, S66.\%, S70.\%, S72.\%, S73.\%, S76.\%, S79.\%, S80.\%, S82.\%, S83.\%, S86.\%, S89.\%, S90.\%, S92.\%, S93.\%, S96.\%, T07\%, T14.\%, T83.\%, T84.\%, T85.\%, X0\%, X11\%, X12\%, X19\%, X21\%, X22\%, X29\%, X31\%, X32\%, X39\%, X41\%, X42\%, X49\%, X51\%, X52\%, X59\%, X61\%, X62\%, X69\%, X71\%, X72\%, X79\%, X8\%, X9\% \\ \addlinespace
Paralysis & G04.1, G11.4, G80.1, G80.2, G81, G82, G83.0, G83.1, G83.2, G83.3, G83.4, G83.9 \\ \addlinespace
Peptic Ulcer Disease & K25.7, K25.9, K26.7, K26.9, K27.7, K27.9, K28.7, K28.9 \\ \addlinespace
Peripheral Vascular Disorders & I70, I71, I73.1, I73.8, I73.9, I77.1, I79.0, I79.2, K55.1, K55.8, K55.9, Z95.8, Z95.9 \\ \addlinespace
Pulmonary Circulation Disorders & I26, I27, I28.0, I28.8, I28.9 \\ \addlinespace
Renal Failure & I12.0, I13.1, N18, N19, N25.0, Z49.0, Z49.1, Z49.2, Z94.0, Z99.2 \\ \addlinespace
Rheumatoid Arthritis/collagen & L94.0, L94.1, L94.3, M05, M06, M08, M12.0, M12.3, M30, M31.0, M31.1, M31.2, M31.3, M32, M33, M34, M35, M45, M46.1, M46.8, M46.9 \\ \addlinespace
Solid Tumor without Metastasis & C00, C01, C02, C03, C04, C05, C06, C07, C08, C09, C10, C11, C12, C13, C14, C15, C16, C17, C18, C19, C20, C21, C22, C23, C24, C25, C26, C30, C31, C32, C33, C34, C37, C38, C39, C40, C41, C43, C45, C46, C47, C48, C49, C50, C51, C52, C53, C54, C55, C56, C57, C58, C60, C61, C62, C63, C64, C65, C66, C67, C68, C69, C70, C71, C72, C73, C74, C75, C76, C97 \\ \addlinespace
Traumatic brain injury & F07.81, S02.0, S02.1, S02.8, S02.9, S04.2, S04.3, S04.4, S06., S07.1, Z87.820 \\ \addlinespace
Valvular Disease & A52.0, I05, I06, I07, I08, I09.1, I09.8, I34, I35, I36, I37, I38, I39, Q23.0, Q23.1, Q23.2, Q23.3, Z95.2, Z95.3, Z95.4 \\ \addlinespace
Weight Loss & E40, E41, E42, E43, E44, E45, E46, R63.4, R64 \\
\end{xltabular}

\clearpage
\subsection*{Mental Health Disorders}

\begin{xltabular}{\textwidth}{l X}
\toprule
\textbf{Predictor} & \textbf{ICD-10 Codes} \\
\midrule
\endfirsthead
\midrule
\multicolumn{2}{c}{{\tablename\ \thetable{} -- continued}} \\
\midrule
\textbf{Predictor} & \textbf{ICD-10 Codes} \\
\midrule
\endhead
\midrule
\multicolumn{2}{r}{{Continued on next page}} \\
\endfoot
\bottomrule
\endlastfoot
Anxiety Disorder & F40, F41, F01.54, F01.A4, F01.B4, F01.C4, F02.84, F02.A4, F02.B4, F02.C4, F03.94, F03.A4, F03.B4, F03.C4, F06.4 \\ \addlinespace
Bipolar disorder & F30, F31 \\ \addlinespace
Dementia & F01, F02, F03, G30, G31.0, G31.83 \\ \addlinespace
Depression & F20.4, F31.3, F31.4, F31.5, F32, F33, F34.1, F41.2, F43.2 \\ \addlinespace
Other Neurological Disorders & G10, G11, G12, G13, G20, G21, G22, G25.4, G25.5, G31.2, G31.8, G31.9, G32, G35, G36, G37, G40, G41, G93.1, G93.4, R47.0, R56 \\ \addlinespace
Posttraumatic Stress Disorder & F43.10, F43.11, F43.12 \\ \addlinespace
Psychoses & F20, F22, F23, F24, F25, F28, F29, F30.2, F31.2, F31.5 \\ \addlinespace
Sleep Disorder & F51.9, F51.8, F51.19, F51.09, F51.1, G47.27, G47.24, G47.22, G47.26, G47.12, G47.54, G47.25, G47.11, F51.13, G47.69, F51.05, G47.29, G47.14, G47.8, G47.53, G47.01, G47.20, G47.6, G47.9, G47.61, F51.4, G47.21, G47.23, F51.12, F51.3, G47.52, F51.04, G47.2, G47.10, G47.62, G47.50, G25.81, G47.00, G47.13, G47.63, F51.03, G47.51, F51.02, F51.11, F51.5, G47.19, F51.01, G47.59, G47, G47.09, G47.1, G47.5, G47.0, G25.81 \\
\end{xltabular}

\clearpage
\subsection*{Substance Abuse Disorders}

\begin{xltabular}{\textwidth}{l X}
\toprule
\textbf{Predictor} & \textbf{ICD-10 Codes} \\
\midrule
\endfirsthead
\midrule
\multicolumn{2}{c}{{\tablename\ \thetable{} -- continued}} \\
\midrule
\textbf{Predictor} & \textbf{ICD-10 Codes} \\
\midrule
\endhead
\midrule
\multicolumn{2}{r}{{Continued on next page}} \\
\endfoot
\bottomrule
\endlastfoot
Alcohol Use Disorder & F10, G62.1, I42.6, K29.2, K70, T51.0 \\ \addlinespace
Cannabis & F12 \\ \addlinespace
Cocaine & F14 \\ \addlinespace
Drug Abuse & F12, F13, F14, F15, F16, F18, F19, Z71.5, Z72.2 \\ \addlinespace
Hallucinogen & F16 \\ \addlinespace
Nicotine dependence & F17 \\ \addlinespace
Opioid Use Disorder & F11.10, F11.11, F11.120, F11.121, F11.122, F11.129, F11.13,
F11.14, F11.150, F11.151, F11.159,
F11.181, F11.182, F11.188, F11.19, F11.20, F11.21, F11.220, F11.221, F11.222, F11.229, F11.23,
F11.24, F11.250,F11.251, F11.259, F11.281, F11.282, F11.288, F11.29 \\ \addlinespace
Other stimulant & F15 \\
\end{xltabular}

\clearpage
\subsection*{Social and Behavioral Factors of Health}

\begin{xltabular}{\textwidth}{l X X}
\toprule
\textbf{Predictor} & \textbf{ICD-10 Codes} & \textbf{Stop Codes} \\
\midrule
\endfirsthead
\midrule
\multicolumn{3}{c}{{\tablename\ \thetable{} -- continued}} \\
\midrule
\textbf{Predictor} & \textbf{ICD-10 Codes} & \textbf{Stop Codes} \\
\midrule
\endhead
\midrule
\multicolumn{3}{r}{{Continued on next page}} \\
\endfoot
\bottomrule
\endlastfoot
Employment or financial problem & Z56, Z59.5-9 & 208, 222, 535, 555, 568, 574 \\ \addlinespace
Food insecurity & Z59.4 & --- \\ \addlinespace
Housing instability & Z59.1, Z59.81 (CDW) & 504, 507, 508, 511, 522, 528, 529, 530, 555, 556, 590 \\ \addlinespace
Legal problems & Z65.0-4, Y92.14 & 591, 592 \\ \addlinespace
Non-specific psychosocial needs & R41.83, Z53.1, Z55.0-4, Z55.8-9, Z60.0, Z60.3-5, Z60.8-9, Z56.1, Z64.4, Z65.0-4, Z65.8-9, Z73.4-6 & --- \\ \addlinespace
Social or familial problems & Z59.2, Z59.3, Z55, Z60 & --- \\ \addlinespace
Violence Problems & O9A.3\%, O9A.4\%, O9A.5\%, T74\%, T76\%, X92\%, X93\%, X94\%, X95\%, X96\%, X97\%, X98\%, X99\%, Y00\%, Y01\%, Y02\%, Y03\%, Y04\%, Y07\%, Y08\%, Y09\%, Y35\%, Y36\%, Y37\%, Y38\%, Z04.4\%, Z04.7\%, Z04.81\%, Z65.0\%, Z65.1\%, Z65.2\%, Z65.3\%, Z65.4\%, Z65.5\%, Z65.8\%, Z65.9\%, Z69\%, Z91.4\% & 524 \\
\end{xltabular}

\clearpage
\subsection*{Service Utilization Predictors}

\begin{xltabular}{\textwidth}{l X}
\toprule
\textbf{Predictor} & \textbf{Stop Codes} \\
\midrule
\endfirsthead
\midrule
\multicolumn{2}{c}{{\tablename\ \thetable{} -- continued}} \\
\midrule
\textbf{Predictor} & \textbf{Stop Codes} \\
\midrule
\endhead
\midrule
\multicolumn{2}{r}{{Continued on next page}} \\
\footnotemark
\endfoot
\bottomrule
\endlastfoot
Outpatient Visits: Primary care & 170, 171, 172, 301, 318, 322, 323, 338, 348, 350, 704 \\ \addlinespace
Outpatient Visits: Mental health & 156, 157, 292, 464, 502, 509, 510, 516, 524, 525, 527, 531, 533, 534, 535, 536, 538, 539, 542, 546, 547, 550, 552, 561, 562, 564, 565, 566, 567, 568, 571, 572, 573, 574, 575, 576, 577, 579, 582, 583, 584, 586, 587, 593, 596, 597, 598, 599, 713 \\ \addlinespace
Outpatient Visits: Substance abuse & 513, 514, 519, 523, 545, 547, 548, 560, 586, 587, 593, 596, 597, 598, 599, 706, 707, 721, 722, 723, 724 \\ \addlinespace
Outpatient Visits: Specialty care & 120, 290, 324, 331, 336, 143, 231, 293, 302, 303, 304, 305, 306, 307, 308, 309, 310, 311, 312, 313, 314, 315, 316, 317, 321, 325, 327, 329, 330, 333, 334, 335, 337, 339, 340, 344, 345, 346, 349, 356, 369, 391, 392, 394, 420, 602, 603, 604, 606, 607, 608, 611, 118, 119, 121, 173, 174, 175, 176, 177, 178, 190, 191, 319, 326, 347, 351, 352, 353, 354, 658, 680, 682, 291, 401, 402, 403, 404, 405, 406, 407, 408, 409, 410, 411, 413, 414, 415, 418, 419, 424, 427, 428, 429, 430, 432, 434, 435, 441, 486, 487, 488, 489, 718 \\ \addlinespace
Outpatient Visits: Diagnostic / Ancillary & 192, 651, 674, 669, 103, 123, 124, 125, 139, 142, 147, 159, 160, 166, 167, 169, 180, 181, 182, 328, 332, 372, 373, 436, 683, 685, 686, 104, 105, 106, 107, 108, 109, 110, 111, 115, 116, 126, 128, 144, 145, 146, 148, 149, 150, 151, 155, 158, 212, 421, 703 \\ \addlinespace
Outpatient Visits: Rehabilitation & 195, 196, 197, 198, 199, 201, 202, 203, 204, 205, 206, 207, 208, 209, 210, 211, 213, 214, 215, 216, 217, 218, 220, 221, 222, 224, 225, 229, 230, 240, 241, 250, 417, 423, 425, 437, 438, 439 \\ \addlinespace
Emergency / Urgent-care & 130, 131 \\ \addlinespace
Inpatient Visits: Total & 25, 26, 33, 38, 39, 45, 68, 70, 71, 75, 76, 77, 79, 88, 89, 91, 92, 93, 94, 109, 110, 27, 29, 72, 73, 74, 84, 86, 90, 111, 1-19, 24, 30, 31, 34, 83, 48-63, 65, 78, 97, 106, 107, 108, 32, 34, 40, 42, 43, 44, 46, 47, 64, 66, 67, 68, 69, 80, 81, 95, 96, 100, 101, 102, 103, 104, 105, 20, 35, 41, 82, 21, 36, 112, 22, 23, 85 \\
\end{xltabular}

\clearpage
\clearpage

\section*{Supplementary Table 7. Data-split characteristics across prediction windows}\phantomsection
\label{supp_table_7}

*Training set values shown before stratified downsampling. The training set was downsampled to balance classes (matched on gender, age group, and race). validation and test sets retained original prevalence distribution. Data was split at the patient level to prevent data leakage.

{
\renewcommand{\arraystretch}{1.1}

\begin{xltabular}{\textwidth}{l X c c c} \label{tab:data_splits} \\
\toprule
\textbf{Prediction Window} & \textbf{Data Split} & \textbf{N} & \textbf{N (\%)} & \textbf{Events (n)} \\
\midrule
\endfirsthead

\toprule
\textbf{Prediction Window} & \textbf{Data Split} & \textbf{N} & \textbf{N (\%)} & \textbf{Events (n)} \\
\midrule
\endhead

\midrule
\multicolumn{5}{r}{\textit{Continued on next page}} \\
\endfoot

\bottomrule
\endlastfoot

\multirow{3}{*}{3 Months} 
 & Training* & 3,934,290 & 92\% & 12,630 \\
 & Validation & 128,292 & 3\% & 412 \\
 & Test & 213,820 & 5\% & 686 \\
\midrule

\multirow{3}{*}{6 Months} 
 & Training* & 3,934,290 & 92\% & 24,672 \\
 & Validation & 128,292 & 3\% & 805 \\
 & Test & 213,820 & 5\% & 1,341 \\
\midrule

\multirow{3}{*}{9 Months} 
 & Training* & 3,934,290 & 92\% & 36,266 \\
 & Validation & 128,292 & 3\% & 1,183 \\
 & Test & 213,820 & 5\% & 1,971 \\
\midrule

\multirow{3}{*}{12 Months} 
 & Training* & 3,934,290 & 92\% & 46,921 \\
 & Validation & 128,292 & 3\% & 1,531 \\
 & Test & 213,820 & 5\% & 2,550 \\
\end{xltabular}
}

\clearpage
\section*{Supplementary Note 2. Model hyperparameters and training configuration}\phantomsection
\label{supp_note_2}

\textbf{Computational Resources}\\
All language models (ModernBERT, Bioclinical ModernBERT, Llama-3.1-8B, OpenBio) were trained on NVIDIA A100 80GB GPUs. Training did not utilize the full server capacity, with resources allocated as needed for each model's requirements.\\
\textbf{Logistic regression (elastic net)}\\ We fit an elastic net logistic regression model that combines L1 and L2 penalties. We used randomized search with 100 iterations. The search varied the inverse regularization strength C over the values [0.03, 0.1, 0.3, 1, 3, 10, and 30], and varied the elastic net mixing parameter (l1\_ratio) over [0.0, 0.1, 0.3, 0.5, 0.7, 0.9, and 1.0]. The maximum number of iterations was allowed to be [2,000, 4,000, or 6,000]. We used the SAGA solver with an elastic net penalty.\\
\textbf{Random forest}\\ For the random forest model, we used randomized search with 100 iterations. The search space included the number of trees (n\_estimators) with candidate values of [400, 800, 1,200, and 1,600]. We also varied maximum tree depth [None, 10, 20, 40, 80], the minimum number of samples required to split a node (min\_samples\_split) [2, 5, 10, 20, 50], the minimum number of samples required at a leaf node (min\_samples\_leaf) [1, 2, 5, 10, 20], the number of features considered at each split (max\_features) [sqrt, log2, 0.2, 0.4, 0.6, 0.8], the complexity parameter (ccp\_alpha) [0.0, 1e-4, 1e-3], and the fraction of samples used per tree (max\_samples) [None, 0.5, 0.8]. Bootstrap sampling was enabled.\\
\textbf{XGBoost}\\ For gradient boosted trees (XGBoost), we used randomized search with 100 iterations. We varied the number of boosting rounds (n\_estimators) with candidate values of [300, 600, 1,000, and 1,500]. Other tuned hyperparameters included maximum tree depth [3, 6, 9, 12, 15], learning rate [0.01, 0.05, 0.1, 0.2], row subsampling ratio (subsample) [0.6, 0.8, 0.9, 1.0], feature subsampling ratio (colsample\_bytree) [0.6, 0.8, 0.9, 1.0], L1 regularization (reg\_alpha) [0, 0.1, 1, 10], L2 regularization (reg\_lambda) [0, 0.1, 1, 10], minimum child weight [1, 3, 5, 7], and minimum loss reduction (gamma) [0, 0.1, 0.5, 1]. The objective was binary logistic regression, and we used the "hist" tree method for efficiency.\\
\textbf{ModernBERT}\\ We fine-tuned the ModernBERT-large model as a sequence classification model using a fixed configuration without hyperparameter search. We trained for 10 epochs with a learning rate of 0.00005 (5e-5). We used a batch size of 16 with gradient accumulation of 2, resulting in an effective batch size of 32. Optimization used the AdamW optimizer with weight decay of 0.05, cosine learning rate scheduling, and a warmup ratio of 0.1. We set the maximum sequence length to 512 tokens. Training was performed in mixed fp16 precision without quantization. Early stopping was applied based on precision-recall area under the curve (PR-AUC) on the validation set, with a patience of 20 evaluation intervals. A standard classification head was added on top of the encoder\\
\textbf{BioClinical-ModernBERT}\\ We also fine-tuned a clinically pre-trained variant, BioClinical-ModernBERT-large. All training hyperparameters and settings were identical to those used for ModernBERT, including optimizer, learning rate, weight decay, number of epochs, batch size, gradient accumulation, scheduler, sequence length, precision, and early stopping strategy.\\
\textbf{Llama-3.1-8B}\\ We fine-tuned the Meta-Llama-3.1-8B base model using Low-Rank Adaptation (LoRA) with a fixed configuration and no hyperparameter search. The LoRA rank was 16, the LoRA alpha parameter was 32, and the LoRA dropout rate was 0.1. LoRA adapters were applied to the attention and feed-forward projection modules, including query, key, value, output, and intermediate projections (q\_proj, k\_proj, v\_proj, o\_proj, up\_proj, down\_proj, gate\_proj), as well as to the language modeling head (lm\_head). We trained for 10 epochs with a learning rate of 0.0001 (1e-4) and an effective batch size of 64 (batch size of 4 multiplied by gradient accumulation steps of 16). We used the AdamW optimizer in its fused implementation (adamw\_torch\_fused), with a linear learning rate schedule, a warmup ratio of 0.03, a maximum gradient norm of 1.0, and a maximum sequence length of 512 tokens. Training used mixed bfloat16 precision without additional quantization. Gradient checkpointing was enabled to reduce memory usage. Early stopping was based on validation PR-AUC with a patience of 10 evaluation intervals. The language modeling head was fully fine-tuned together with the LoRA adapters. Training uses answer-only loss calculation, where only the final predicted token contributes to the loss, focusing the model on the classification task.\\
\textbf{OpenBio LLM}\\We also fine-tuned a clinically oriented model, OpenBioLLM-8B, using the same LoRA and training configuration as for Llama-3.1-8B. Keeping these settings identical ensured comparability between the two large language models.

\clearpage
\section*{Supplementary Note 3. Illustrative example of EHR representation and model inputs}\phantomsection
\label{supp_note_3}

Illustrative example of how a single patient’s electronic health record (EHR) data are transformed into the representations used by our machine learning (ML), masked language model (MLM), and large language model (LLM) approaches. The schematic version of this pipeline appears in Figure 1 of the main text. The example is purely illustrative and does not correspond to a real patient.

\subsection*{Stage 1: Raw longitudinal EHR table}

Consider a hypothetical patient (Patient 001) with three visits during the observation year 2016. For simplicity, we show only age, gender, three diagnosis groups (Anxiety disorder, Cancer, Influenza), and an indicator for legal problems. The first two visits (February and April) fall in the first time interval (H1; e.g., January-June 2016), and the September visit falls in the second interval (H2; e.g., July-December 2016).
\\
For Patient 001 (Age 32, Male), the visit-level records are:

\newcolumntype{Y}{>{\centering\arraybackslash}X}

{
\renewcommand{\arraystretch}{1.3} 

\begin{xltabular}{\textwidth}{*{7}{Y}}
\label{tab:longitudinal_full_width} \\
\toprule
\textbf{Visit Date} & \textbf{Age} & \textbf{Gender} & \makecell{\textbf{Anxiety}\\\textbf{Disorder}} & \textbf{Cancer} & \textbf{Influenza} & \makecell{\textbf{Legal}\\\textbf{Problems}} \\
\midrule
\endfirsthead

\toprule
\textbf{Visit Date} & \textbf{Age} & \textbf{Gender} & \makecell{\textbf{Anxiety}\\\textbf{Disorder}} & \textbf{Cancer} & \textbf{Influenza} & \makecell{\textbf{Legal}\\\textbf{Problems}} \\
\midrule
\endhead

\midrule
\multicolumn{7}{r}{\textit{Continued on next page}} \\
\endfoot

\bottomrule
\endlastfoot

10/02/16 & 32 & M & 1 & 0 & 1 & 1 \\ \addlinespace
22/04/16 & 32 & M & 1 & 0 & 0 & 0 \\ \addlinespace
15/09/16 & 32 & M & 0 & 1 & 0 & 0 \\
\end{xltabular}
}

\subsection*{Stage 2: Longitudinal representation without condition persistence rules}

 We first aggregate visit-level information within each interval. A value of 1 indicates at least one occurrence of the diagnosis/event in that period. This preserves temporal ordering (H1 vs H2) but does not yet incorporate condition persistence rules.

{
\renewcommand{\arraystretch}{1.3}

\begin{xltabular}{\textwidth}{*{3}{Y}}
\label{tab:aggregated_h1_h2} \\
\toprule
\textbf{Variable} & \makecell{\textbf{H1}\\\textbf{(2016 Jan - Jun)}} & \makecell{\textbf{H2}\\\textbf{(2016 Jul - Dec)}} \\
\midrule
\endfirsthead

\toprule
\textbf{Variable} & \makecell{\textbf{H1}\\\textbf{(2016 Jan - Jun)}} & \makecell{\textbf{H2}\\\textbf{(2016 Jul - Dec)}} \\
\midrule
\endhead

\midrule
\multicolumn{3}{r}{\textit{Continued on next page}} \\
\endfoot

\bottomrule
\endlastfoot

Anxiety Disorder & 1 & 0 \\ \addlinespace
Cancer & 0 & 1 \\ \addlinespace
Influenza & 1 & 0 \\ \addlinespace
Legal Problems & 1 & 0 \\
\end{xltabular}
}

\subsection*{Stage 3: Applying the condition persistence framework}

As described in Supplementary Table 9, each predictor is assigned a \textbf{condition persistence rule} (e.g., ever-history, recurrent time-limited, episodic). In this toy example:

\begin{itemize}
    \item Anxiety disorder: recurrent time-limited
    \item Legal problems: recurrent time-limited
    \item Cancer: ever-history
    \item Influenza: episodic
\end{itemize}

Both Anxiety disorder and Legal problems use a recurrent time-limited rule with persistence for T = 2 time intervals. Cancer uses an ever-history rule, and Influenza has no persistence beyond the interval in which it is recorded.

Applying these rules to the interval-level representation yields (with * indicating values updated by persistence):

{
\renewcommand{\arraystretch}{1.3}

\begin{xltabular}{\textwidth}{*{3}{Y}}
\label{tab:aggregated_carry_forward} \\
\toprule
\textbf{Variable} & \makecell{\textbf{H1}\\\textbf{(2016 Jan - Jun)}} & \makecell{\textbf{H2}\\\textbf{(2016 Jul - Dec)}} \\
\midrule
\endfirsthead

\toprule
\textbf{Variable} & \makecell{\textbf{H1}\\\textbf{(2016 Jan - Jun)}} & \makecell{\textbf{H2}\\\textbf{(2016 Jul - Dec)}} \\
\midrule
\endhead

\midrule
\multicolumn{3}{r}{\textit{Continued on next page}} \\
\endfoot

\bottomrule
\endlastfoot

Anxiety Disorder & 1 & 1* \\ \addlinespace
Cancer & 1* & 1 \\ \addlinespace
Influenza & 1 & 0 \\ \addlinespace
Legal Problems & 1 & 1* \\
\end{xltabular}
}

In the actual pipeline, these condition persistence rules are applied at the end of the observation window to construct features for prediction and do not introduce information from future time periods into earlier prediction windows.

\subsection*{Stage 4: Construction of ML feature representation}

For Elastic Net Logistic Regression, random forest, and related tabular models, we collapse the longitudinal information into a single row per patient at the end of the observation window, using interval-specific indicators (e.g., anxiety\_H1, anxiety\_H2). The full models use a much richer feature set spanning demographics, mental and physical health, substance use, social factors of health, and utilization. A simplified subset of the resulting feature vector for Patient 001 is:

{
\renewcommand{\arraystretch}{1.2}

\begin{xltabular}{\textwidth}{*{2}{Y}}
\label{tab:feature_vector} \\
\toprule
\textbf{Feature} & \textbf{Value} \\
\midrule
\endfirsthead

\toprule
\textbf{Feature} & \textbf{Value} \\
\midrule
\endhead

\midrule
\multicolumn{2}{r}{\textit{Continued on next page}} \\
\footnotemark
\endfoot

\bottomrule
\endlastfoot

age\_30\_39 & 1 \\ \addlinespace
sex\_male & 1 \\ \addlinespace
anxiety\_H1 & 1 \\ \addlinespace
anxiety\_H2 & 1 \\ \addlinespace
cancer\_H1 & 1 \\ \addlinespace
cancer\_H2 & 1 \\ \addlinespace
influenza\_H1 & 1 \\ \addlinespace
influenza\_H2 & 0 \\ \addlinespace
legal\_problems\_H1 & 1 \\ \addlinespace
legal\_problems\_H2 & 1 \\
\end{xltabular}
}

\subsection*{Stage 5: Text Prompt for MLM and LLMs}

The same patient profile is converted into a compact natural language prompt. For this toy example (3-month prediction window), the prompt is:

\begin{tcolorbox}[
    colback=gray!5, 
    colframe=black, 
    width=\textwidth, 
    arc=2mm, 
    boxrule=0.5pt,
    title=\textbf{Patient Prediction Task: 3-Month Homelessness Risk},
    fonttitle=\sffamily\bfseries
]
\texttt{Patient Information:} \\
\texttt{demographics: \{Gender: male, Age: 32\};} \\
\texttt{mental\_health\_disorders: \{H1: Anxiety disorder; H2: Anxiety disorder\};} \\
\texttt{physical\_health: \{H1: Influenza; H2: Cancer\};} \\
\texttt{social\_and\_behavioral\_factors: \{H1-H2: Legal problems\}.} 

\vspace{1em}
\textbf{Task:} Given the patient information, predict \textit{yes} if this patient will be homeless in the next 3 months, \textit{no} otherwise.
\end{tcolorbox}

The prompt structure follows this template, where features are organized by clinical domain and time periods:

\begin{tcolorbox}[colback=white, colframe=black!70, sharp corners, boxrule=0.5pt, width=\textwidth]
\small
\texttt{Patient Information: [domain1]: \{[static features] or \{[time\_period]: [features]\}\}; [domain2]: \{[time\_period]: [features]\}; ... Given the patient information, predict yes if this patient will be homeless in the next [M] months, no otherwise.}
\end{tcolorbox}

\noindent\textbf{Where:}

\begin{itemize}[leftmargin=*]
    \item \textbf{[domain]} represents one of: demographics, utilization, mental\_health\_disorders, physical\_health, substance\_abuse, military\_history, or social\_and\_behavioral\_factors
    \item \textbf{[time\_period]} is one of: Q1, Q2, Q3, Q4 (for quarterly aggregation) or H1, H2 (for half-year aggregation)
    \item \textbf{[M]} is the prediction window: 3 months, 6 months, 9 months, or 12 months
    \item \textbf{Static features} (e.g., demographics) appear as \{Feature: value, Feature: value\}
    \item \textbf{Time-varying features} appear as \{Q1: feature1, feature2; Q2: feature3\} with features listed within each time period
    \item \textbf{Only present features} are included (absent features are omitted)
    \item \textbf{Count/utilization features} include their values (e.g., Primary care visits: 3), while binary features appear as names only when present
\end{itemize}

\clearpage
\section*{Supplementary Table 8. Selected aggregation level for time-varying representation}\phantomsection
\label{supp_table_8}

Temporal aggregation granularity (quarterly vs half-year) for the time-varying representation was treated as a hyperparameter and selected using validation PR-AUC within each model and prediction horizon. The table reports the selected aggregation level.
{
\renewcommand{\arraystretch}{1.1}

\begin{xltabular}{\textwidth}{l X c c c c}\label{tab:model_frequencies} \\
\toprule
\textbf{Model Class} & \textbf{Model} & \textbf{3 Months} & \textbf{6 Months} & \textbf{9 Months} & \textbf{12 Months} \\
\midrule
\endfirsthead

\toprule
\textbf{Model Class} & \textbf{Model} & \textbf{3 Months} & \textbf{6 Months} & \textbf{9 Months} & \textbf{12 Months} \\
\midrule
\endhead

\midrule
\multicolumn{6}{r}{\textit{Continued on next page}} \\
\endfoot

\bottomrule
\endlastfoot

\multirow{2}{*}{\makecell[l]{Large Language\\Models}} 
 & Llama-3.1-8B & Quarterly & Half-Year & Quarterly & Quarterly \\
 & OpenBioLLM-8B & Half-Year & Quarterly & Half-Year & Quarterly \\
\midrule

\multirow{3}{*}{Machine Learning} 
 & Elastic Net LR & Quarterly & Quarterly & Half-Year & Quarterly \\
 & Random Forest & Quarterly & Half-Year & Half-Year & Quarterly \\
 & XGBoost & Quarterly & Quarterly & Quarterly & Quarterly \\
\midrule

\multirow{2}{*}{\makecell[l]{Masked Language\\Models}} 
 & BioClinical ModernBERT & Quarterly & Half-Year & Quarterly & Quarterly \\
 & ModernBERT & Quarterly & Half-Year & Quarterly & Quarterly \\
\end{xltabular}
}

\clearpage
\section*{Supplementary Table 9. Fill strategy rules used to construct time-varying predictors}\phantomsection
\label{supp_table_9}

{
\scriptsize
\renewcommand{\arraystretch}{0.95}

\begin{xltabular}{\textwidth}{l X X c}
\toprule
\textbf{Predictor Type} & \textbf{Predictor} & \textbf{Mode} & \textbf{Quarters} \\
\midrule
\endfirsthead

\toprule
\textbf{Predictor Type} & \textbf{Predictor} & \textbf{Mode} & \textbf{Quarters} \\
\midrule
\endhead

\midrule
\multicolumn{4}{r}{\textit{Continued on next page}} \\
\endfoot

\bottomrule
\endlastfoot

\textbf{Mental Health} & Anxiety Disorder & Recurrent Time-Limited & 2 \\
 & Bipolar Disorder & Chronic Persistent & \\
 & Dementia & Chronic Persistent & \\
 & Depression & Recurrent Time-Limited & 2 \\
 & Other Neurological Disorders & Chronic Persistent & \\
 & Posttraumatic Stress Disorder & Recurrent Time-Limited & 2 \\
 & Psychoses & Chronic Persistent & \\
 & Sleep Disorder & Recurrent Time-Limited & 2 \\
\midrule

\textbf{Physical Health} & AIDS/HIV & Chronic Persistent & \\
 & Blood Loss Anemia & Recurrent Time-Limited & 2 \\
 & Cardiac Arrhythmia & Chronic Persistent & \\
 & Cardiovascular Disease & Chronic Persistent & \\
 & Chronic Pulmonary Disease & Chronic Persistent & \\
 & Cirrhosis & Chronic Persistent & \\
 & Coagulopathy & Chronic Persistent & \\
 & Congestive Heart Failure & Chronic Persistent & \\
 & Deficiency Anemia & Recurrent Time-Limited & 2 \\
 & Diabetes & Chronic Persistent & \\
 & Fluid And Electrolyte Disorders & Episodic & \\
 & Hepatitis & Recurrent Time-Limited & 2 \\
 & Hypertension & Chronic Persistent & \\
 & Hypothyroidism & Chronic Persistent & \\
 & Influenza & Episodic & \\
 & Liver Disease & Chronic Persistent & \\
 & Lymphoma & Ever-History & \\
 & Metastatic Cancer & Ever-History & \\
 & Obesity & Chronic Persistent & \\
 & Pain & Recurrent Time-Limited & 1 \\
 & Paralysis & Chronic Persistent & \\
 & Peptic Ulcer Disease & Recurrent Time-Limited & 2 \\
 & Peripheral Vascular Disorders & Chronic Persistent & \\
 & Pulmonary Circulation Disorders & Recurrent Time-Limited & 2 \\
 & Renal Failure & Chronic Persistent & \\
 & Rheumatoid Arthritis/Collagen & Chronic Persistent & \\
 & Solid Tumor Without Metastasis & Ever-History & \\
 & Traumatic Brain Injury & Recurrent Time-Limited & 2 \\
 & Valvular Disease & Chronic Persistent & \\
 & Weight Loss & Recurrent Time-Limited & 2 \\
\midrule

\textbf{Substance Abuse} & Alcohol Use Disorder & Chronic Persistent & \\
 & Cannabis & Recurrent Time-Limited & 2 \\
 & Cocaine & Recurrent Time-Limited & 2 \\
 & Drug Abuse & Recurrent Time-Limited & 2 \\
 & Hallucinogen & Recurrent Time-Limited & 2 \\
 & Nicotine Dependence & Chronic Persistent & \\
 & Opioid Use Disorder & Chronic Persistent & \\
 & Other Stimulant & Recurrent Time-Limited & 2 \\
\midrule

\textbf{SDOH Factors} & Employment Or Financial Problems & Recurrent Time-Limited & 2 \\
 & Food Insecurity & Recurrent Time-Limited & 2 \\
 & Housing Problems & Recurrent Time-Limited & 2 \\
 & Legal Problems & Recurrent Time-Limited & 2 \\
 & Non Specific Psychosocial Needs & Recurrent Time-Limited & 2 \\
 & Social Or Familial Problems & Recurrent Time-Limited & 2 \\
 & Violence Problems & Episodic & \\
\end{xltabular}
}

%% file: sample.bib
@article{tsai2021problem,
  title={The problem of veteran homelessness: An update for the new decade},
  author={Tsai, Jack and Pietrzak, Robert H and Szymkowiak, Dorota},
  journal={American journal of preventive medicine},
  volume={60},
  number={6},
  pages={774--780},
  year={2021},
  publisher={Elsevier}
}

@article{saleem2013next,
  title={The next-generation electronic health record: perspectives of key leaders from the US Department of Veterans Affairs},
  author={Saleem, Jason J and Flanagan, Mindy E and Wilck, Nancy R and Demetriades, Jim and Doebbeling, Bradley N},
  journal={Journal of the American Medical Informatics Association},
  volume={20},
  number={e1},
  pages={e175--e177},
  year={2013},
  publisher={BMJ Publishing Group BMA House, Tavistock Square, London, WC1H 9JR}
}

@article{fink2022comparing,
  title={Comparing mental and physical health of US veterans by VA healthcare use: implications for generalizability of research in the VA electronic health records},
  author={Fink, David S and Stohl, Malka and Mannes, Zachary L and Shmulewitz, Dvora and Wall, Melanie and Gutkind, Sarah and Olfson, Mark and Gradus, Jaimie and Keyhani, Salomeh and Maynard, Charles and others},
  journal={BMC health services research},
  volume={22},
  number={1},
  pages={1500},
  year={2022},
  publisher={Springer}
}

@article{tsai2017one,
  title={One-year incidence and predictors of homelessness among 300,000 US Veterans seen in specialty mental health care.},
  author={Tsai, Jack and Hoff, Rani A and Harpaz-Rotem, Ilan},
  journal={Psychological Services},
  volume={14},
  number={2},
  pages={203},
  year={2017},
  publisher={Educational Publishing Foundation}
}

@article{elbogen2025identifying,
  title={Identifying Prevention Targets for Homelessness Among Recently Discharged US Veterans Across Systems},
  author={Elbogen, Eric B and Pugh, Mary Jo and Amuan, Megan and Blakey, Shannon M and Graziano, Robert C and Nelson, Richard E and Jones, Audrey L and Tsai, Jack},
  journal={Health Services Insights},
  volume={18},
  pages={11786329251375179},
  year={2025},
  publisher={SAGE Publications Sage UK: London, England}
}

@article{tsai2025retrospective,
  title={Retrospective Study of Homelessness among Transitioning Service Members Within Two Years after Military Service},
  author={Tsai, Jack and Szymkowiak, Dorota},
  journal={Administration and Policy in Mental Health and Mental Health Services Research},
  pages={1--10},
  year={2025},
  publisher={Springer}
}

@article{tsai2024predicting,
  title={Predicting homelessness among transitioning US Army soldiers},
  author={Tsai, Jack and Szymkowiak, Dorota and Hooshyar, Dina and Gildea, Sarah M and Hwang, Irving and Kennedy, Chris J and King, Andrew J and Koh, Katherine A and Luedtke, Alex and Marx, Brian P and others},
  journal={American journal of preventive medicine},
  volume={66},
  number={6},
  pages={999--1007},
  year={2024},
  publisher={Elsevier}
}

@inproceedings{chatterjee2025measurement,
  title={Measurement bias in documentation of social risk among medicare beneficiaries},
  author={Chatterjee, Paula and Macneal, Eliza and Roberts, Eric T},
  booktitle={JAMA Health Forum},
  volume={6},
  number={7},
  pages={e251923--e251923},
  year={2025},
  organization={American Medical Association}
}

@article{devanarayan2025association,
  title={Association of ICD-10 Z code-documented social determinants of health with emergency department outcomes in ectopic pregnancy},
  author={Devanarayan, Priya and Farber, Cassandra and Stancliff, Hayes and Marco, Catherine A},
  journal={The American Journal of Emergency Medicine},
  year={2025},
  publisher={Elsevier}
}

@inproceedings{hau2025social,
  title={Social Determinants of Health ICD-10 Code Use by a Large Integrated Healthcare System},
  author={Hau, Cynthia and Grubber, Janet M and Ferguson, Ryan E and Cushman, William C and Ishani, Areef and Glassman, Peter A and Hynes, Colleen A and Leatherman, Sarah M},
  booktitle={Healthcare},
  volume={13},
  number={21},
  pages={2710},
  year={2025},
  organization={MDPI}
}

@article{montgomery2020housing,
  title={Housing instability and homeless program use among veterans: The intersection of race, sex, and homelessness},
  author={Montgomery, Ann Elizabeth and Szymkowiak, Dorota and Tsai, Jack},
  journal={Housing Policy Debate},
  volume={30},
  number={3},
  pages={396--408},
  year={2020},
  publisher={Taylor \& Francis}
}

@article{o2016tailoring,
  title={Tailoring care to vulnerable populations by incorporating social determinants of health: the Veterans Health Administration’s “Homeless Patient Aligned Care Team” Program},
  author={O’Toole, Thomas P and Johnson, Erin E and Aiello, Riccardo and Kane, Vincent and Pape, Lisa},
  journal={Preventing chronic disease},
  volume={13},
  pages={E44},
  year={2016}
}

@article{montgomery2020demographic,
  title={Demographic correlates of veterans’ adverse social determinants of health},
  author={Montgomery, Ann Elizabeth and Tsai, Jack and Blosnich, John R},
  journal={American journal of preventive medicine},
  volume={59},
  number={6},
  pages={828--836},
  year={2020},
  publisher={Elsevier}
}

@article{tsai2015risk,
  title={Risk factors for homelessness among US veterans},
  author={Tsai, Jack and Rosenheck, Robert A},
  journal={Epidemiologic reviews},
  volume={37},
  number={1},
  pages={177--195},
  year={2015},
  publisher={Oxford University Press}
}

@article{yang2025predicting,
  title={Predicting suicide death among veterans after psychiatric hospitalization using transformer based models with social determinants and NLP},
  author={Yang, Zhichao and Mitra, Avijit and Hu, Wen and Berlowitz, Dan and Yu, Hong},
  journal={Scientific Reports},
  year={2025},
  publisher={Nature Publishing Group UK London}
}

@article{russell2023implementing,
  title={Implementing a social needs screening and referral program among veterans: Assessing Circumstances \& Offering Resources for Needs (ACORN)},
  author={Russell, Lauren E and Cohen, Alicia J and Chrzas, Steven and Halladay, Christopher W and Kennedy, Meaghan A and Mitchell, Kathleen and Moy, Ernest and Lehmann, Lisa Soleymani},
  journal={Journal of general internal medicine},
  volume={38},
  number={13},
  pages={2906--2913},
  year={2023},
  publisher={Springer}
}

@article{davidson2019screening,
  title={Screening for social determinants of health: the known and unknown},
  author={Davidson, Karina W and McGinn, Thomas},
  journal={Jama},
  volume={322},
  number={11},
  pages={1037--1038},
  year={2019},
  publisher={American Medical Association}
}

@article{washington2010risk,
  title={Risk factors for homelessness among women veterans},
  author={Washington, Donna L and Yano, Elizabeth M and McGuire, James and Hines, Vivian and Lee, Martin and Gelberg, Lillian},
  journal={Journal of health care for the poor and underserved},
  volume={21},
  number={1},
  pages={82--91},
  year={2010},
  publisher={Johns Hopkins University Press}
}

@book{tsai2019homelessness,
  title={Homelessness among US veterans: Critical perspectives},
  author={Tsai, Jack},
  year={2019},
  publisher={Oxford University Press}
}

@article{o2018population,
  title={Population-tailored care for homeless veterans and acute care use, cost, and satisfaction: a prospective quasi-experimental trial},
  author={O’Toole, Thomas P and Johnson, Erin E and Borgia, Matthew and Noack, Amy and Yoon, Jean and Gehlert, Elizabeth and Lo, Jeanie},
  journal={Preventing chronic disease},
  volume={15},
  pages={E23},
  year={2018}
}

@article{gundlapalli2017characteristics,
  title={Characteristics of the highest users of emergency services in veterans affairs hospitals: homeless and non-homeless},
  author={Gundlapalli, Adi V and Jones, Audrey L and PETTEY, Warren BP and MOHANTY, April and BRIGNONE, Emily and GAWRON, Lori and VANNEMAN, Megan and SAMORE, Matthew H and D FARGO, Jamison and others},
  journal={Studies in health technology and informatics},
  volume={238},
  pages={24},
  year={2017}
}

@article{tsai2013risk,
  title={Risk factors for ED use among homeless veterans},
  author={Tsai, Jack and Rosenheck, Robert A},
  journal={The American journal of emergency medicine},
  volume={31},
  number={5},
  pages={855--858},
  year={2013},
  publisher={Elsevier}
}

@article{chen2023algorithmic,
  title={Algorithmic fairness in artificial intelligence for medicine and healthcare},
  author={Chen, Richard J and Wang, Judy J and Williamson, Drew FK and Chen, Tiffany Y and Lipkova, Jana and Lu, Ming Y and Sahai, Sharifa and Mahmood, Faisal},
  journal={Nature biomedical engineering},
  volume={7},
  number={6},
  pages={719--742},
  year={2023},
  publisher={Nature Publishing Group UK London}
}

@article{harris2025evaluating,
  title={Evaluating the accuracy of the Veterans Health Administration’s REACH VET suicide prediction model for legal involved veterans},
  author={Harris, Alex HS and Finlay, Andrea K and Meerwijk, Esther L},
  journal={npj Mental Health Research},
  volume={4},
  number={1},
  pages={53},
  year={2025},
  publisher={Nature Publishing Group UK London}
}

@article{kessler2023evaluation,
  title={Evaluation of a model to target high-risk psychiatric inpatients for an intensive postdischarge suicide prevention intervention},
  author={Kessler, Ronald C and Bauer, Mark S and Bishop, Todd M and Bossarte, Robert M and Castro, Victor M and Demler, Olga V and Gildea, Sarah M and Goulet, Joseph L and King, Andrew J and Kennedy, Chris J and others},
  journal={JAMA psychiatry},
  volume={80},
  number={3},
  pages={230--240},
  year={2023},
  publisher={American Medical Association}
}

@article{vickers2016net,
  title={Net benefit approaches to the evaluation of prediction models, molecular markers, and diagnostic tests},
  author={Vickers, Andrew J and Van Calster, Ben and Steyerberg, Ewout W},
  journal={bmj},
  volume={352},
  year={2016},
  publisher={British Medical Journal Publishing Group}
}

@article{tsai2022developing,
  title={Developing an operational definition of housing instability and homelessness in Veterans Health Administration’s medical records},
  author={Tsai, Jack and Szymkowiak, Dorota and Jutkowitz, Eric},
  journal={PLoS One},
  volume={17},
  number={12},
  pages={e0279973},
  year={2022},
  publisher={Public Library of Science San Francisco, CA USA}
}

@article{nilsson2019individual,
  title={Individual-level predictors for becoming homeless and exiting homelessness: a systematic review and meta-analysis},
  author={Nilsson, Sandra Feodor and Nordentoft, Merete and Hjorth{\o}j, Carsten},
  journal={Journal of urban health},
  volume={96},
  number={5},
  pages={741--750},
  year={2019},
  publisher={Springer}
}

@article{hu2022lora,
  title={Lora: Low-rank adaptation of large language models.},
  author={Hu, Edward J and Shen, Yelong and Wallis, Phillip and Allen-Zhu, Zeyuan and Li, Yuanzhi and Wang, Shean and Wang, Lu and Chen, Weizhu and others},
  journal={ICLR},
  volume={1},
  number={2},
  pages={3},
  year={2022}
}

@inproceedings{warner2025smarter,
  title={Smarter, better, faster, longer: A modern bidirectional encoder for fast, memory efficient, and long context finetuning and inference},
  author={Warner, Benjamin and Chaffin, Antoine and Clavi{\'e}, Benjamin and Weller, Orion and Hallstr{\"o}m, Oskar and Taghadouini, Said and Gallagher, Alexis and Biswas, Raja and Ladhak, Faisal and Aarsen, Tom and others},
  booktitle={Proceedings of the 63rd Annual Meeting of the Association for Computational Linguistics (Volume 1: Long Papers)},
  pages={2526--2547},
  year={2025}
}

@article{pedregosa2011scikit,
  title={Scikit-learn: Machine learning in Python},
  author={Pedregosa, Fabian and Varoquaux, Ga{\"e}l and Gramfort, Alexandre and Michel, Vincent and Thirion, Bertrand and Grisel, Olivier and Blondel, Mathieu and Prettenhofer, Peter and Weiss, Ron and Dubourg, Vincent and others},
  journal={the Journal of machine Learning research},
  volume={12},
  pages={2825--2830},
  year={2011},
  publisher={JMLR. org}
}

@article{chen2016xgboost,
  title={XGBoost: A Scalable Tree Boosting System},
  author={Chen, Tianqi},
  journal={Cornell University},
  year={2016}
}

@article{paszke2019pytorch,
  title={Pytorch: An imperative style, high-performance deep learning library},
  author={Paszke, Adam and Gross, Sam and Massa, Francisco and Lerer, Adam and Bradbury, James and Chanan, Gregory and Killeen, Trevor and Lin, Zeming and Gimelshein, Natalia and Antiga, Luca and others},
  journal={Advances in neural information processing systems},
  volume={32},
  year={2019}
}

@misc{2024AnnualHomelessness,
  title = {The 2024 {{Annual Homelessness Assessment Report}} ({{AHAR}} to {{Congress}}) {{Part}} 1: {{Point-In-Time Estimates}} of {{Homelessness}}, {{December}} 2024},
  howpublished = {\url{https://www.huduser.gov/portal/sites/default/files/pdf/2024-AHAR-Part-1.pdf}},
  note = {Accessed: 2025-12-15}
}

@misc{PEFT,
  title = {PEFT: Parameter-Efficient Fine-Tuning},
  howpublished = {\url{https://huggingface.co/docs/peft/en/index}},
  note = {Accessed: 2025-12-15}
}

@misc{2024MetallamaLlama318BHugging,
  title = {Meta-Llama/{{Llama-3}}.1-{{8B}} · {{Hugging Face}}},
  year = {2024},
  howpublished = {\url{https://huggingface.co/meta-llama/Llama-3.1-8B}},
  note = {Accessed: 2025-12-15}
}

@misc{AadityaLlama3OpenBioLLM8BHugging,
  title = {Aaditya/{{Llama3-OpenBioLLM-8B}} · {{Hugging Face}}},
  howpublished = {\url{https://huggingface.co/aaditya/Llama3-OpenBioLLM-8B}},
  note = {Accessed: 2025-12-15}
}

@misc{Transformers,
  title = {Transformers Documentation},
  howpublished = {\url{https://huggingface.co/docs/transformers/en/index}},
  note = {Accessed: 2025-12-15}
}

@article{sounack2025bioclinical,
  title={BioClinical ModernBERT: A State-of-the-Art Long-Context Encoder for Biomedical and Clinical NLP},
  author={Sounack, Thomas and Davis, Joshua and Durieux, Brigitte and Chaffin, Antoine and Pollard, Tom J and Lehman, Eric and Johnson, Alistair EW and McDermott, Matthew and Naumann, Tristan and Lindvall, Charlotta},
  journal={arXiv preprint arXiv:2506.10896},
  year={2025}
}
